%% file: main.tex
\documentclass[10pt,twocolumn,letterpaper]{article}

\usepackage{cvpr}              

\input{preamble}

\definecolor{cvprblue}{rgb}{0.21,0.49,0.74}
\usepackage[pagebackref,breaklinks,colorlinks,allcolors=cvprblue]{hyperref}

\def\paperID{xxxx} 
\def\confName{CVPR}
\def\confYear{2026}

\title{MV-Fashion: Towards Enabling Virtual Try-On and Size Estimation with Multi-View Paired Data}

\author{
Hunor Laczkó$^{1,2}$ \quad Libang Jia$^{3}$ \quad Loc-Phat Truong$^{2,3}$ \quad Diego Hernández$^{2,3}$ \\
Sergio Escalera$^{2,3}$ \quad Jordi Gonzalez$^{1,2}$ \quad Meysam Madadi$^{2,3}$ \vspace{0.1cm} \\
{\tt\small \{hunor.laczko, jordi.gonzalez\}@uab.cat, \{ljiajiax213, dhernaan8\}@alumnes.ub.edu} \\
{\tt\small \{phattruong, mmadadi, sescalera\}@ub.edu} \vspace{0.1cm} \\
$^{1}$Universitat Autònoma de Barcelona \quad $^{2}$Computer Vision Center \quad $^{3}$Universitat de Barcelona 
}

\begin{document}

\twocolumn[{%
\renewcommand\twocolumn[1][]{#1}%
\maketitle
\begin{center}
    \centering
    \captionsetup{type=figure}
    \vspace{-5mm}
    \includegraphics[width=\textwidth]{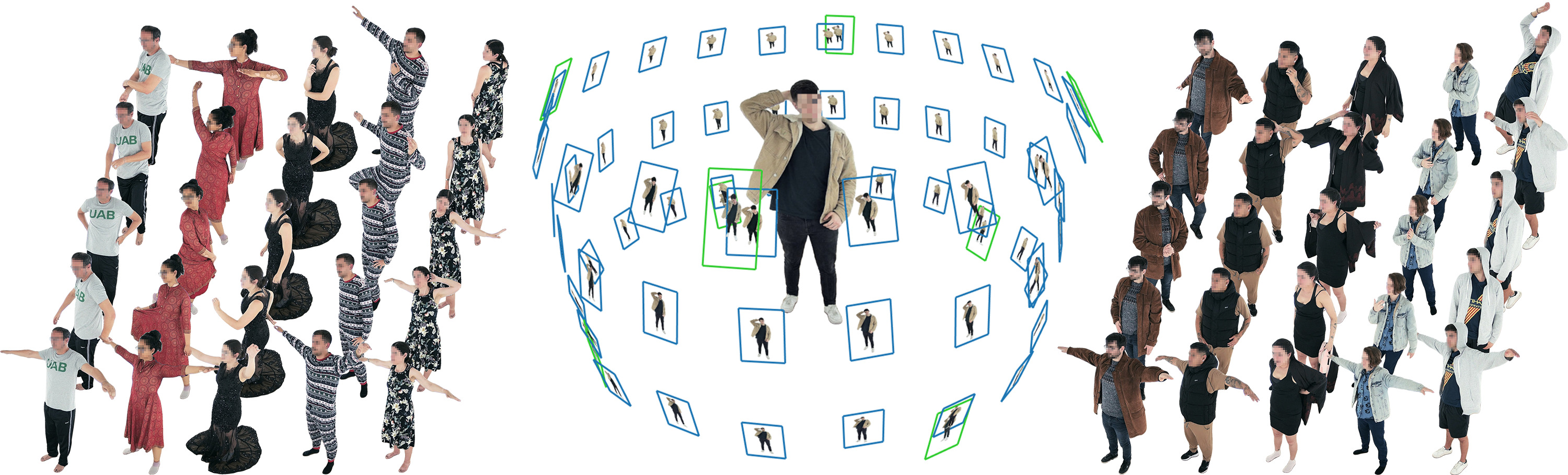}
    \captionof{figure}{We present \textbf{MV-Fashion}, a multi-view synchronized video dataset with 72.5 million frames. The dataset contains diverse clothing, multi-layered outfits annotated with draping styles, and paired catalogue domain data, ready for virtual try-on and fashion-centric tasks. The capture setup, shown in the middle, features 60 RGB cameras (blue) and 8 RGB-depth cameras (green) with 4K footage.}
    \label{fig:intro}
\end{center}%
}]

\input{sec/0_abstract}

\input{sec/1_intro}

\input{sec/2_related}
\input{sec/3_dataset}

\input{sec/4_baselines}
\input{sec/5_experiments}
\input{sec/6_conclusion}


\clearpage
{
    \small
    \bibliographystyle{ieeenat_fullname}
    \bibliography{main}
}
\input{sec/X_suppl_arxiv}

\end{document}

%% file: preamble.tex
\renewcommand{\paragraph}[1]{\vspace{.5em}\noindent\textbf{#1.}}

\usepackage{graphicx}
\usepackage[table]{xcolor}
\usepackage{subcaption}
\usepackage{pifont}   
\usepackage{multirow}

%% file: sec/0_abstract.tex
\begin{abstract}
Existing 4D human datasets fall short for fashion-specific research, lacking either realistic garment dynamics or task-specific annotations. Synthetic datasets suffer from a realism gap, whereas real-world captures lack the detailed annotations and paired data required for virtual try-on (VTON) and size estimation tasks. To bridge this gap, we introduce MV-Fashion, a large-scale, multi-view video dataset engineered for domain-specific fashion analysis. MV-Fashion features 3,273 sequences (72.5 million frames) from 80 diverse subjects wearing 3-10 outfits each. It is designed to capture complex, real-world garment dynamics, including multiple layers and varied styling (e.g. rolled sleeves, tucked shirt). A core contribution is a rich data representation that includes pixel-level semantic annotations, ground-truth material properties like elasticity, and 3D point clouds. Crucially for VTON applications, MV-Fashion provides paired data: multi-view synchronized captures of worn garments alongside their corresponding flat, catalogue images. We leverage this dataset to establish baselines for fashion-centric tasks, including virtual try-on, clothing size estimation, and novel view synthesis. The dataset is available at \url{https://hunorlaczko.github.io/MV-Fashion}.

\end{abstract}

%% file: sec/1_intro.tex
\newcommand{\cmark}{\textcolor{green!60!black}{\ding{51}}} 
\newcommand{\xmark}{\textcolor{red!70!black}{\ding{55}}}   

\begin{table*}[t]
\centering
  
\begin{subtable}[t]{0.49\textwidth}
\scriptsize
\setlength{\tabcolsep}{4pt}
\setlength\extrarowheight{-5pt}
\rowcolors{2}{gray!10}{white}
\centering

\begin{tabular}{llllllllll}
\toprule
\rowcolor{gray!10}
\textbf{Dataset} & \rotatebox{90}{\textbf{Sizes}} & \rotatebox{90}{\textbf{Styles}} & \rotatebox{90}{\textbf{Layered}} & \rotatebox{90}{\textbf{Subjects}} & \rotatebox{90}{\textbf{Sequences}} & \rotatebox{90}{\textbf{Outfits}} & \rotatebox{90}{\textbf{Frames}} & \rotatebox{90}{\textbf{RGB}} & \rotatebox{90}{\textbf{Depth}} \\ \hline
4D-Dress~\cite{wang4DDRESS4DDataset2024} & \xmark & \xmark & \cmark & 32 & 520 & 64 & 78K & 53 & \xmark \\
GeneBody~\cite{chengGeneralizableNeuralPerformer2022} & \xmark & \xmark & \xmark & 100 & 370 & \xmark & 2.95M & 48 & \xmark \\
ActorsHQ~\cite{isikHumanRFHighFidelityNeural2023} & \xmark & \xmark & \xmark & 8 & 52 & 8 & 40K & 160 & \xmark \\
SIZER~\cite{tiwariSIZERDatasetModel2020} & \cmark & \xmark & \xmark & 100 & \xmark & 10 & 2K & 130 & \xmark \\
X-humans~\cite{shenXAvatarExpressiveHuman2023} & \xmark & \xmark & \xmark & 20 & \xmark & \xmark & 35k & 53 & \xmark \\
4DHumanOutfit~\cite{armando4DHumanOutfitMultisubject4D2023b} & \xmark & \xmark & \xmark & 20 & 1,540 & 140 & 459K & 68 & \xmark \\
FeeMan~\cite{wangFreeManBenchmarking3D2024} & \xmark & \xmark & \xmark & 40 & 8,000 & \xmark & 11.3M & 8 & \xmark \\
UltraStage~\cite{zhouRelightableNeuralHuman2023} & \xmark & \xmark & \xmark & 100 & \xmark & \xmark & 192k & 32 & \xmark \\
Neural Actor~\cite{liuNeuralActorNeural2021} & \xmark & \xmark & \xmark & 4 & 4 & 4 & 87K & 86 & \xmark \\
ZJU-Mocap~\cite{pengNeuralBodyImplicit2021} & \xmark & \xmark & \xmark & 9 & 9 & 9 & 161K & 21 & \xmark \\
MPI-INF-3DHP~\cite{mehtaMonocular3DHuman2017} & \xmark & \xmark & \xmark & 8 & 64 & 16 & 1.3M & 14 & \xmark \\
NHR~\cite{wuMultiViewNeuralHuman2020} & \xmark & \xmark & \xmark & 3 & 5 & 3 & 100K & 80 & \xmark \\
CMU Panoptic~\cite{jooPanopticStudioMassively2015} & \xmark & \xmark & \xmark & 97 & 65 & 97 & 15.3M & 480 & 10 \\
AIST++~\cite{liAIChoreographerMusic2021} & \xmark & \xmark & \xmark & 30 & 1408 & 30 & 10.1M & 9 & \xmark \\
HUMBI~\cite{yuHUMBILargeMultiview2020} & \xmark & \xmark & \xmark & 772 & \xmark & 772 & 26M & 107 & \xmark \\
Humman~\cite{caiHuMManMultimodal4D2022} & \xmark & \xmark & \xmark & 1,000 & 400K & 1,000 & 60M & 10 & 10 \\
Human3.6M~\cite{ionescuHuman36MLargeScale2014} & \xmark & \xmark & \xmark & 11 & \xmark & 11 & 3.6M & 4 & 1 \\
THuman 4.0~\cite{zhengStructuredLocalRadiance2022} & \xmark & \xmark & \xmark & 3 & \xmark & 3 & 10K & 24 & \xmark \\
DyMVHumans~\cite{zhengPKUDyMVHumansMultiViewVideo2024} & \xmark & \xmark & \xmark & 32 & 2,668 & 45 & 8.2M & 60 & \xmark \\
DNA-Rendering~\cite{chengDNARenderingDiverseNeural2023} & \xmark & \xmark & \xmark & 500 & 5K & 1,500 & 67.5M & 60 & 8 \\
MVHumanNet++~\cite{liMVHumanNetLargescaleDataset2025} & \xmark & \xmark & \xmark & 4,500 & 60K & 9,000 & 645.1M & 48 & \xmark \\
\midrule 
\textbf{Ours} & \cmark & \cmark & \cmark & 80 & 3,273 & 474 & 72.5M & 68 & 8 \\ \hline
\end{tabular}
\caption{Comparison of multi-view human datasets.}
\label{table:multiview_datasets}
\end{subtable}
\hfill
\begin{subtable}[t]{0.49\textwidth}
\scriptsize
\setlength{\tabcolsep}{4pt}
\setlength\extrarowheight{-5pt}
\rowcolors{2}{gray!10}{white}
\centering
\begin{tabular}{lcccccc}
\toprule
\rowcolor{gray!10}
\textbf{Name} &
\rotatebox{90}{\textbf{MV}} &
\rotatebox{90}{\textbf{Poses}} &
\rotatebox{90}{\textbf{Paired}} &
\rotatebox{90}{\textbf{Video}} &
\rotatebox{90}{\textbf{Outfits}} &
\rotatebox{90}{\textbf{Frames}} \\
\midrule
DeepFashion~\cite{liuDeepFashionPoweringRobust2016} & \xmark & \cmark\ (4–5) & \xmark & \xmark & -- & 800K \\
DeepFashion2~\cite{geDeepFashion2VersatileBenchmark2019} & \xmark & \cmark & \xmark & \xmark & 43.8K & 491K \\
FashionPedia~\cite{jiaFashionpediaOntologySegmentation2020} & \xmark & \xmark & \xmark & \xmark & -- & 48,825 \\
Fashion-Gen~\cite{rostamzadehFashionGenGenerativeFashion2018} & \xmark & \cmark\ (1–6) & \xmark & \xmark & -- & 293,008 \\
Fashionista~\cite{yamaguchiParsingClothingFashion2012} & \xmark & \xmark & \xmark & \xmark & -- & 158,235 \\
CloSe-D~\cite{anticCloSe3DClothing2024} & \cmark\ (130) & \cmark\ (7) & \xmark & \xmark & -- & -- \\
4DHumanOutfit~\cite{armando4DHumanOutfitMultisubject4D2023b} & \cmark\ (68) & \cmark\ (multiple) & \xmark & \cmark & 140 & 459K \\
DeepFashion3D~\cite{zhuDeepFashion3DDataset2020} & \cmark\ (50) & \xmark & \xmark & \xmark & 563 & 2,078 \\
4D-DRESS~\cite{wang4DDRESS4DDataset2024} & \cmark\ (53) & \cmark\ (multiple) & \xmark & \cmark & 64 & 78K \\
VITON~\cite{hanVITONImageBasedVirtual2018} & \xmark & \xmark & \cmark & \xmark & 16,253 & 32,506 \\
VITON-HD~\cite{choiVITONHDHighResolutionVirtual2021} & \xmark & \xmark & \cmark & \xmark & 13,679 & 13,679 \\
LookBook~\cite{yooPixelLevelDomainTransfer2016} & \xmark & \cmark\ ($\sim$8) & \cmark & \xmark & 9,732 & 84,748 \\
FashionTryOn~\cite{zhengVirtuallyTryingNew2019} & \xmark & \cmark\ (2) & \cmark & \xmark & 4,327 & 86,142 \\
FashionOn~\cite{hsiehFashionOnSemanticguidedImagebased2019} & \xmark & \cmark\ (2) & \cmark & \xmark & 10,895 & 32,685 \\
MPV~\cite{dongMultiPoseGuidedVirtual2019} & \xmark & \cmark\ ($\sim$3) & \cmark & \xmark & 13,524 & 35,687 \\
DressCode~\cite{morelliDressCodeHighResolution2022b} & \xmark & \cmark\ (2–5) & \cmark & \xmark & 53,792 & 107,584 \\
IGPair~\cite{shenIMAGDressingv1CustomizableVirtual2024} & \xmark & \cmark\ (2-5) & \cmark & \xmark & 86,873 & 649,714 \\
MVG~\cite{wangMVVTONMultiViewVirtual2025b} & \xmark & \cmark\ (5) & \cmark & \xmark & 1,009 & 5,045 \\
VVT~\cite{dongFWGANFlowNavigatedWarping2019} & \xmark & \cmark\ (multiple) & \cmark & \cmark & 791 & 190,101 \\
TikTokDress~\cite{nguyenSwiftTryFastConsistent2025} & \xmark & \cmark\ (multiple) & \cmark & \cmark & 817 & 270K \\
ViViD~\cite{fangViViDVideoVirtual2024a} & \xmark & \cmark\ (multiple) & \cmark & \cmark & 9,700 & 1,2M \\
\midrule
\textbf{Ours} & \cmark\ (68) & \cmark\ (multiple) & \cmark & \cmark & 474 & 72.5M \\
\bottomrule
\end{tabular}
\caption{Comparison of fashion-centric datasets.}
\label{table:vton_datasets}
\end{subtable}

\caption{We compare the proposed \textbf{MV-Fashion dataset} with existing datasets, which we divide into two categories: (a) multi-view human, and (b) fashion-centric datasets. For the multi-view human comparison, we evaluate attributes such as \textit{Size} (inclusion of clothing measurements), \textit{Styles} (style variations of the same outfit), \textit{Layered} (multiple clothing layers in the same outfit), \textit{Subjects} (number of participants), and \textit{Sequences} (number of videos), along with the number of \textit{RGB} and \textit{Depth} cameras in the capture setup. For the fashion-centric comparison, our criteria include: \textit{MV} for Multi-View (use of multiple synchronized cameras), \textit{Poses} (different poses in the same clothing; 'multiple' refers to video sequences with many poses), \textit{Paired} (linking images of the worn outfit to garment photos) and \textit{Video} (static images vs. video). In both cases, we also add \textit{Outfits} (number of outfits), and \textit{Frames} (number of RGB frames). In this table, wherever used, 'K' denotes thousands ($10^{3}$) and 'M' denotes millions ($10^{6}$). \cmark\ indicates available/yes, \xmark\ indicates not available/no/unspecified.}
\label{tab:main}
\end{table*}

\section{Introduction}
\label{sec:intro}

Understanding and modeling human clothing is fundamental for applications ranging from 3D games and digital avatars to animation and virtual try-on (VTON) systems. Accurately modeling 3D clothes enhances VTON, reduces designer workloads, and enables data-driven physics simulations. In e-commerce, accurately predicting fit and size is crucial to reduce retailer losses from returns caused by poor online fit assessment~\cite{duhouxVolumesDestructionReturned2024}. Developing realistic reconstruction and simulation algorithms also advances sustainability by reducing physical samples, overproduction, and the environmental impact of returns~\cite{duhouxVolumesDestructionReturned2024}.

However, this research area presents significant challenges because garments exhibit substantial variability in terms of size, shape, topology, and fabric, which complicates the generation of representative 3D data. A core technical hurdle is modeling dynamic deformations, particularly those associated with loose clothing like dresses or coats, which often fail to be captured realistically by models relying solely on basic skeletal or parametric body priors. 

While many fashion-centric datasets have been introduced, the current landscape remains fragmented. Research is largely split between two types of datasets, neither of which is sufficient on its own. Standard 2D VTON datasets~\cite{choiVITONHDHighResolutionVirtual2021, morelliDressCodeHighResolution2022b,shenIMAGDressingv1CustomizableVirtual2024} offer vital catalogue-to-worn image pairs but lack 3D geometric or multi-view data. Conversely, existing 3D and 4D fashion datasets~\cite{zhuDeepFashion3DDataset2020, wang4DDRESS4DDataset2024} capture geometry and motion but omit the paired catalogue data essential for VTON tasks. This creates a critical gap: no single dataset provides the synchronized multi-view data and the explicit catalogue-to-worn pairings required to jointly tackle garment dynamics, size estimation, and VTON.

Furthermore, existing datasets for dynamic human modeling, despite rapid progress, remain ill-suited for fashion-specific research. Synthetic datasets~\cite{berticheCLOTH3DClothed3D2020,zouCLOTH4DDatasetClothed2023} offer large-scale ground-truth but suffer from a realism gap, failing to capture complex garment dynamics. Meanwhile, real-world capture datasets like ActorsHQ~\cite{isikHumanRFHighFidelityNeural2023}, DNA-Rendering~\cite{chengDNARenderingDiverseNeural2023} and MVHumanNet++~\cite{liMVHumanNetLargescaleDataset2025} provide high-fidelity appearance and large-scale datasets but are designed for general human motion. They lack the specific focus, annotations, and data pairings useful for fashion-centric tasks. Finally, these high-resolution, multi-view datasets introduce significant computational complexity and scalability challenges. 

In this work, we introduce MV-Fashion, a new multi-view video dataset engineered specifically to bridge the gap between general multi-view human capture and domain-specific fashion analysis. We balance data quality and acquisition efficiency through a custom multi-view capture setup based on affordable, off-the-shelf hardware. Our dataset provides a unique combination of scale and fashion-centric detail, curated to address fundamental challenges in garment-centric applications. It explicitly captures complex clothing behaviors, such as multi-layering and diverse styling configurations (e.g., tucked shirts, rolled sleeves), thereby providing a robust benchmark for future research. To support applications in virtual try-on and clothing size estimation, we provide extensive ground-truth annotations. These include manually captured material properties (elasticity, fabric type), sizing charts, categorical labels, garment-level segmentations, body pose and shape estimations, and point clouds recovered from depth sensors. A key innovation for virtual try-on is the inclusion of paired flat/catalogue images of the garments for every multi-view video sequence. See Tab.~\ref{tab:main} for a comprehensive comparison of MV-Fashion with other multi-view and fashion datasets. We leverage this comprehensive data to establish several baselines for virtual try-on, cloth size estimation, and novel view synthesis.

In summary, our key contributions are:
\begin{itemize}
    \item \textbf{A large-scale, multi-view video fashion dataset}, captured with a resource-friendly, multi-view setup using 68 cameras.
    \item \textbf{Unique, challenging capture scenarios} designed to test garment dynamics, with multi-layered outfits and diverse styling variations.
    \item \textbf{Rich, multi-modal data} essential for fashion applications, including pixel-level segmentations, ground-truth garment material/elasticity data for physics, categorical labels, detailed sizing charts, SMPL-X~\cite{pavlakosExpressiveBodyCapture2019} pose and shape parameters, and 3D point clouds.
    \item \textbf{Catalogue-to-multi-view pairing} providing flat garment images for every multi-view video sequence, unlocking new avenues for VTON and size estimation.
    \item \textbf{Baseline benchmarks} for virtual try-on, cloth size estimation and novel view synthesis.
\end{itemize}

%% file: sec/2_related.tex
\section{Related work}
\label{sec:related}

Progress in learning-based clothing reconstruction, simulation, and view synthesis depends on dataset realism and task alignment. Existing datasets can be grouped into those generated synthetically, those captured from the real world, and those tailored specifically for VTON applications.

\subsection{Synthetic Datasets}
Synthetic datasets are created using graphics engines~\cite{blender, unrealengine} and simulation tools~\cite{clo}, enabling scalable generation of massive volumes of data with perfect ground-truth semantic labels available by design~\cite{zouCLOTH4DDatasetClothed2023}. Datasets like CLOTH3D and CLOTH4D~\cite{berticheCLOTH3DClothed3D2020,zouCLOTH4DDatasetClothed2023} are large-scale with rich data on garment dynamics, size, and fabric. BEDLAM~\cite{blackBEDLAMSyntheticDataset2023} and BEDLAM2.0~\cite{tesch2025bedlam2} focuses on lifelike animated motion across body shapes and outfit types. Other specialized datasets, such as D-LAYERS~\cite{shaoMultiLayered3DGarments2023} and ReSynth~\cite{maPowerPointsModeling2021}, focus on multi-layer clothing and point cloud representations respectively. However, these datasets suffer from a realism gap, struggling to model complex human appearance, motion, and clothing deformation. Even photorealistic attempts~\cite{blackBEDLAMSyntheticDataset2023} fail to precisely capture the dynamics of real-world garments.

\subsection{Real-World Datasets}
To capture authentic clothing dynamics and appearance, many researchers employ multi-view volumetric capture systems~\cite{colletHighQualityStreamableFreeViewpoint2015, jooPanopticStudioMassively2015} to record clothed humans. These datasets vary significantly in scale, fidelity, and the types of data and annotations. For a direct comparison, refer to Tab.~\ref{table:multiview_datasets}.

\textbf{Foundational and Large-Scale Datasets} such as Human3.6M~\cite{ionescuHuman36MLargeScale2014} and CMU Panoptic Studio~\cite{jooPanopticStudioMassively2015} are established benchmarks for large-scale human capture. Human3.6M provides 3.6 million 3D poses~\cite{ionescuHuman36MLargeScale2014}, while CMU Panoptic Studio utilizes a massive camera array for capturing social interactions~\cite{jooPanopticStudioMassively2015}. More recent large-scale efforts include HuMMan~\cite{caiHuMManMultimodal4D2022}, which offers multi-modal data for 1,000 subjects with a focus on versatile sensing and modeling. DNA-Rendering~\cite{chengDNARenderingDiverseNeural2023} increases diversity by featuring 500 subjects/1,500 outfits captured by a multi-camera setup. Currently, the largest dataset by frame count, MVHumanNet++~\cite{liMVHumanNetLargescaleDataset2025}, provides 645 million frames from 4,500 subjects in daily dressing, incorporating enhanced annotations such as pseudo normal and depth maps. Some clothing-specific large-scale datasets include 4D-DRESS~\cite{wang4DDRESS4DDataset2024}, a 4D dataset with high-quality textured scans accompanied by vertex-level semantic annotations and segmented garment meshes for 64 outfits/520 sequences. SIZER~\cite{tiwariSIZERDatasetModel2020} focuses specifically on garment fitting and sizing, capturing 100 diverse subjects. Similarly, X-Humans~\cite{shenXAvatarExpressiveHuman2023} and 4DHumanOutfit~\cite{armando4DHumanOutfitMultisubject4D2023b} provide clothed scans but lack the depth of semantic annotations and sizing variations.

Despite their impressive scale, these datasets have limitations for fashion-centric applications. They lack fine-grained semantic garment annotations, material data, and paired catalogue images. Furthermore, the resources required for processing this immense volume of data are substantial, restricting their accessibility for constrained setups.

\textbf{High-Fidelity and Specialised Datasets} prioritize high-quality capture for specific applications. Examples include ActorsHQ~\cite{isikHumanRFHighFidelityNeural2023} for novel view synthesis, UltraStage~\cite{zhouRelightableNeuralHuman2023} for relighting, AIST++~\cite{liAIChoreographerMusic2021} with paired music and 3D dance motion, Hi4D~\cite{yinHi4D4DInstance2023} with human-human interaction, and FreeMan~\cite{wangFreeManBenchmarking3D2024}, collected entirely ``in-the-wild" using synchronized smartphone cameras. While these datasets offer unique features or high-quality data, their applicability is limited for general-purpose research.

\subsection{Fashion and Virtual Try-On Datasets} 
\label{related_work:vton_ds}
Datasets for fashion analysis and VTON vary significantly in dimensionality and in the availability of paired data, i.e. the explicit link between a garment's catalogue (product) image and its worn (on-person) state. This pairing is crucial for VTON~\cite{choiVITONHDHighResolutionVirtual2021}. Existing works, also shown in Tab.~\ref{table:vton_datasets}, fall into the following distinct categories: \textit{(1) Unpaired 2D datasets} are foundational, large-scale 2D image collections used for recognition, retrieval, and parsing. However, they lack the paired images needed for VTON. This category includes DeepFashion~\cite{liuDeepFashionPoweringRobust2016}, DeepFashion2~\cite{geDeepFashion2VersatileBenchmark2019}, Fashionista~\cite{yamaguchiParsingClothingFashion2012}, Fashion-Gen~\cite{rostamzadehFashionGenGenerativeFashion2018}, and FashionPedia~\cite{jiaFashionpediaOntologySegmentation2020}. \textit{(2) Paired 2D datasets} provide the paired catalogue-to-worn images essential for VTON synthesis but lack 3D geometric information, for example, the standard single-pose datasets VITON~\cite{hanVITONImageBasedVirtual2018} and VITON-HD~\cite{choiVITONHDHighResolutionVirtual2021}. Datasets like Dress Code~\cite{morelliDressCodeHighResolution2022b}, IGPair~\cite{shenIMAGDressingv1CustomizableVirtual2024}, MVG~\cite{wangMVVTONMultiViewVirtual2025b}, MPV~\cite{dongMultiPoseGuidedVirtual2019}, LookBook~\cite{yooPixelLevelDomainTransfer2016}, FashionOn~\cite{hsiehFashionOnSemanticguidedImagebased2019}, and FashionTryOn~\cite{zhengVirtuallyTryingNew2019}, further use multi-pose setup with 2 to 8 different poses per subject. VVT~\cite{dongFWGANFlowNavigatedWarping2019}, TikTokDress~\cite{nguyenSwiftTryFastConsistent2025}, and ViViD~\cite{fangViViDVideoVirtual2024a} capture subjects in motion from a single video. Despite their diversity, these datasets lack the synchronized multi-view data to unlock novel approaches. \textit{(3) 3D/4D geometric} datasets provide crucial shape and dynamics information. Examples include 3D datasets like DeepFashion3D~\cite{zhuDeepFashion3DDataset2020}, GarVerseLOD~\cite{luoGarVerseLODHighFidelity3D2024}, and CloSe-D~\cite{anticCloSe3DClothing2024}, and dynamic 4D datasets like Thuman 4.0~\cite{zhengStructuredLocalRadiance2022}, 4D-DRESS~\cite{wang4DDRESS4DDataset2024}, and 4DHumanOutfit~\cite{armando4DHumanOutfitMultisubject4D2023b}. While invaluable for shape and motion analysis, their lack of paired data restricts their use in VTON.

Overall, the field lacks a dataset that bridges the gap between VTON and multi-view capture. VTON-focused datasets~\cite{choiVITONHDHighResolutionVirtual2021, morelliDressCodeHighResolution2022b,shenIMAGDressingv1CustomizableVirtual2024} are 2D-limited, while 4D geometric ones~\cite{wang4DDRESS4DDataset2024, tiwariSIZERDatasetModel2020} lack the necessary catalogue-to-worn pairings. Existing real-world capture methods often require resource-heavy setups~\cite{isikHumanRFHighFidelityNeural2023}, and few datasets adequately include complex, unexplored scenarios like multi-layering or diverse styling. We therefore introduce MV-Fashion, providing an accessible, multi-view dataset that explicitly links multi-view video data of complex styling with static catalogue images to jointly address garment dynamics, size estimation, and virtual try-on.

%% file: sec/3_dataset.tex
\section{MV-Fashion}
\label{sec:dataset}
This section provides an overview of MV-Fashion, outlining the data acquisition setup and protocol, summarizing relevant statistics, and describing the high-quality annotations designed to facilitate downstream applications.

\subsection{Dataset Acquisition }

\textbf{Setup.} The acquisition setup is built upon a custom hybrid multi-view capture system, utilizing accessible and readily available hardware. Specifically, we use 60 Raspberry Pi global shutter cameras (1.6 MP), and 8 Orbbec Femto Bolt cameras to provide complementary depth and 4K video streams. The cameras are mounted in three rows on a 20-sided aluminium frame to provide dense coverage around the subject. To ensure precise temporal alignment, all cameras operate under an external electronic synchronization, achieving a frame alignment of less than 2 milliseconds across all cameras. The controlled illumination setup, consisting of 40 high-power LED panels, ensures consistent lighting and allows for low exposure times to mitigate motion blur during sequence recording. Fig.~\ref{fig:intro} illustrates the capture setup and shows representative examples of recorded subjects and outfits. 

\begin{figure*}
    \centering
    \includegraphics[width=\linewidth]{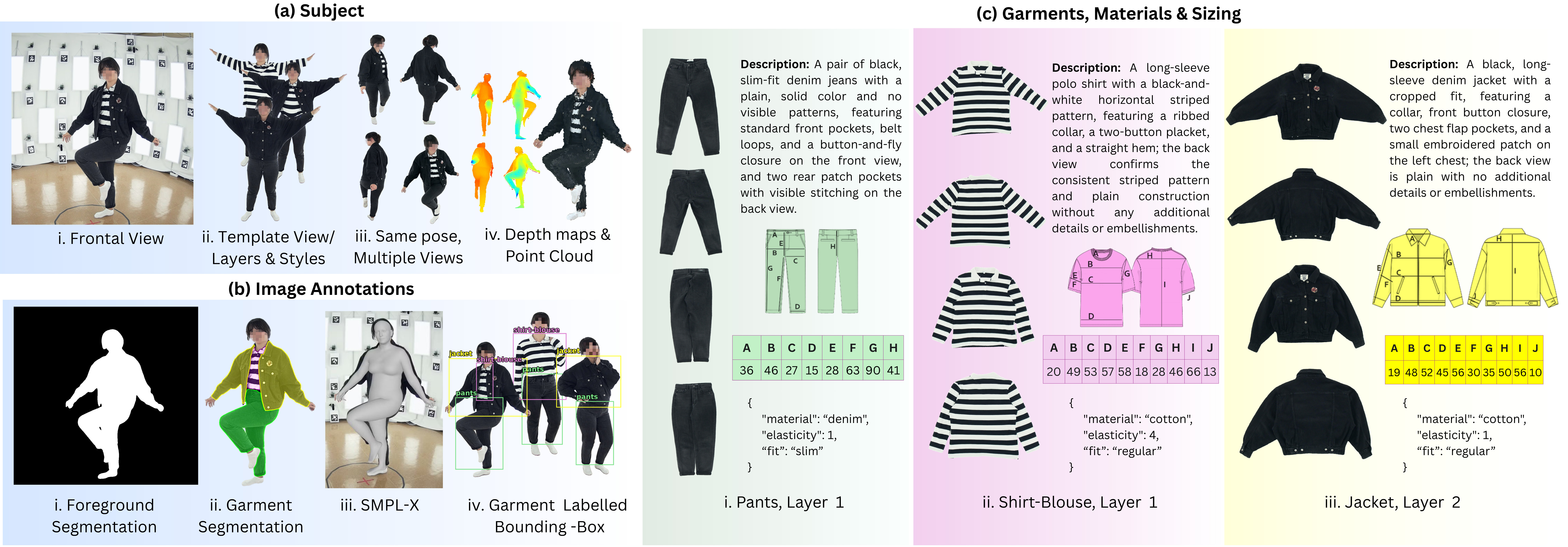}
    \caption{Available data for one subject in a single clothing set. 
\textit{(a) Subject:} Frontal view, clothing layers (w \& w/o jacket), styles (open \& closed jacket) and template recordings, four representative RGB views, and depth images with the reconstructed point cloud. 
\textit{(b) Image Annotations:} Foreground segmentation isolating the subject, garment segmentation for each clothing item, fitted SMPL-X body model, and labelled bounding boxes for all garments in each frame. 
\textit{(c) Garments, Materials \& Sizing:} For each garment, frontal and back views in stretched and normal states, text description of the clothing, sizing chart with corresponding measurements (in centimeters), and properties.}
    \label{fig:annotations}
\end{figure*}

\textbf{Collection Protocol.} Our recording protocol defines three sequence types, each recorded as a 20-second video at 15 fps: \textit{ (1) Body sequence} serves as a baseline, capturing the subject in minimal clothing performing fixed, canonical poses for body-shape estimation. \textit{(2) Template sequence} captures the static geometry of each outfit using the same fixed poses. \textit{(3) Motion sequence} captures the subject performing five randomly selected fashion-oriented poses to model dynamic garment behavior. To acquire rich garment details and variety, we record multiple motion sequences for each outfit. This process includes capturing different styling variations when possible, as well as multi-layer scenarios where new garments are added. To complement the visual data, we record garment measurements, material data, and other metadata, as described in Sec.~\ref{sec:annotations}. Additional technical details about the capture setup and recording protocols are provided in the Suppl. Sec.~\ref{supp:data_acq}. 

\textbf{Ethics Considerations.} The data collection was reviewed and approved by the authors' university data ethics board, and all participants provided written informed consent prior to the study. Although we aimed to collect a balanced dataset, geographical constraints led to a dominance of European and American participants, which may bias both the data and the derived baselines. Consequently, any models trained on MV-Fashion may encode inherent biases related to underrepresented ethnicities, cultural backgrounds, clothing norms or fashion trends, or they might not generalize well for different populations. 

\subsection{Dataset Statistics}
The data collection process systematically captures a wide range of human dynamics and appearances, featuring video recordings of 80 subjects performing diverse dynamic actions. This cohort has a gender distribution of 50.6\% male, 45.7\% female, and 3.7\% non-binary. The subjects' origins are primarily Europe (56.8\%), followed by South America (28.4\%), North America (9.9\%), and Asia (4.9\%).

Each subject is recorded in 3 to 10 different outfits, encompassing multiple garment types, including challenging configurations with multiple layers of clothing. The resulting dataset comprises 3,273 video sequences. In total, there are 474 sets of outfits, constituted by 754 individual pieces of garments, resulting in a total volume of 72.5 million frames. Out of these 474 outfits, 326 have a single layer of clothing, 145 have two layers, and there are 3 cases with three layers. Tab.~\ref{tab:dataset_stats} reports the distribution of garment categories in the dataset, including their frequency in layered and styled outfits.

\begin{table}[h]
\centering
\scriptsize
\setlength{\tabcolsep}{1.62pt}
\begin{tabular}{l *{14}{c}}
\toprule
 & 
\rotatebox[origin=l]{90}{\parbox{1cm}{\raggedright shirt- \\ blouse}} & 
\rotatebox[origin=l]{90}{\parbox{1cm}{\raggedright pants}} & 
\rotatebox[origin=l]{90}{\parbox{1cm}{\raggedright shorts}} & 
\rotatebox[origin=l]{90}{\parbox{1cm}{\raggedright dress}} & 
\rotatebox[origin=l]{90}{\parbox{1cm}{\raggedright jacket}} & 
\rotatebox[origin=l]{90}{\parbox{1cm}{\raggedright sweater}} & 
\rotatebox[origin=l]{90}{\parbox{1cm}{\raggedright sweatshirt}} & 
\rotatebox[origin=l]{90}{\parbox{1cm}{\raggedright skirt}} & 
\rotatebox[origin=l]{90}{\parbox{1cm}{\raggedright coat}} & 
\rotatebox[origin=l]{90}{\parbox{1cm}{\raggedright jumpsuit}} & 
\rotatebox[origin=l]{90}{\parbox{1cm}{\raggedright vest}} & 
\rotatebox[origin=l]{90}{\parbox{1cm}{\raggedright cardigan}} & 
\rotatebox[origin=l]{90}{\parbox{1cm}{\raggedright blazer}} & 
\rotatebox[origin=l]{90}{\parbox{1cm}{\raggedright tights- \\ stockings}} \\
\midrule
\textbf{Total} & 38.3 & 21.2 & 11.7 & 5.2 & 5.1 & 3.8 & 3.6 & 3.4 & 3.0 & 1.3 & 1.1 & 0.9 & 0.7 & 0.1 \\
\textbf{Layer} & 51.6 & 0.0 & 0.0 & 33.3 & 97.4 & 82.1 & 81.5 & 0.0 & 100 & 70.0 & 100 & 100 & 100 & 0.0 \\
\textbf{Style} & 7.7 & 0.0 & 0.0 & 0.0 & 86.8 & 39.2 & 55.5 & 0.0 & 59.1 & 10.0 & 62.5 & 28.6 & 60.0 & 0.0 \\
\bottomrule
\end{tabular}
\caption{This table shows the distribution of different clothing categories across the dataset (Total). It also shows how often each clothing type shows up in multi-layered outfits (Layer) or in outfits with style variations (Style). All values are percentages.}
\label{tab:dataset_stats}
\end{table}

\subsection{Dataset Annotation}
\label{sec:annotations}

\textbf{Annotation Pipelines.} Our dataset curation process includes camera calibration, pixel-level segmentation, 3D point cloud reconstruction, and 3D body model annotation. We first calibrate camera intrinsics, distortion, and extrinsics using AprilTags~\cite{multiview_calib1,multiview_calib2}, perform stereo calibration between RGB and depth sensors, refine depth alignment with ColorICP~\cite{park2017colored} into a 3D point cloud, and adjust colors using polynomial correction~\cite{polynomialCC}. We then generate foreground and layered garment masks using a unified two-stage pipeline, employing initializers like YOLOv8~\cite{ultralytics_yolov8} and the semantics-aware Qwen3-VL~\cite{qwen3vl2025}, followed by segmentation with SAM2~\cite{ravi2024sam2}, and manual quality checks. Finally, we annotate body pose and shape using the {SMPL-X}~\cite{pavlakosExpressiveBodyCapture2019} model by detecting 2D keypoints with RTMW-x~\cite{jiang2024rtmw}, triangulating them to 3D using the calibrated camera parameters, and optimizing SMPL-X parameters via a modified SMPLify-X~\cite{pavlakosExpressiveBodyCapture2019} pipeline that incorporates point cloud data for better shape fitting on the minimally clothed scans.

\textbf{Garment Attributes.} Garment sizing charts are standard tools for clothes designers and online shops for assigning labels like S, M, and L to clothes and determine their fit. To the best of our knowledge, there is no public dataset providing detailed sizing charts for clothes. In MV-Fashion, we provide (1) 14 clothes category labels, (2) detailed sizing charts computed with a measuring tape, (3) fitting appearance of the clothes on subjects as categorical labels, (4) draping styles, (5) materials and elasticity, and (6) detailed clothes descriptions computed by large language models.

To standardize the analysis of the sizing charts, we combine the clothes categories into six groups based on the specific body parts measured: \textbf{G1} (shirt-blouse), \textbf{G2} (pants, shorts, tights-stockings), \textbf{G3} (skirt), \textbf{G4} (jacket, sweater, sweatshirt, coat, vest, cardigan, blazer), \textbf{G5} (jumpsuit), and \textbf{G6} (dress). Beyond numerical measurements, we also annotate the garments based on their fitting appearance on the person with three categories: (1) slim with a tight fit, (2) regular, and (3) loose with an oversized look. We enrich our data by categorizing the styles into groups such as rolled-up sleeves, fully/partially closed jackets, and tucked-in shirts. Additionally, we annotate physical properties by manually extracting the garment material (e.g., denim, cotton) and estimating elasticity on a discrete scale ranging from 1 (rigid) to 5 (highly stretchable). Finally, we include textual descriptions for each item, generated by applying Qwen3-VL~\cite{qwen3vl2025} to the frontal and back garment images. 

For examples of the available data and its annotations, please refer to Fig.~\ref{fig:annotations}. For more technical details, discussions, and further examples, see Suppl. Sec.~\ref{supp:data_cu_ann}.

%% file: sec/4_baselines.tex
\section{Baselines}
\label{sec:baselines}

\subsection{Virtual Try-On}
\label{sec:vton_baseline}
Latest image-based VTON approaches \cite{kim2024stableviton, xu2025oot, choi2024idm, chong2025catvton, zhou2025leffa} strongly rely on single-view datasets~\cite{morelliDressCodeHighResolution2022b, choiVITONHDHighResolutionVirtual2021}, limiting progress toward fine-grained controllability and multi-view virtual try-on. MV-Fashion bridges this limitation by providing a rich dataset, as stated in Sec.~\ref{sec:dataset},
enabling conditioning signals far beyond conventional single-view settings.

To assess its applicability, we employ two state-of-the-art VTON approaches, diffusion based IDM-VTON \cite{choi2024idm} and flow matching based InsertAnything \cite{song2025insert}. Specifically, we design three baseline experiments to evaluate both compatibility with existing VTON pipelines and the novel capabilities enabled by our dataset: (1) Single-View, (2) Semantic Controllability, and (3) Multi-View Geometric Analysis.
\textit{Single-View Baselines:} We train both models on a frontal-paired subset replicating well-established single-view settings \cite{choiVITONHDHighResolutionVirtual2021,morelliDressCodeHighResolution2022b}.
\textit{Semantic Controllability:} We extend IDM-VTON by augmenting the input prompt using a text template based on our categorical draping styles (\eg~\textit{base garment is invisible, outerwear is fully closed, stretched sleeves, untucked, hood down}). 
This tests the augmented model's semantic alignment with appearance change instructions at test time by comparing no-style versus styling prompts. 
\textit{Multi-View Geometric Analysis:} We design a two-part geometric benchmark using a subset of synchronized views to quantify cross-view consistency. 
This benchmark utilizes a mixed subset of person views (frontal and rear target poses) for both experiments. 
The first \textit{Cross-View Geometric Test} directly quantifies IDM-VTON limitations under cross-perspective geometric mismatch by requiring it to synthesize the target pose's appearance using only the frontal garment catalogue image, while in the second experiment, we validate our dataset's capacity for \textit{View-Adaptive Try-On}. 
In the latter case, we extend IDM-VTON to utilize both frontal and rear garment catalogue images, passing them to a shared IP-Adapter module \cite{ye2023ipadapter}. 
This design allows the model to select the most geometrically compatible conditioning view for each target pose.

\subsection{Size Estimation} \label{sec:size_baseline}
Inferring accurate garment measurements from images of clothed humans is a challenging and largely unexplored task due to distortions from arbitrary poses, drapes, and wrinkles. To date, most works have focused on automatic garment size estimation from controlled, frontal 2D images, often by detecting keypoints on the normalized item~\cite{kowaleczko2022neural, automatic2022garment}. A more robust alternative is to infer the 2D sewing pattern (\cite{Chen_2024_WACV}), which provides clothing's direct measurements. This is not limited to garment images; clothed humans~\cite{liu2023sewformer, dress1to3} and even text guidance~\cite{bian2024chatgarment} can also be used. Taking all that into account, we leverage our data to predict garment size measurements directly from images of clothed individuals.

Our approach adapts SPnet~\cite{lim2024spnet}, a recent model designed to predict sewing parameters from posed people, in which an encoder-decoder network $\Psi(G^s,P^s,P^t)$ is used to predict the target garment normal image $G^t$ in a canonical pose, where $G^s$, $P^s$ and $P^t$ are image representations for the input garment normal, input body pose, and output canonical pose, respectively. Then, a separate network $\Phi(G^t)$ is trained to estimate the sewing parameters. We refer the reader to~\cite{lim2024spnet} for the original architecture details. Given motion sequences, we obtain $G^s$ with our segmentation and Sapiens~\cite{sapiens}, and $P^s$ from our SMPL-X estimation. Likewise, $G^t$ and $P^t$ are computed from the template recordings. This allows us to train on our real data rather than the synthetic one used in SPnet.

We follow the same architecture and training protocol as SPnet for the network $\Psi$. However, we update $\Phi$ to regress normalized garment sizes, scaled to $[0, 1]$ via dividing by the maximum size in the dataset. We evaluate three variants of $\Phi$: (1) individual models trained per garment group (Sec.~\ref{sec:annotations}), as in SPnet; (2) a single multi-task model trained jointly on all groups; and (3) a multi-task model where the SegNet~\cite{segnet} encoder in (2) is replaced with SwinV2~\cite{liu2021swinv2} and additionally conditioned on $P^t$, to better align garment regions with the corresponding sizing chart entries.

\subsection{Novel View Synthesis (NVS)}
\label{sec:nvs_baseline}
Novel View Synthesis is a critical component of modern virtual try-on and digital fashion applications, enabling the creation of realistic, 360-degree views of clothed individuals. Recent state-of-the-art methods for clothed human reconstruction~\cite{li2024gaussianbodyclothedhumanreconstruction}, garment generation~\cite{rong2024gaussiangarments}, and 3D virtual try-on~\cite{cao2024gsvtoncontrollable3dvirtual} increasingly rely on NVS techniques to handle complex clothing deformations and textures. Therefore, to validate our dataset's suitability for these advanced applications, we benchmark its performance using popular NVS methods. We select three models implemented within the Nerfstudio framework~\cite{nerfstudio}: Instant-NGP~\cite{instantngp}, Nerfacto, and 3D Gaussian Splatting (splatfacto)~\cite{3dgaussiansplatting}. 

Our investigation is two-fold. First, we conduct a comparative analysis of these three methods on MV-Fashion to establish baseline NVS performance. Then, we perform an ablation study on the number of input views to measure the effect of view density on reconstruction quality.

For more details on the baselines, see Suppl. Sec.~\ref{supp:baseline_tech_details}.

%% file: sec/5_experiments.tex
\section{Experiments}
\label{sec:experiments}
In this section, we outline the data, training protocols, and evaluation metrics, as well as discuss the baseline results.

\subsection{Virtual Try-On} \label{sec:VTON_experiments}
\textbf{Data and Training Protocols.} We utilize a subject-based split for all VTON experiments, reserving $10\%$ of subjects as the test set. For the Frontal Single-View Subset, we sample i.i.d. from sequences captured across seven distinct frontal-facing camera angles to ensure view diversity. This subset totals 15,803 training images and 2,188 testing images. The Multi-View Subset incorporates images from seven additional rear-facing cameras for cross-perspective challenges, totaling 14,689 training images and 2,073 testing images. Experiments rely on implementations of IDM-VTON~\cite{choi2024idm} and InsertAnything~\cite{song2025insert}, using their standard training protocols with modifications described in Sec.~\ref{sec:vton_baseline}.

\textbf{Evaluation Metrics.}
We report three standard VTON metrics: SSIM~\cite{ssim}, LPIPS~\cite{lpips}, and FID \cite{fid}.

\begin{figure}[t]
    \centering
    \includegraphics[width=0.45\textwidth]{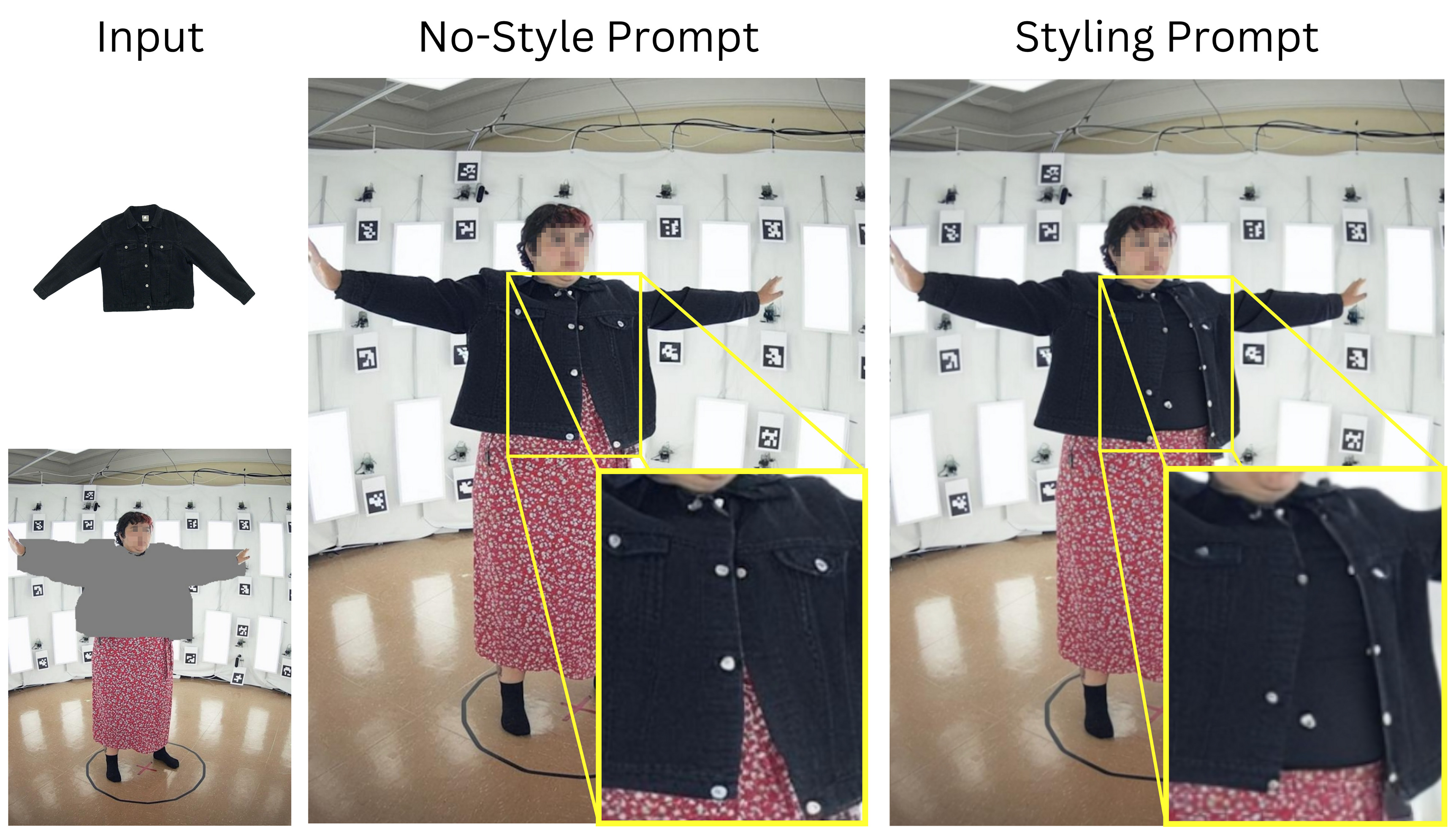}
    \caption{Qualitative comparison showing models trained with styling-augmented data respond to fine-grained styling prompts (\textit{a jacket, the outer wear is fully open}), producing outputs that are sometimes distinguishable from no-style prompts (\textit{a jacket}).
    }
    \label{fig:vton_styling_qualitative}
\end{figure}

\textbf{Results.}
Quantitative results for the \textit{Single-View} Baseline (Tab.~\ref{tab:vton_benchmark}) show that the frontal-paired subset of MV-Fashion can be directly used with existing VTON pipelines. Both IDM-VTON and InsertAnything achieve high-fidelity scores on all metrics. 
The capacity to control try-on results based on styling annotations (\textit{Semantic Controllability}) presents a potential further research directions. As illustrated in Fig.~\ref{fig:vton_styling_qualitative}, IDM-VTON trained with augmented prompt data demonstrates the ability to react to the specific instruction within the styling augmented prompt, a capability absent in the model pre-trained on the VITON-HD dataset. However, we found that the model often struggles to follow these additional semantic signals outside its primary guidance (\eg, from IP-Adapter). Even in successful control cases, subtle artifacts remain present, demonstrating that achieving fine-grained styling control is currently a highly challenging task. 

\begin{figure}[t]
    \centering
    \includegraphics[width=0.45\textwidth]{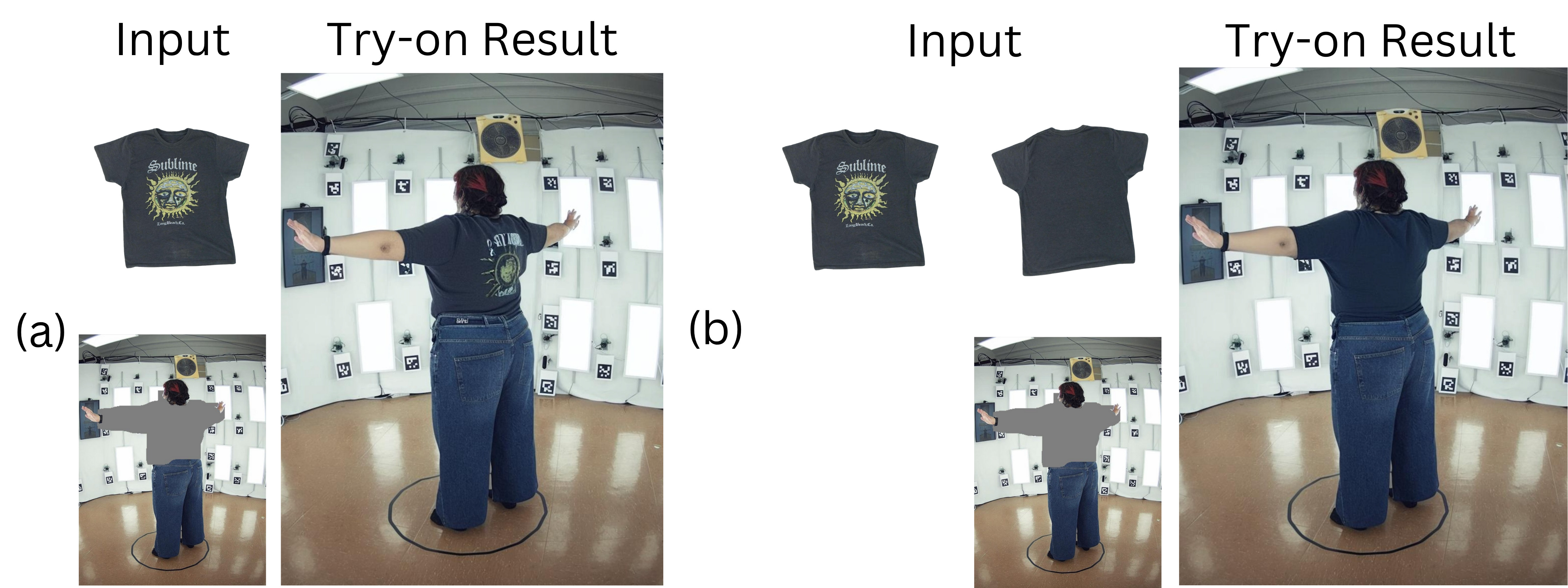}
    \caption{Results on (a) \textit{Cross-View Geometric Test} vs (b) \textit{View-Adaptive Try-On}. The updated IDM-VTON architecture can map between the catalogue view and the person's pose when both front and rear images of the garment are provided to the model.}
    \label{fig:vton_mv_qualitative}
\end{figure}

For Multi-View Geometric Analysis, the \textit{Cross-View Geometric Test} reveals a performance drop across all metrics (Tab.~\ref{tab:vton_benchmark}), directly quantifying the difficulty models face in maintaining geometric consistency under cross-perspective constraints. The \textit{View-Adaptive Try-On}, conversely, shows a partial reduction in failure cases when models adaptively use both frontal and rear garment views (Fig.~\ref{fig:vton_mv_qualitative}), validating the research potential enabled by these multi-view pairs.

Since new architectural design is outside our scope, the development of advanced control mechanisms and multi-view models is left as a valuable future investigation.

\begin{table}[ht]
\centering
\scriptsize
\setlength{\tabcolsep}{4pt}
\label{tab:combined_vton_res}
\begin{tabular}{lcccc}
\toprule
\textbf{Benchmark} & \textbf{Experiment} & \textbf{SSIM$\uparrow$} & \textbf{LPIPS$\downarrow$} & \textbf{FID$\downarrow$} \\
\midrule
\multirow{2}{*}{Single-View} & IDM-VTON~\cite{choi2024idm} & $0.881$ & $0.086$ & $10.187$ \\
 & InsertAnything~\cite{song2025insert} & $0.927$ & $0.065$ & $8.192$ \\
\midrule
\multirow{2}{*}{Multi-View (IDM-VTON)} & Cross-view & $0.868$ & $0.098$ & $12.907$ \\
 & View-adaptive & $0.873$ & $0.093$ & $12.775$ \\
\bottomrule
\end{tabular}
\caption{Quantitative results benchmarking Single-View Baselines and Multi-View Geometric Analysis using the proposed MV-Fashion's subsets described in Sec.~\ref{sec:vton_baseline}.}
\label{tab:vton_benchmark}
\end{table}

\subsection{Size Estimation} \label{sec:sizing_exp}

\textbf{Data and Training Protocols.} 
We train $\Psi$ and $\Phi$ independently, exclusively utilizing data from outer-layer garments to ensure full visibility, and without applying any data augmentation. For $\Psi$, we sample 30 uniformly distributed frames from five random frontal cameras per motion sequence to capture diverse poses. Each source tuple ($P^s, G^s$) is paired with the canonical configuration ($P^t, G^t$), derived from the initial frame of template sequences for the two most frontal views, yielding 300 training pairs for every outer-layer garment present in a sequence. For $\Phi$, we use the 1,508 canonical $G^t$ images paired directly with our annotated measurements. All versions of $\Phi$ are trained 5 times and the results are averaged to avoid randomness effect. The test split follows Sec.~\ref{sec:VTON_experiments}, ensuring strict separation between train and test subjects.

\textbf{Evaluation Metrics.} We report two metrics for $\Psi$: MAE to measure pixel-level accuracy, and SSIM for perceptual quality. For the size predictor ($\Phi$), we also provide MAE, in centimeters, along with its standard deviation.

\begin{figure}
    \centering
    \includegraphics[width=0.45\textwidth]{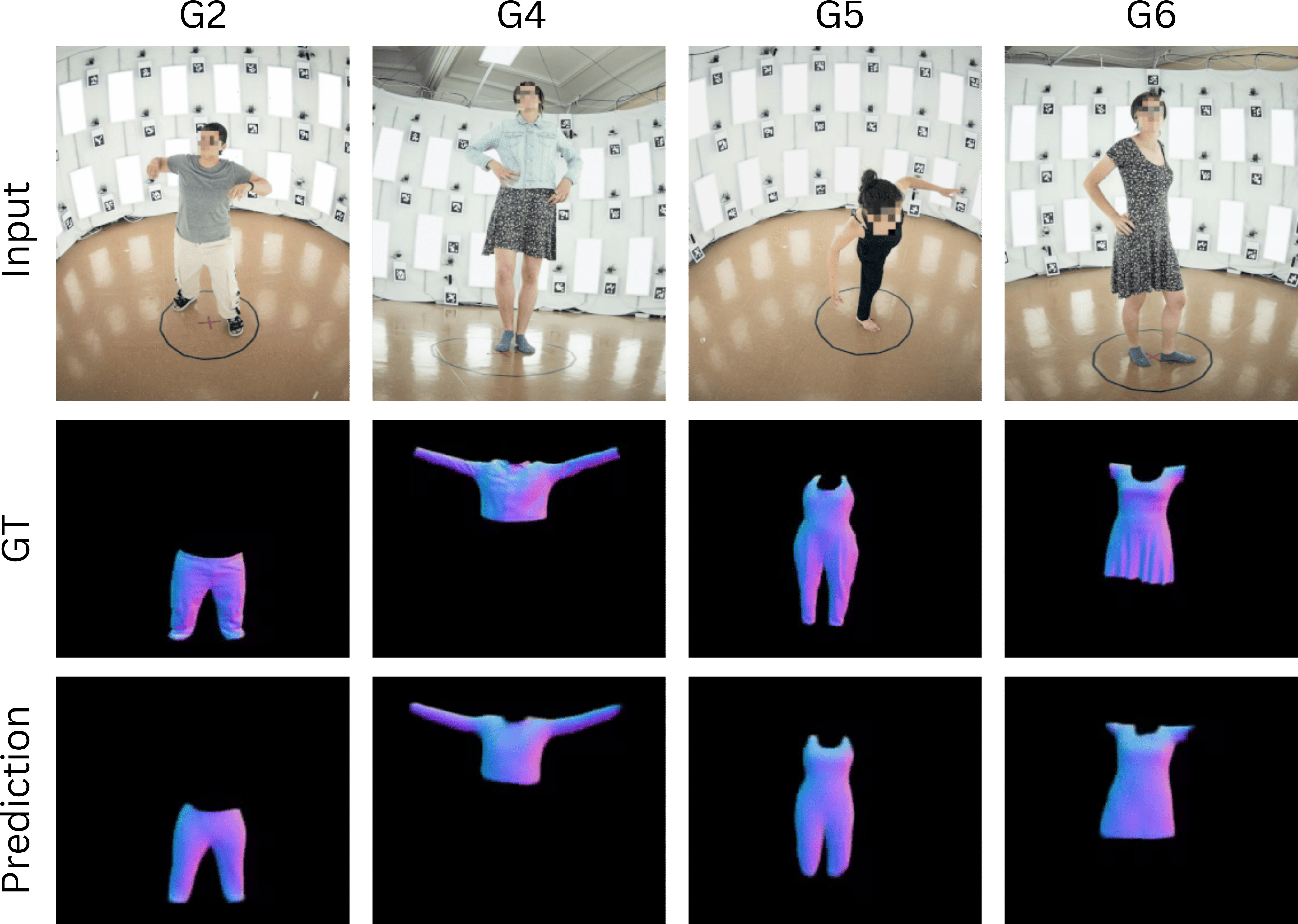}
    \caption{Qualitative results of the canonical garment normal predictor $\Psi$ for groups G2, G4, G5 and G6; showing input frame, ground truth ($G^t$), and the predicted normal image.}
    \label{fig:sizing_qualitative}
\end{figure}

\textbf{Results and Analysis.} Qualitative results for $\Psi$ (Fig.~\ref{fig:sizing_qualitative}) demonstrate effective disentanglement of pose from garment geometry, producing canonical, smoothed normals that recover the garment's intrinsic shape even in challenging views. Quantitatively, $\Psi$ achieves a global MAE of 0.0163 and SSIM of 0.9355, indicating high fidelity and structural preservation. For $\Phi$, Tab.~\ref{tab:sizing_metrics} validates the effectiveness of our multi-task approach (exploiting common measurements across garment groups) compared to the per-group training originally proposed in SPnet. The SwinV2-based model with $P^t$ guidance yields the most robust performance, achieving the lowest MAE (4.279 cm). These baselines demonstrate that MV-Fashion contains sufficient signal to learn accurate sizing from real-world images.

\begin{table}[h]
\centering
\scriptsize
\setlength{\tabcolsep}{4pt}
\begin{tabular}{l c c c}
\toprule
\textbf{Group} & \textbf{Per-Group} & \textbf{Multi-Task} & \textbf{Multi-Task + SwinV2} \\
\midrule
\textbf{G1} & $4.069$ $(5.861)$ & $4.361$ $(6.061)$ & $3.933$ $(5.994)$ \\
\textbf{G2} & $3.006$ $(2.541)$ & $3.185$ $(2.619)$ & $3.067$ $(2.762)$ \\
\textbf{G3} & $9.951$ $(8.437)$ & $6.847$ $(9.174)$ & $6.643$ $(10.135)$ \\
\textbf{G4} & $5.092$ $(7.377)$ & $5.644$ $(8.563)$ & $4.436$ $(6.537)$ \\
\textbf{G5} & $12.109$ $(10.313)$ & $4.295$ $(2.746)$ & $4.889$ $(4.633)$ \\
\textbf{G6} & $7.767$ $(7.954)$ & $6.549$ $(7.234)$ & $6.389$ $(6.562)$ \\
\midrule
\textbf{Average} & $4.904$ $(6.533)$ & $4.710$ $(6.392)$ & $4.279$ $(5.870)$ \\
\bottomrule
\end{tabular}
\caption{MAE results (cm) for the size predictor $\Phi$, reported as mean (std). We compare the three variants from Sec.~\ref{sec:size_baseline} across all garment groups (defined in Sec.~\ref{sec:annotations}) and the average.}
\label{tab:sizing_metrics}
\end{table}

\subsection{Novel View Synthesis Benchmarking}
\label{sec:nvs_exp}

\begin{table}[t]
    \centering
    \scriptsize
    \begin{tabular}{lccc}
    \toprule
    \textbf{Method} & \textbf{PSNR $\uparrow$} & \textbf{SSIM $\uparrow$}& \textbf{LPIPS $\downarrow$} \\
    \midrule
    
    instant-ngp~\cite{instantngp} & $29.120$ $(1.555)$ & $0.936$ $(0.015)$ & $0.089$ $(0.022)$ \\
    nerfacto~\cite{nerfstudio} & $26.535$ $(2.047)$ & $0.952$ $(0.011)$ & $0.052$ $(0.009)$ \\
    splatfacto~\cite{3dgaussiansplatting} & & & \\
    \quad 4 views & $24.633$ $(2.230)$ & $0.933$ $(0.015)$ & $0.074$ $(0.019)$ \\
    \quad 8 views & $27.299$ $(1.850)$ & $0.944$ $(0.013)$ & $0.050$ $(0.012)$ \\
    \quad 16 views& $28.500$ $(1.709)$ & $0.949$ $(0.013)$ & $0.039$ $(0.008)$ \\
    \quad 32 views& $29.211$ $(1.764)$ & $0.954$ $(0.013)$ & $0.032$ $(0.007)$ \\
    \quad 56 views& $29.567$ $(1.877)$ & $0.957$ $(0.013)$ & $0.028$ $(0.006)$ \\
     
    \bottomrule
    \end{tabular}
    \caption{Comparison of image quality metrics as mean (std). We compare against other methods and provide an ablation study on the number of views for splatfacto.}
    \label{tab:nvs_metrics}
\end{table}

\begin{figure}[t]
    \centering
    \includegraphics[width=0.45\textwidth]{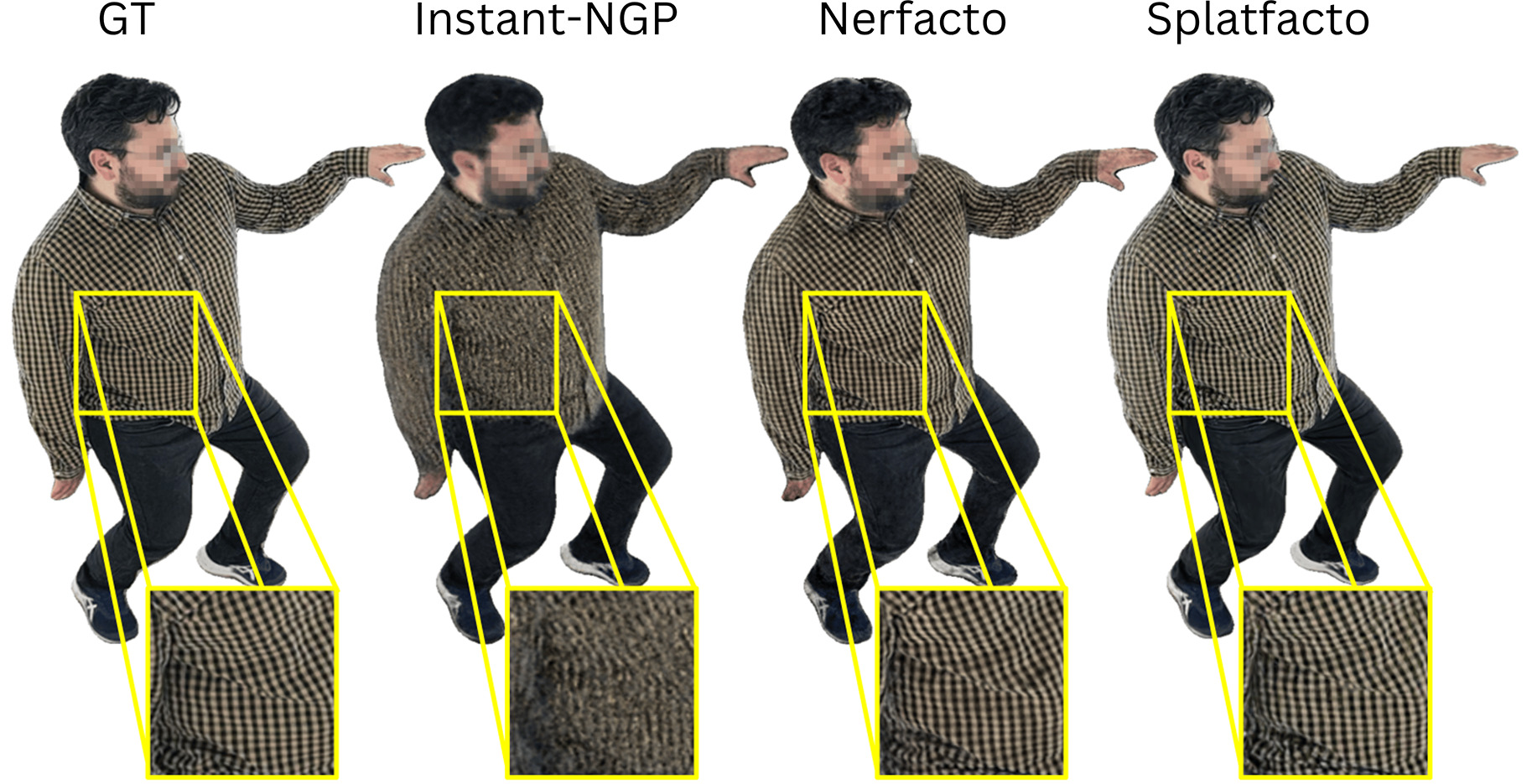}
    \caption{Qualitative examples of the presented NVS methods. It shows how Nerf-based methods struggle with fine details, but Gaussian Splatting can reconstruct fine textures and the wrinkles.}
    \label{fig:nvs_qualitative}
\end{figure}

\textbf{Data and Training Protocols.} We sample one random static frame from all subjects to capture data diversity. All methods are trained and evaluated independently on these scenes, with their average as the reported result. For compatibility across methods, we only use the 60 main cameras, excluding the 4K cameras. Our primary experiment uses a 56/4 train/test split, with test views selected from challenging viewpoints (e.g. top-side angle). A second experiment ablates the number of training views (e.g., 56, 32, 16, 8, 4) to evaluate the impact of view density. To prevent a proximity bias between the train and test views, the training split is randomized for each scene.

\textbf{Evaluation Metrics.} We report three standard NVS metrics: PSNR, SSIM~\cite{ssim}, and LPIPS~\cite{lpips}.

\textbf{Results and Analysis.} Quantitative results (Tab.~\ref{tab:nvs_metrics}) confirm that all benchmarked methods achieve high-fidelity reconstructions, validating our dataset's suitability for NVS research. Qualitative results (Fig.~\ref{fig:nvs_qualitative}) further show that the methods can reproduce intricate garment details such as fine textures and wrinkles. Our view-density ablation (Tab.~\ref{tab:nvs_metrics}) with the splatfacto method shows that all metrics improve with more training views, but performance saturates at 56 views. This suggests that adding more cameras to the system would only provide diminishing returns, demonstrating that our multi-view setup is appropriate for this task.

The results collectively validate MV-Fashion as a robust testbed for photometric, geometric, and semantic consistency across fashion-centric tasks. For more baseline results and discussions, see Suppl. Sec.~\ref{supp:experiments}.

%% file: sec/6_conclusion.tex
\section{Conclusion}
\label{sec:conclusion}
We introduced MV-Fashion, a comprehensive multi-view video dataset designed specifically for fashion-centric applications. The dataset offers extensive real-world garment diversity, rich annotations, and paired multi-view captures that support key tasks such as Virtual Try-On, size estimation, and novel view synthesis. We also established benchmark evaluations using state-of-the-art methods to provide a foundation for future research. To promote reproducibility and community-driven progress, both the MV-Fashion dataset and baseline implementations will be publicly released upon publication. We believe MV-Fashion will serve as a valuable resource and catalyst for advancing research in fashion understanding and digital human modeling.

\textbf{Limitations and Future Works.}
We provide several types of annotations and the data used for our baseline models. Most annotations are generated using off-the-shelf, well-established methods complemented by targeted manual QA; we therefore focus our evaluation on downstream task performance rather than exhaustively benchmarking every intermediate annotation, and encourage future work to further analyze and improve these components. We report baselines for the most common downstream tasks, but the dataset is not limited to these. It can also support tasks such as instance segmentation or garment classification. 

\section{Acknowledgements}
This work has been partially supported by the Spanish projects TED2021-131317B-I00, PID2022-136436NB-I00 and PDC2022-133305-I00, Catalan project IDC-2024-PROD-00013, and by ICREA under the ICREA Academia programme. This work has been supported by the Spanish grants PID2020-120611RB-I00 and PID2024-162984NB-I00 (funded by ERDF/ UE and MICIU/AEI/10.13039/501100011033).

The project’s diverse responsibilities were distributed as follows: Hunor Laczko and Meysam Madadi handled hardware design and setup, while Pavel Aguilar served as the operator; Hunor Laczko also managed scanning software, data preprocessing, and novel view synthesis, whereas Libang Jia focused on data curation and segmentation; Meysam Madadi further oversaw calibration, camera fusion, and body pose and shape, complemented by Phat Truong on virtual try-on, Diego Hernandez on size estimation, and guidance from supervisors Meysam Madadi, Sergio Escalera, and Jordi Gonzalez.

%% file: sec/X_suppl_arxiv.tex
\clearpage
\maketitlesupplementary

\begin{figure*}[!h]
    \centering
    \includegraphics[width=\linewidth]{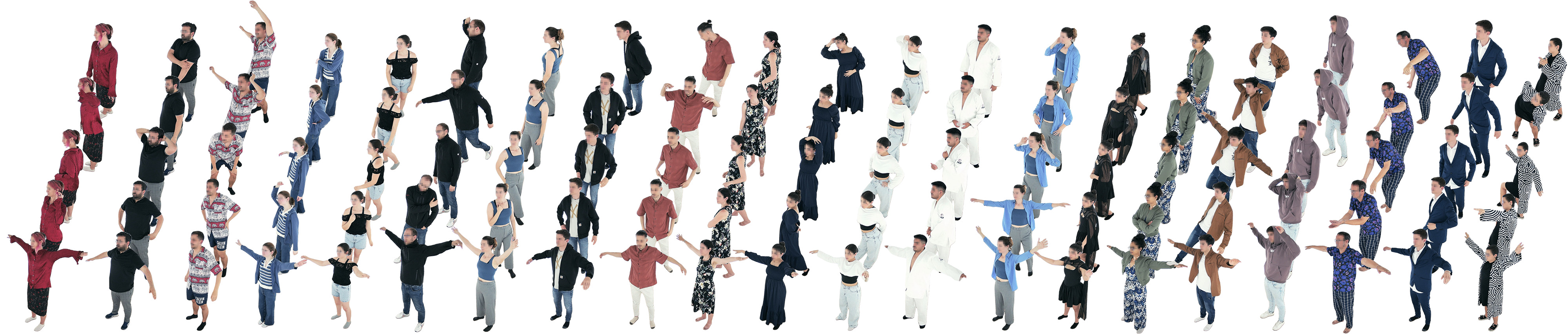}
    \caption{Representative samples from MV-Fashion illustrating the range of garment typologies (single vs. multi-layer) and poses captured in the dataset.}
    \label{fig:supp_extra_samples}
\end{figure*}

In this supplementary document we describe the capture setup and dataset annotation processes in more detail and provide additional statistics. Later, we detail the benchmark setups and provide additional quantitative and qualitative results.

\section{Dataset Acquisition} \label{supp:data_acq}
\subsection{Setup}

As described in the main paper, we use 60 Raspberry Pi (RPi) global shutter cameras (1.6 MP), and 8 Orbbec Femto Bolt cameras. We use global shutter cameras rather than rolling shutter ones, avoiding motion artifacts and making the data more consistent. The Bolt cameras are Time-of-Flight (ToF) devices comparable to the Azure Kinect which record 4K color footage and lower resolution (640x576) depth information. All the cameras are connected to the same external trigger signal that is generated by a Raspberry Pi Pico module. It is a PWM signal that triggers the exposure. To create a single depth frame, the Bolt cameras take several infrared (IR) captures. In order to avoid interference between these, there is a sub-millisecond delay between each Bolt camera. This results in an interleaved pattern for the IR captures, where only one Bolt camera is capturing at a time. This also means that there is a higher chance of motion blur in the captured depth images, as the exposure times are fixed. Most consumer depth cameras operate in this manner, and this cannot be avoided. The overall synchronization is verified using a fast flashing LED strip. The LED strip uses individually addressable LEDs and is connected to an Arduino microcontroller. This setup can achieve sub-millisecond flashes of unique patterns that can be later identified in each frame and cross-referenced. Our setup achieves sub-2-millisecond synchronization across cameras. The lighting is controlled by 40 LED panels. These provide uniform and diffused light, with a high color rendering index (CRI) rate. Using panels help with distributing the power load, thus minimizing the electromagnetic interference. They are also flicker-free, not affecting the exposure. 

All RPis are connected through ethernet cables and switches to a central computer, and all recordings are saved directly to this computer over the network. The Bolt cameras are connected to a mini computer each. These cameras are USB cameras, and connecting them to the same computer can cause instabilities, even when using dedicated USB expansion cards. That is why we opted to connect each to a mini computer (Asus NUC) first. They still record directly to the central unit over the ethernet. 

We developed a custom application that allows us to control all cameras and record the sequences in a structured manner.

\subsection{Collection Protocol} \label{supp:collection_protocol}
We have three main types of sequences: \textit{Body} sequences, which record the subjects in minimal clothing to capture the body shapes accurately; \textit{Template} sequences use fixed poses and aim to capture each piece of garment separately in these predefined poses; \textit{Motion} sequences that record the dynamic clothing following random poses. We also record the individual catalogue/flat images of each garment with a top-down view camera. This top camera is mounted in the center of the capture setup, at the top. We use a temporary white background and lay the garments flat. Each flat garment image is recorded in two styles. The first variation captures the garment fully flattened to reveal its complete structure. The second presents a compact, less structured arrangement, designed to mimic real-world images taken by users.

As auxiliary data, each time before a subject re-enters the capturing area, we record an empty scene for the extrinsics calibration purposes to ensure consistency across sessions. We use this information to recalibrate the cameras to avoid any potential error due to the movement of the cameras. At the beginning of each day, we also performed manual checks and removed any dust that had accumulated during the previous day.  

A typical recording session lasted 2 hours. Each subject could participate in up to three different sessions. We explained the protocol to the subjects, described their task and asked them to sign all necessary documents. During each recording, the subjects received live instructions shown on a monitor inside the capturing setup as well as a live preview of themselves. Specifically, we showed 5 random fashion images of different poses that we asked them to imitate to the best of their abilities (see Fig.~\ref{fig:supp_extra_samples}). These images were collected from the internet, and they portrayed full-body images of fashion-oriented standing poses, similar to the ones found in many online clothing retail websites. Since we observed different trends in women's and men's fashion, we grouped the gathered images into two categories. One focused on women's style, while the other focused on men's style. Each participant was allowed to choose whichever category they preferred. The subjects were instructed to bring a wide variety of clothing styles (sports, home, formal, etc), including those with multiple layers whenever possible. In these multilayer cases, we first recorded a few sequences with only one layer, followed by additional clothing layers in further sequences. For each of these, we recorded new template sequences, as well as whenever the style of the clothing item changed (eg. the coat is open or closed). At the end of the session, the subjects were requested to review the recordings and confirm with an additional form.

\section{Dataset Curation and Annotation} 
\label{supp:data_cu_ann}

\subsection{Quality Assurance}
To ensure the reliability of the captured data, we apply a data-cleaning and validation pipeline combining automated procedures with manual inspection. The process consists of two main components:

\begin{itemize}[nosep,leftmargin=*]
    \item Performing frame-count and timestamp alignment to enforce cross-camera and cross-modal consistency;
    \item Conducting a light quality check to filter out improperly exposed frames.
\end{itemize}

\textbf{Camera Alignment.}
Each camera records at a nominal rate of 15\,fps for about 20\,seconds, producing approximately 300 frames per sequence. Minor variations in frame counts may occur due to startup latency or early termination. To obtain a clean and temporally consistent subset, we apply a light-weight alignment and cleaning procedure: we match RGB and depth frames using timestamp proximity, remove occasional extra or invalid frames, and standardize all camera streams to a unified timeline. This yields a synchronized multi-camera set with consistent views and modalities across the entire sequence.

\begin{figure}[h]
    \centering
    \includegraphics[width=\linewidth]{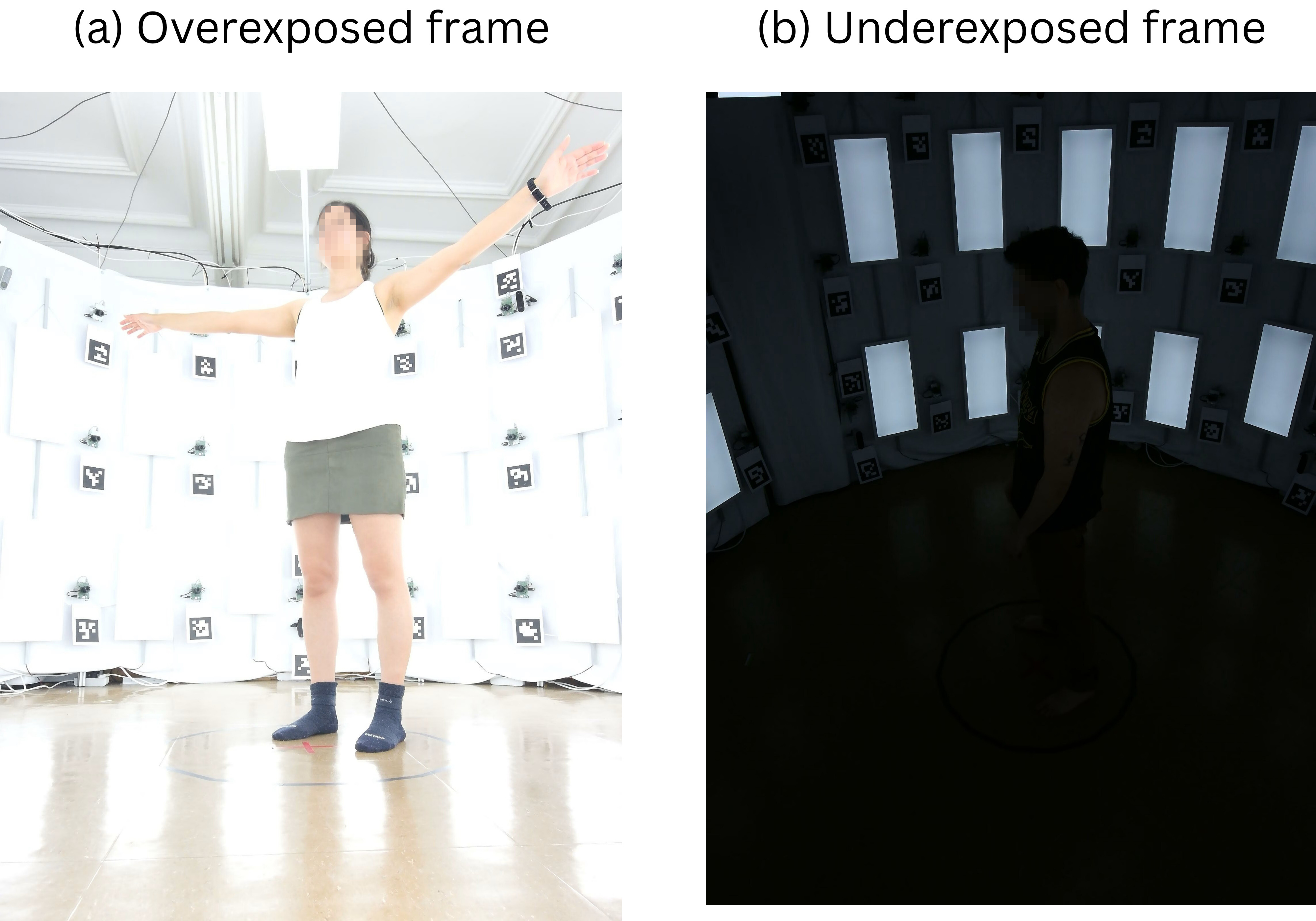}
    \caption{Examples of exposure anomalies:
    (a) An overexposed frame where strong illumination causes saturation and severe loss of texture.
    (b) An underexposed frame where insufficient lighting leads to a near-black appearance and loss of structural details.}
    \label{fig:exposure_anomalies}
\end{figure}

\textbf{Exposure Anomaly Detection.}
To avoid occasional exposure issues from affecting the final data quality, we perform a lightweight exposure check (see Fig.~\ref{fig:exposure_anomalies}). When a recording contains improperly exposed segments or incomplete captures, we discard the affected take or segment.

\begin{figure}[h]
\centering
\begin{subfigure}{0.3\linewidth}
\includegraphics[width=\textwidth]{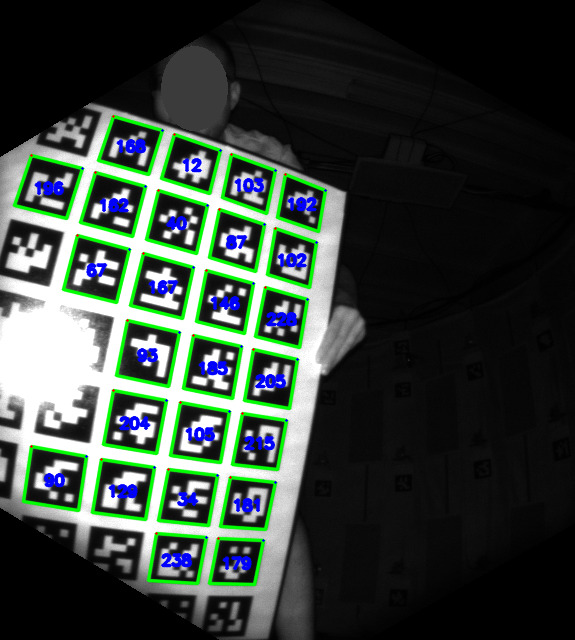}
\caption{Bolt IR.}
\label{fig:first}
\end{subfigure}
\hfill
\begin{subfigure}{0.3\linewidth}
\includegraphics[width=\textwidth]{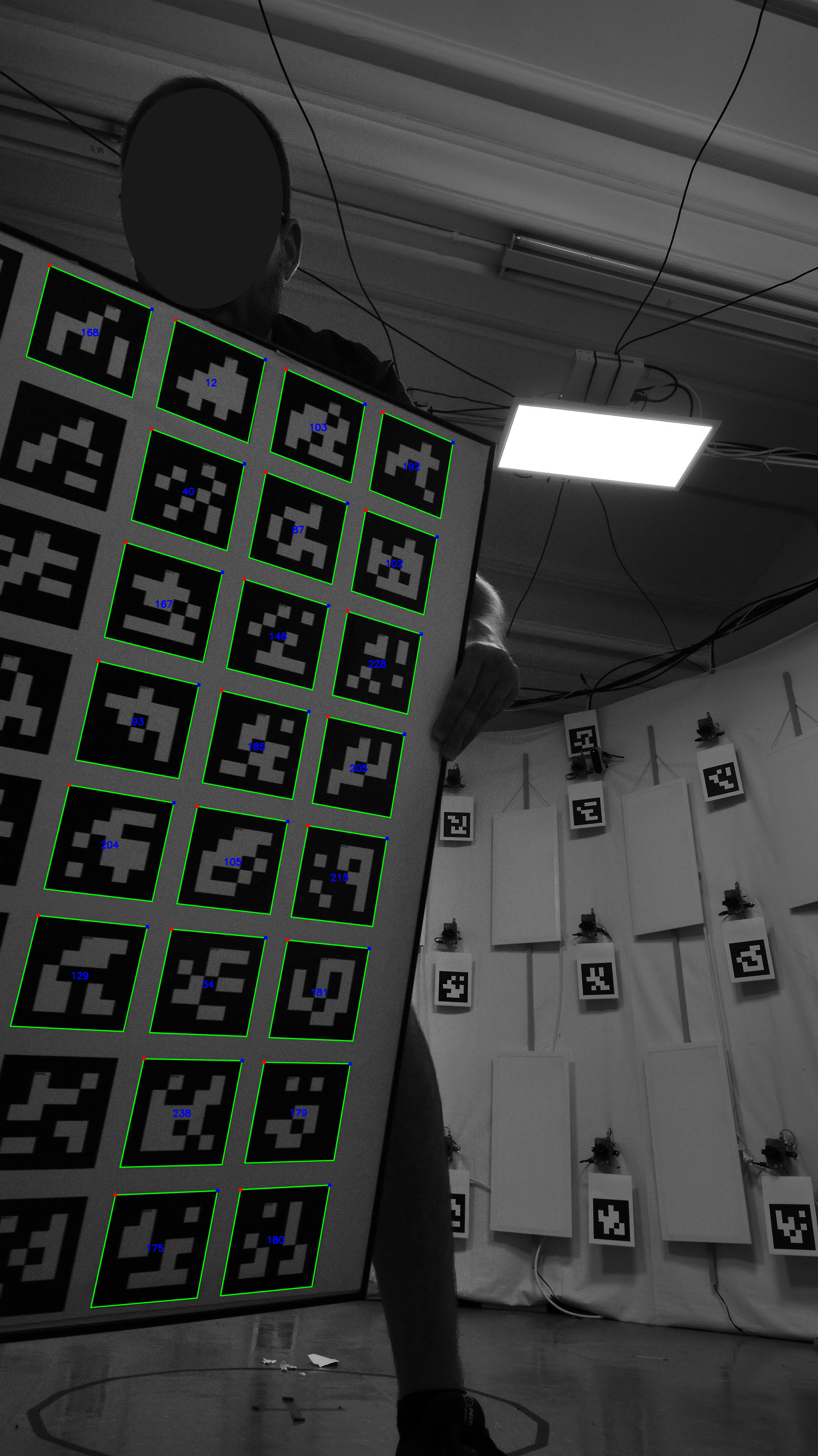}
\caption{Bolt RGB.}
\label{fig:second}
\end{subfigure}
\hfill
\begin{subfigure}{0.3\linewidth}
\includegraphics[width=\textwidth]{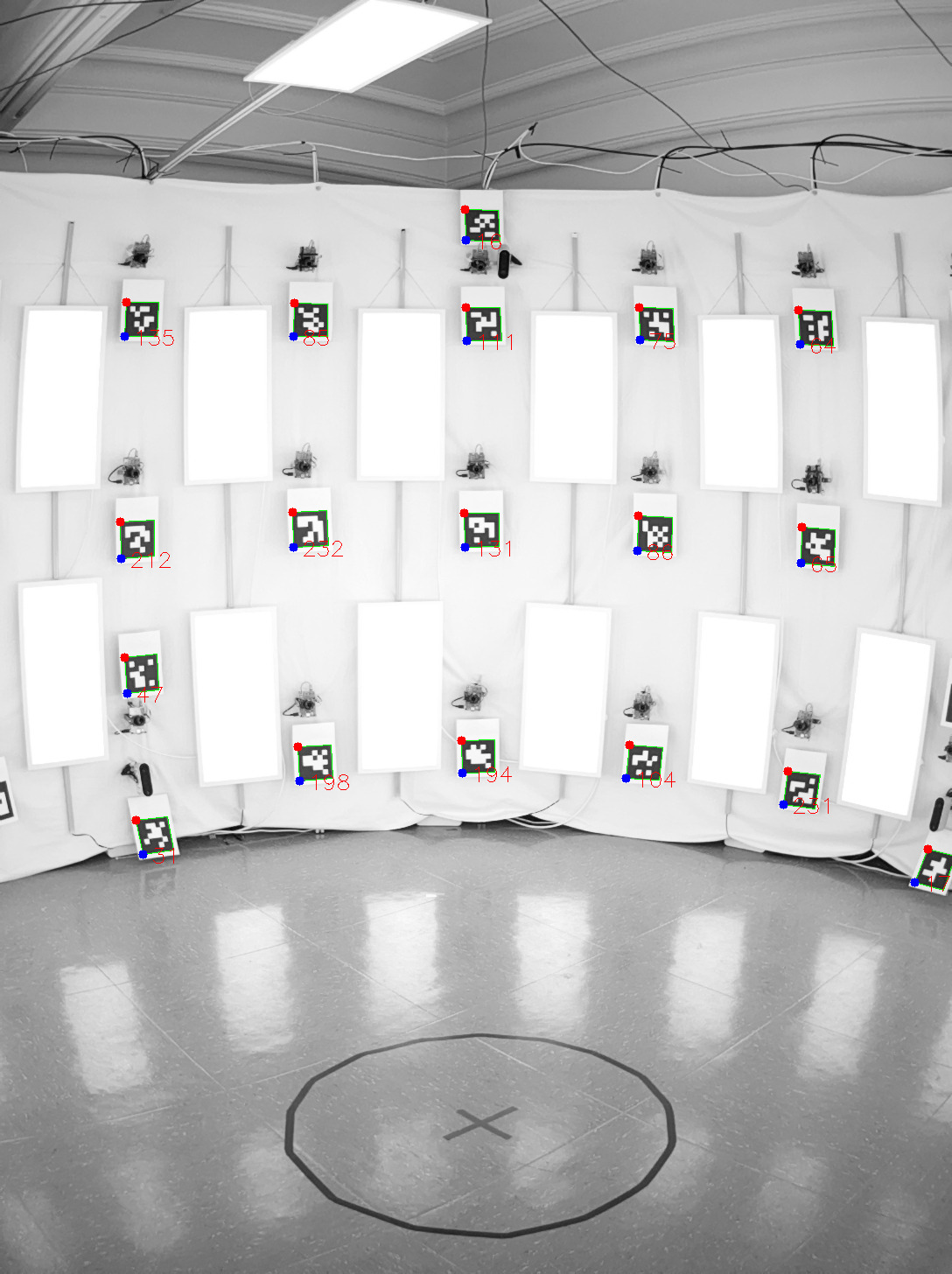}
\caption{RPi.}
\label{fig:third}
\end{subfigure}
\caption{Examples of images and detected keypoints (in green rectangles) used in the intrinsics (a and b) and extrinsics (c) calibration.}
\label{fig:calibration}
\end{figure}

\subsection{Camera Calibration} 

\begin{figure}[h]
    \centering
    \includegraphics[width=\linewidth]{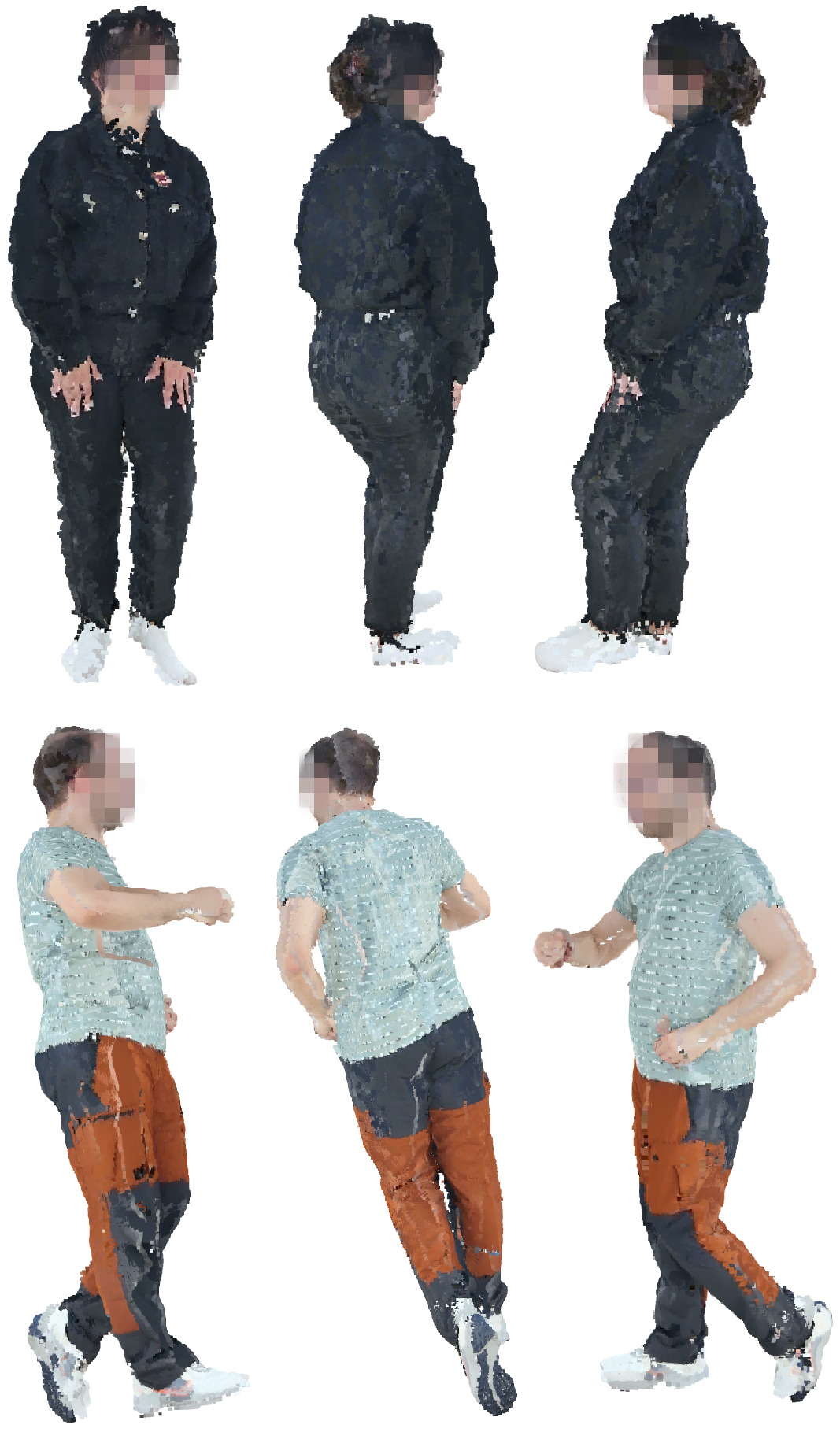}
    \caption{Examples of the reconstructed point clouds visualized from different viewpoints. We apply \textbf{no post-processing} on the point clouds, including 3D and color smoothing. We preprocess the depth images to filter the boundary noise by masking out the pixels that have an absolute difference above 15\,mm with their nearby pixels.}
    \label{fig:pointcloud}
\end{figure}

We calibrate the intrinsics and eight distortion parameters with a matte printed AprilTag board using standard OpenCV functions. AprilTag has several advantages over checkerboard in precise multi-camera calibration, specifically due to the unique coding of feature points, the calibration is more robust when corners are partially visible. We repeat the intrinsics calibration three times: an initial pass to detect outlier keypoints, a refined second pass with outliers removed, consolidating the calibration. Finally, we correct misdetected keypoints in the undistorted images by intersecting the AprilTag edge lines at expected corner locations before performing the final calibration round. For the depth camera calibration, we adjust the IR image contrast to improve landmark detection. We then apply a stereo calibration between the depth and its corresponding RGB cameras. As a result, we achieve an average reprojection error of $0.4$, $0.4$ and $1.7$ pixels for the RPi, Bolt IR and RGB cameras, respectively. For the extrinsics calibration among RGB cameras, including RPis and Bolts, we use AprilTag markers installed beside each camera using A5 flat boards. We apply open toolboxes \cite{multiview_calib1,multiview_calib2} for this task and achieve an average reprojection error of $0.3$. See Fig.~\ref{fig:calibration} for examples of AprilTag boards and detected keypoints used for calibration. Finally, for better alignment of the depth cameras and 3D fusion of 2.5D point clouds, we run ColorICP~\cite{park2017colored} between the overlapping depth cameras. See Fig.~\ref{fig:pointcloud} for examples of the reconstructed point clouds.

\begin{figure}[h]
    \centering
    \includegraphics[width=\linewidth]{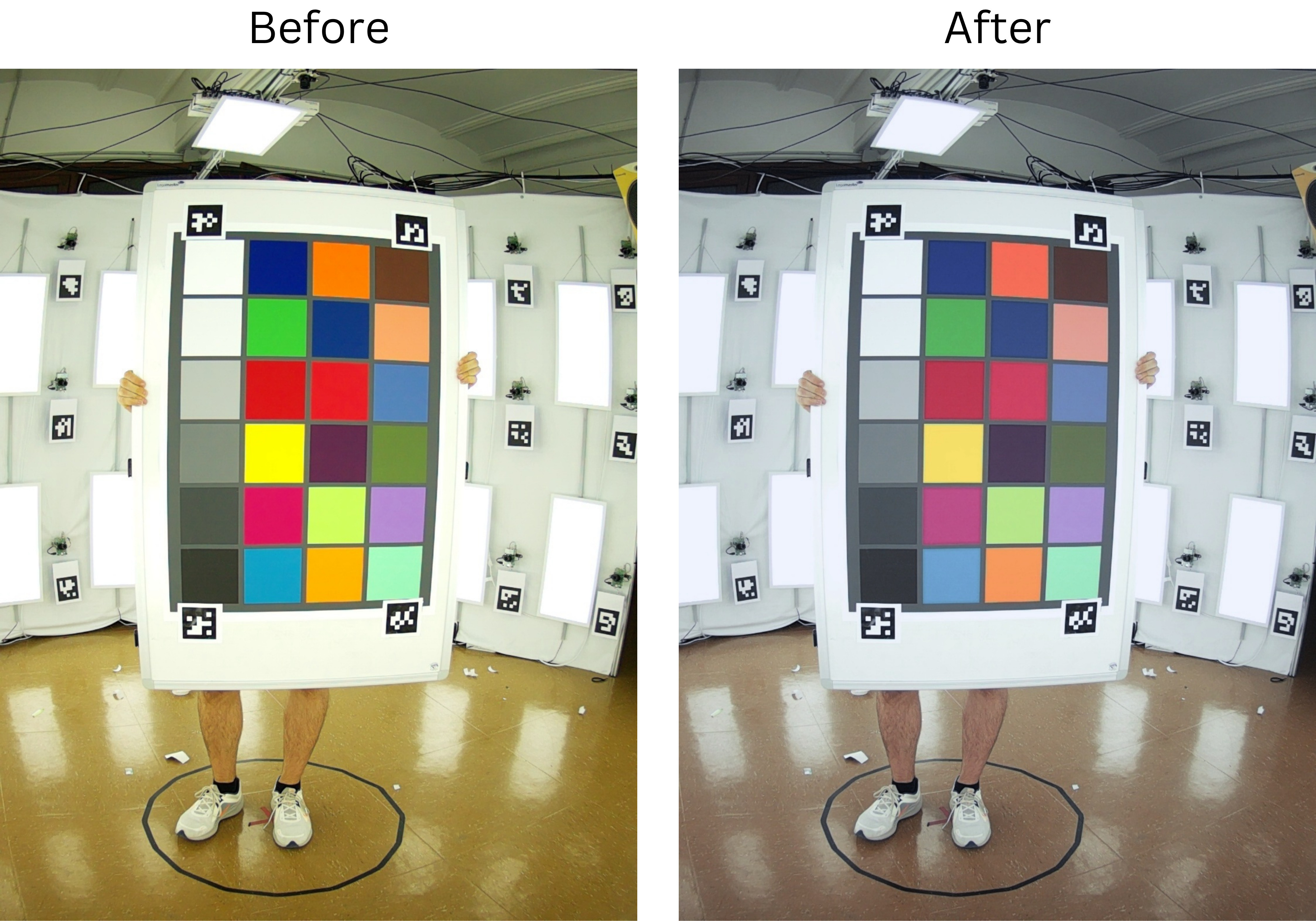}
    \caption{A sample image of the color calibration target before and after the color adjustments.}
    \label{fig:supp_color_calib}
\end{figure}

\textbf{Color Calibration.} We perform standard polynomial color correction~\cite{polynomialCC} between all cameras. We create a custom color calibration target as seen in Fig.~\ref{fig:supp_color_calib}. While we aim for a faithful reproduction of the colors, we cannot guarantee the exact color reproduction due to printing inaccuracies. Nonetheless, our main goal is to have consistent color profiles across each camera, which we can achieve with this method.

\subsection{Segmentation}
\label{sec:suppl_segmentation}

\begin{figure}[t]
    \centering
    \includegraphics[width=\linewidth]{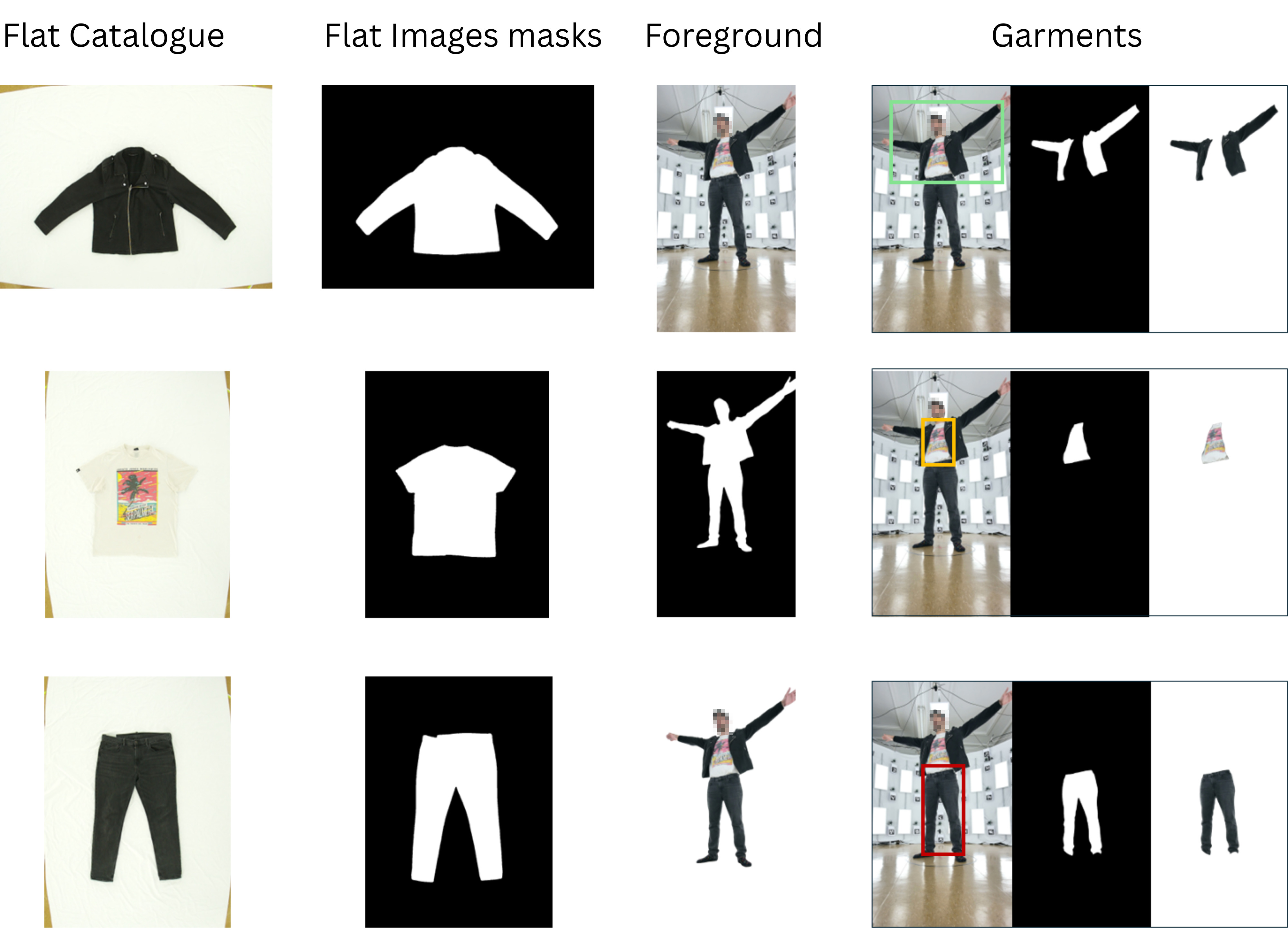}
    \caption{
        Representative segmentation outputs produced by our pipeline.
        From left to right: flat-catalogue garment images, corresponding 
        flat garment masks, foreground masks, and layered on-body 
        garment segmentation.
    }
    \label{fig:segmentation_overview}
\end{figure}

\textbf{Segmentation Pipeline Details.}
As described in the main paper, we provide human foreground masks and 
layered garment masks for each frame. Representative examples of these 
segmentation outputs are shown in Fig.~\ref{fig:segmentation_overview}.

We adopt a unified prompt-driven two-stage segmentation pipeline: a task-appropriate initializer first produces a coarse guidance (e.g., via a text prompt or a bounding box), and then SAM2~\cite{ravi2024sam2} performs pixel-level segmentation. We instantiate this pipeline in three settings, with full implementation details as follows:

\textbf{Flat Catalogue Segmentation.} For static flat garment images with clean backgrounds and a single semantic category, we develop a batch-processing segmentation tool based on the open-source Lang-Segment-Anything framework~\cite{medeiros2024lang}. Specifically, we apply GroundingDINO~\cite{liu2025grounding} to each RGB image using the fixed text prompt ``clothes'' (box threshold = 0.3, text threshold = 0.25) to detect garment regions. We retain the highest-confidence bounding box and use it as a prompt for SAM2.1-Hiera-Small to obtain a refined pixel-level garment mask. The SAM2~\cite{ravi2024sam2} probability map is then binarized using a threshold of 0.5 and, when necessary, resized to match the original image resolution. For each image, we generate two outputs: (i) an RGBA foreground image, where the alpha channel encodes the predicted garment mask, and (ii) a 0/255 binary mask PNG. All outputs follow the original directory structure of the dataset, enabling direct use in downstream modeling and analysis tasks.

\textbf{Foreground Segmentation.} To obtain human foreground masks for all multi-view sequences, we adopt a video-level SAM2~\cite{ravi2024sam2} pipeline initialized from single-frame detection. For each camera sequence, we first apply YOLOv8~\cite{ultralytics_yolov8} to the initial frame, retain person detections with confidence above 0.6, and keep the highest-confidence bounding box. This box is expanded by a factor of 1.05 to better cover fine structures such as hair, arms, and garment hems, and is used as the prompt for the SAM2.1-Hiera-Large video segmentation predictor. Since the first real frame may show boundary artifacts, we insert a duplicated version as a virtual frame~0 for SAM2 initialization. 
Segmentation runs from this virtual frame, but we keep results only from the original frame~1 onward to avoid these initialization artifacts. SAM2 then propagates the mask across the remaining frames. The output logits are binarized (threshold~$>$~0) and resized to the original resolution when necessary. This pipeline is automatically applied to all sequences, and the resulting foreground masks follow the same directory organization as the raw captures to support downstream tasks.

\textbf{Garment Segmentation.} When obtaining layered garment annotations on dressed humans, we observe that existing methods (e.g., Sapiens~\cite{sapiens} used in MVHumanNet++~\cite{liMVHumanNetLargescaleDataset2025}) often confuse inner and outer layers in multi-layer clothing scenarios. To address this issue, we leverage the semantic understanding capabilities of large vision--language models and design a two-stage fine-grained segmentation pipeline: (1) we prompt Qwen3-VL (Qwen3-VL-30B-A3B-Instruct-FP8)~\cite{qwen3vl2025} with the image and the available metadata to produce initial garment bounding boxes, which serve as prompts; and (2) we apply a single SAM2~\cite{ravi2024sam2} model to propagate these masks across the sequence and obtain per-frame layered garment segmentation.

Unlike foreground segmentation, which only requires detecting a single human instance, garment segmentation must correctly identify multiple clothing layers (e.g., inner/middle/outer upper garments and the lower garment). For each camera sequence, we apply Qwen3-VL to the first frame to obtain a structured JSON output containing 2D bounding boxes for initialization. These normalized coordinates are then converted to pixel space to initialize the SAM2.1-Hiera-Large video predictor. For every detected garment, we assign a unique object ID and perform joint tracking within the same video predictor. Similar to our foreground pipeline, we duplicate the first frame as a virtual frame~0 to stabilize SAM2 initialization; effective segmentation begins at the original frame~1, avoiding occasional boundary fluctuations in the true first frame. During initialization, all garment bounding boxes are added at frame~0, after which SAM2 propagates the mask of each garment across the sequence. For each garment, we binarize the per-frame logits and resize them to the original resolution when necessary. Finally, we store masks individually for each garment to support downstream tasks such as size estimation, layered garment modeling, and virtual try-on.

\textbf{Segmentation Verification.} To ensure the reliability of large-scale automatic segmentation, we conduct lightweight human quality checks across all sequences. Flat Catalogue Segmentation and Foreground Segmentation operate in clean, single-category environments and therefore exhibit almost no noticeable errors in our random inspections. In contrast, Garment Segmentation involves multi-layer clothing and multiple object labels, making it more susceptible to layer-related ambiguities. For each sequence, we therefore generate a collage of the first frames and manually inspect potential failure cases. The primary segmentation-related issue we observe is that multiple garment layers may occasionally be merged into a single mask in complex layered-outfit scenarios. In such cases, we refine the results by adjusting prompts, replacing initializers, or manually correcting ambiguous masks, ensuring accurate and consistent final annotations.

\subsection{Body Pose and Shape}

Our body pose and shape annotations are based on SMPL-X~\cite{pavlakosExpressiveBodyCapture2019}, a widely used human parametric model. Given a sequence of $n$ frames, we register body shape $\beta \in \mathbb{R}^{10}$,  pose $\theta \in \mathbb{R}^{n \times 156}$ (body and hand pose, as well as global orientation), and translation $t \in \mathbb{R}^{n \times 3}$. 

We infer these parameters with an automatic pipeline inspired by DNA-Rendering~\cite{chengDNARenderingDiverseNeural2023} and HuMMan~\cite{caiHuMManMultimodal4D2022}. (1) We detect 2D keypoints in the COCO-Wholebody format~\cite{jin2020whole} for each camera view with a pretrained RTMW-x model~\cite{jiang2024rtmw}. Keypoints detected outside the segmentation mask are discarded. (2) We leverage the calibrated intrinsic and extrinsic parameters of the cameras to triangulate 2D keypoints, obtaining several 3D keypoints. These estimated 3D keypoints are averaged per joint and further improved through bundle adjustment, in which outlier 2D keypoints are filtered by a threshold. Additionally, a temporal median filter smooths their movement among frames. (3) Finally, we optimize the SMPL-X parameters via a modified version of SMPLify-X~\cite{pavlakosExpressiveBodyCapture2019} in which we add the chamfer distance between SMPL-X vertices and the reconstructed point cloud. This extra loss provides enough guidance for the shape parameters fitting and is only performed for minimally clothed scans. Then, for all the clothed sequences of a subject, the shape remains fixed. 

\begin{figure}[htbp]
    \centering
    \includegraphics[width=\linewidth]{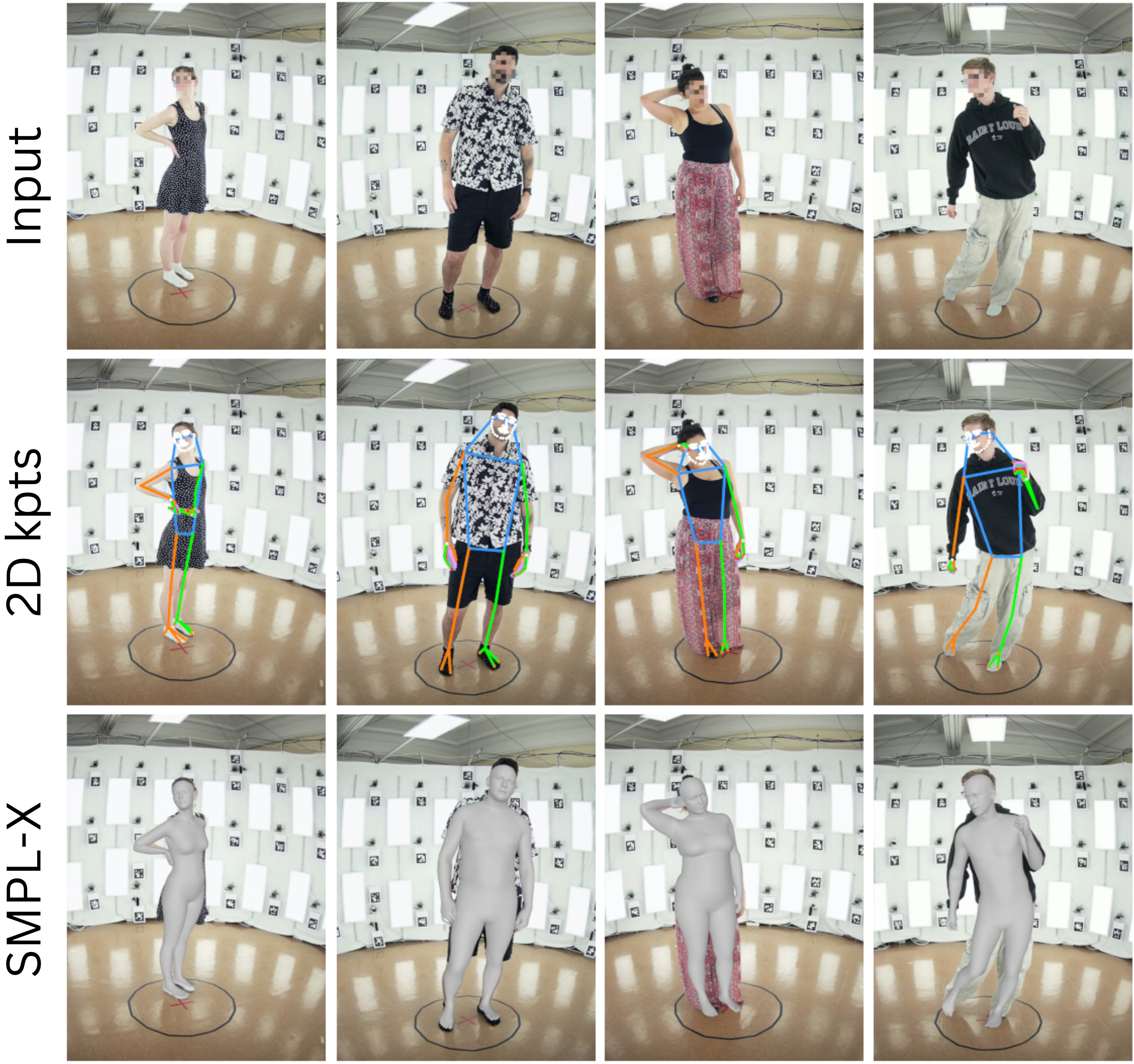}
    \caption{Qualitative results of our SMPL-X fitting pipeline. We display the input images, the 2D keypoints detected by RTMW-x in COCO-Wholebody format, and the final fitted SMPL-X model.}
    \label{fig:supp_smplx}
\end{figure}

Quantitatively, our SMPLify-X pipeline achieves a mean per joint position error (MPJPE) of 23.49\,mm and a chamfer distance of 55.13\,mm. See Fig.~\ref{fig:supp_smplx} for qualitative examples of the fitted SMPL-X.

\subsection{Garment Attributes} 
\label{supp:garment_attr}

\begin{figure*}[h]
    \centering
    \includegraphics[width=\linewidth]{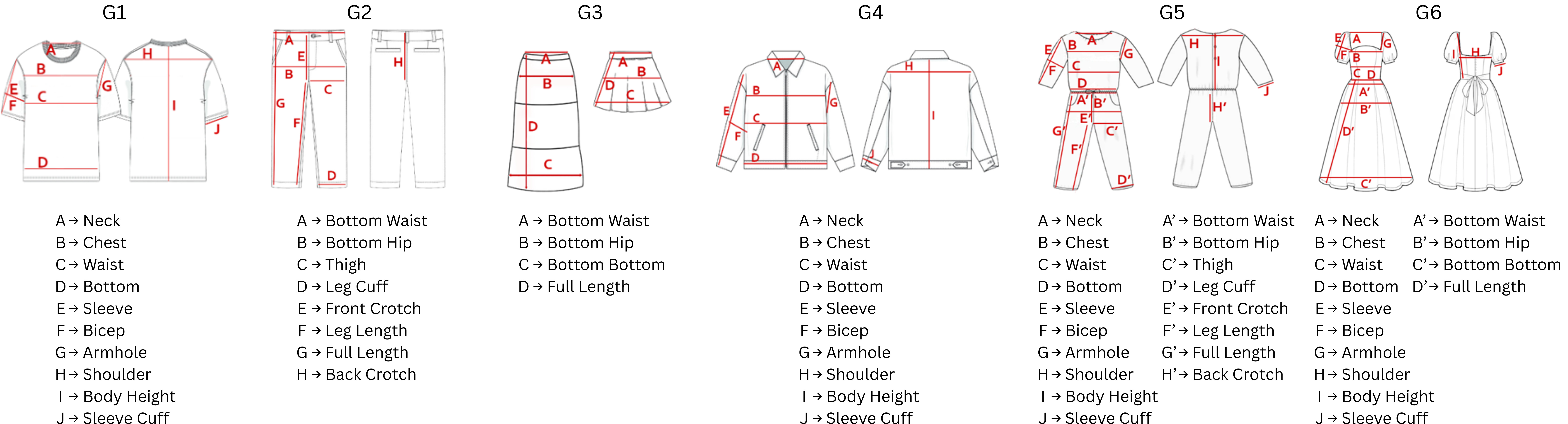}
    \caption{Visual definition of the sizing charts for each garment group (G1-G6). We annotate the measuring tape locations for all 19 measurement parts used in our dataset. Red lines indicate the distances measured.}
    \label{fig:supp_sizing_charts}
\end{figure*}

We define sizing chart annotations from a comprehensive set of 19 distinct measurement parts that span across the six garment groups (G1-G6) introduced in Sec.~\ref{sec:annotations} of the main paper. In Fig.~\ref{fig:supp_sizing_charts}, we provide a visual definition of these measurements, illustrating the specific start and end points for each metric. Each garment group utilizes a specific subset of these measurement parts, which refer to the upper body (\textit{Neck}, \textit{Chest}, \textit{Waist}, \textit{Bottom}, \textit{Sleeve}, \textit{Bicep}, \textit{Armhole}, \textit{Shoulder}, \textit{Body Height}, and \textit{Sleeve Cuff}) or the lower body (\textit{Bottom Waist}, \textit{Bottom Hip}, \textit{Bottom Bottom}, \textit{Thigh}, \textit{Leg Cuff}, \textit{Front Crotch}, \textit{Back Crotch}, \textit{Leg Length}, and \textit{Full Length}). Importantly, not all measurements within a group are applicable to every garment instance (e.g., a sleeveless T-shirt lacks \textit{Sleeve}, \textit{Bicep}, and \textit{Sleeve Cuff}). In such cases, we assign a special invalidity mark to these entries in the annotations. This mechanism is critical for the size estimation baseline described in Sec.~\ref{sec:size_baseline}, as it allows the model to predict only the measurements present on the garment.

\begin{figure}[h]
    \centering
    \includegraphics[width=\linewidth]{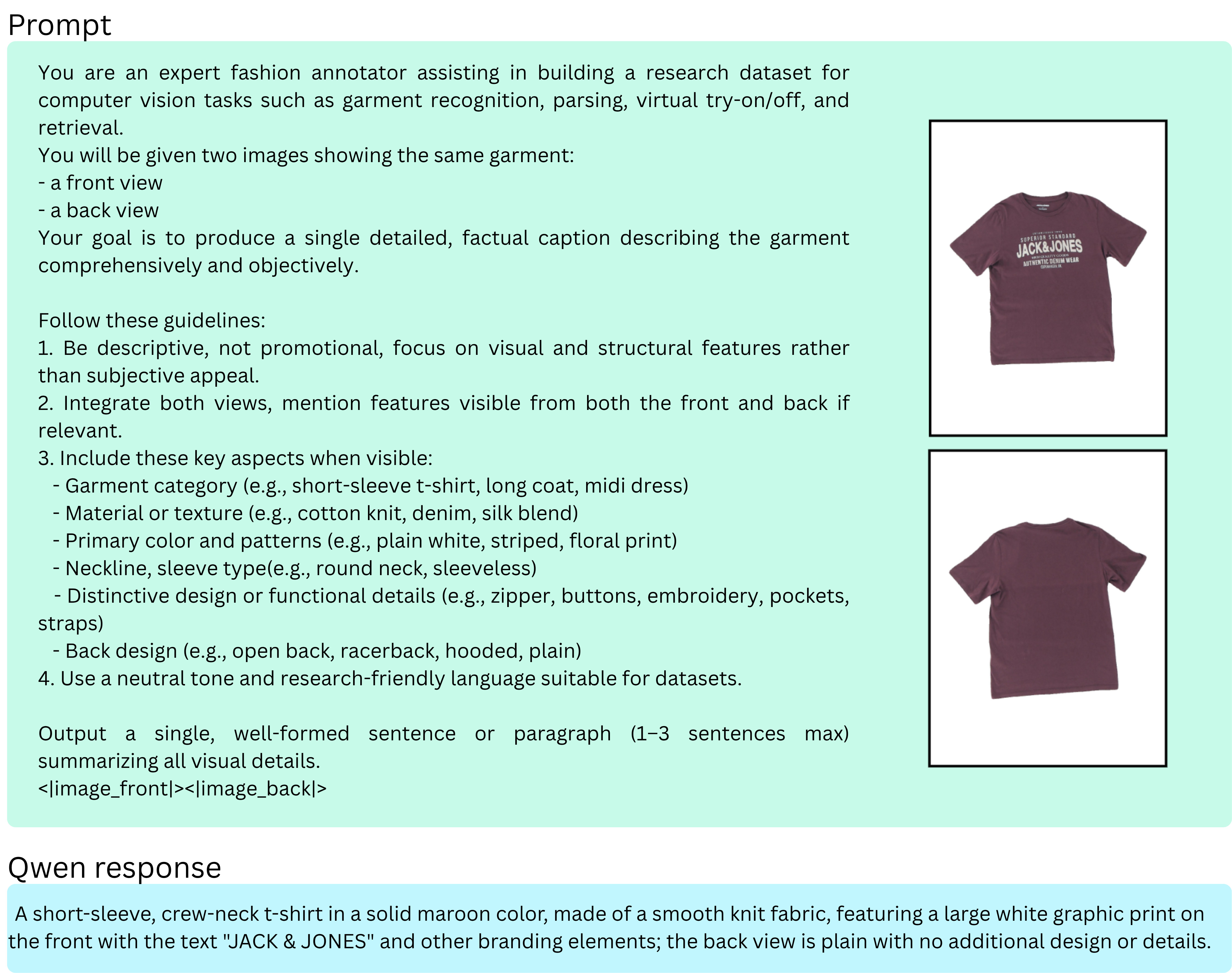}
    \caption{Qwen3-VL \cite{qwen3vl2025} garment description protocol. The input includes a detailed prompt and both views of the garment item to guide Qwen to generate descriptive and clear caption.}
    \label{fig:supp_qwen}
\end{figure}

Regarding annotations beyond numerical measurements, we generate the textual description for each garment item in MV-Fashion using the multi-modal large language model, Qwen3-VL (Qwen3-VL-30B-A3B-Instruct-FP8)~\cite{qwen3vl2025}. Fig.~\ref{fig:supp_qwen} illustrates the complete input for Qwen3-VL, including the textual prompt and both views of the garment, and a representative output. 

\begin{table*}[h]
    \centering
    \scriptsize
    \begin{tabular}{lccc}
        \toprule
        \textbf{Attribute} & \textbf{Layer} & \textbf{Categories} & \textbf{Encoding} \\
        \midrule
        Torso Closure Style & L1 \& L2 & n/a / fully closed / partially closed / fully open & 0 / 1 / 2 / 3 \\
        Long Sleeve Style & L1 \& L2 & n/a / rolled up / rolled down & 0 / 1 / 2 \\
        Tucking Style & L1 \& L2 & n/a / tucked / outside & 0 / 1 / 2 \\
        Hood Style & L2 only & n/a / hood up / hood down & 0 / 1 / 2 \\
        \bottomrule
    \end{tabular}
    \caption{Draping Style Attribute Encodings. n/a stands for not applicable.}
    \label{tab:draping_styles}
\end{table*}

We further enrich the dataset with four draping styles, defined across two layers: layer 1 (L1) for base garments, layer 2 (L2) for outerwear. The specific categories and their numerical encodings are detailed in Tab.~\ref{tab:draping_styles}.
Finally, we annotate the draping style for each sequence as a composite style code that concatenates the numerical encodings of each attribute from L1 and L2, using the underscore (\_) as a separator: (L1 Torso) \_ (L1 Sleeve) \_ (L1 Tucking) \_ (L2 Torso) \_ (L2 Sleeve) \_ (L2 Hood) \_ (L2 Tucking). 

\begin{figure}[h]
    \centering
    \includegraphics[width=\linewidth]{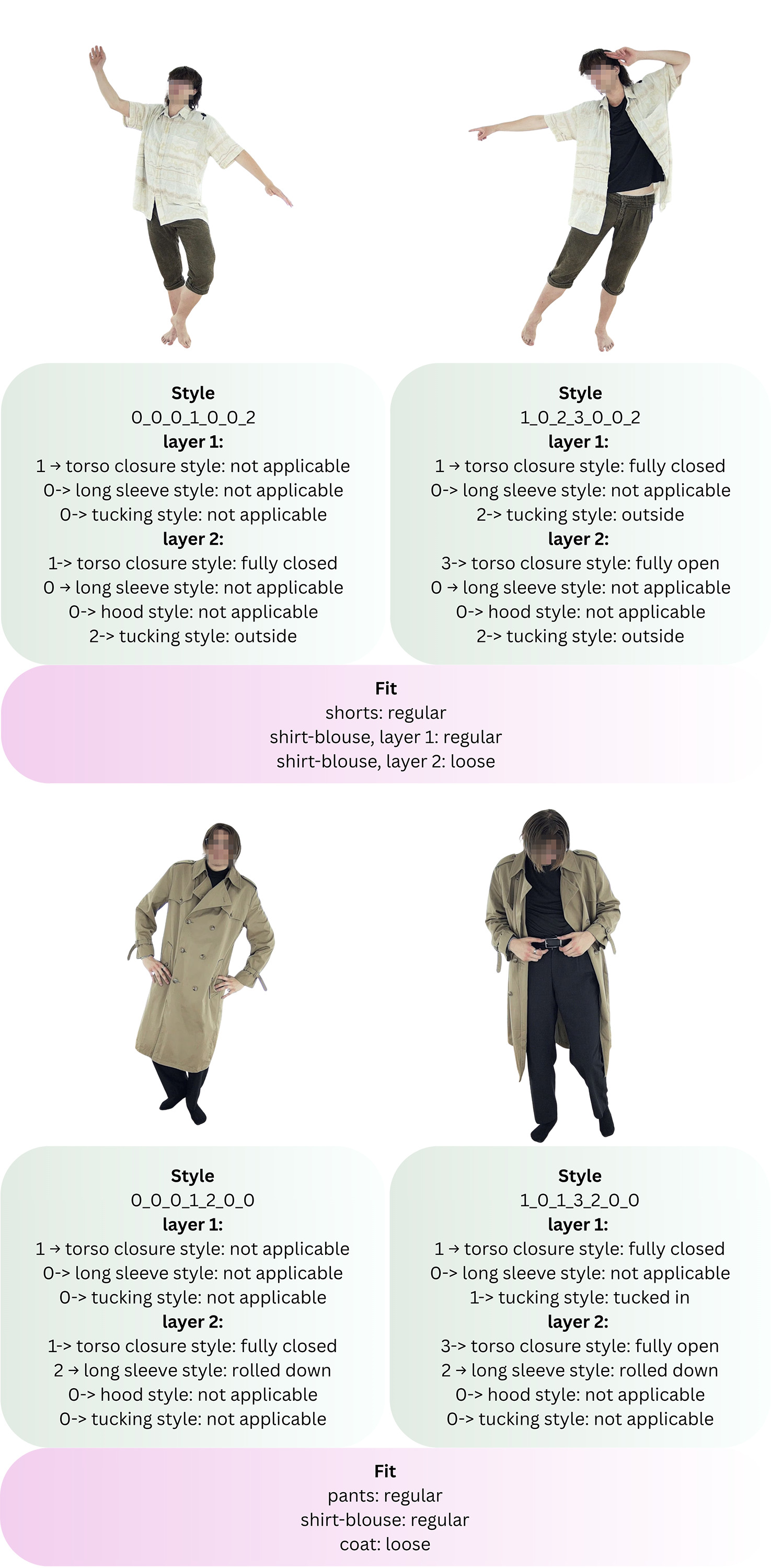}
    \caption{Some examples of the styling labels MV-Fashion provides. It is a unified string of numbers that encodes several styles we defined. We also provide the garment fit for each piece of clothing.}
    \label{fig:supp_styles}
\end{figure}

\begin{figure}[h]
    \centering
    \includegraphics[width=\linewidth]{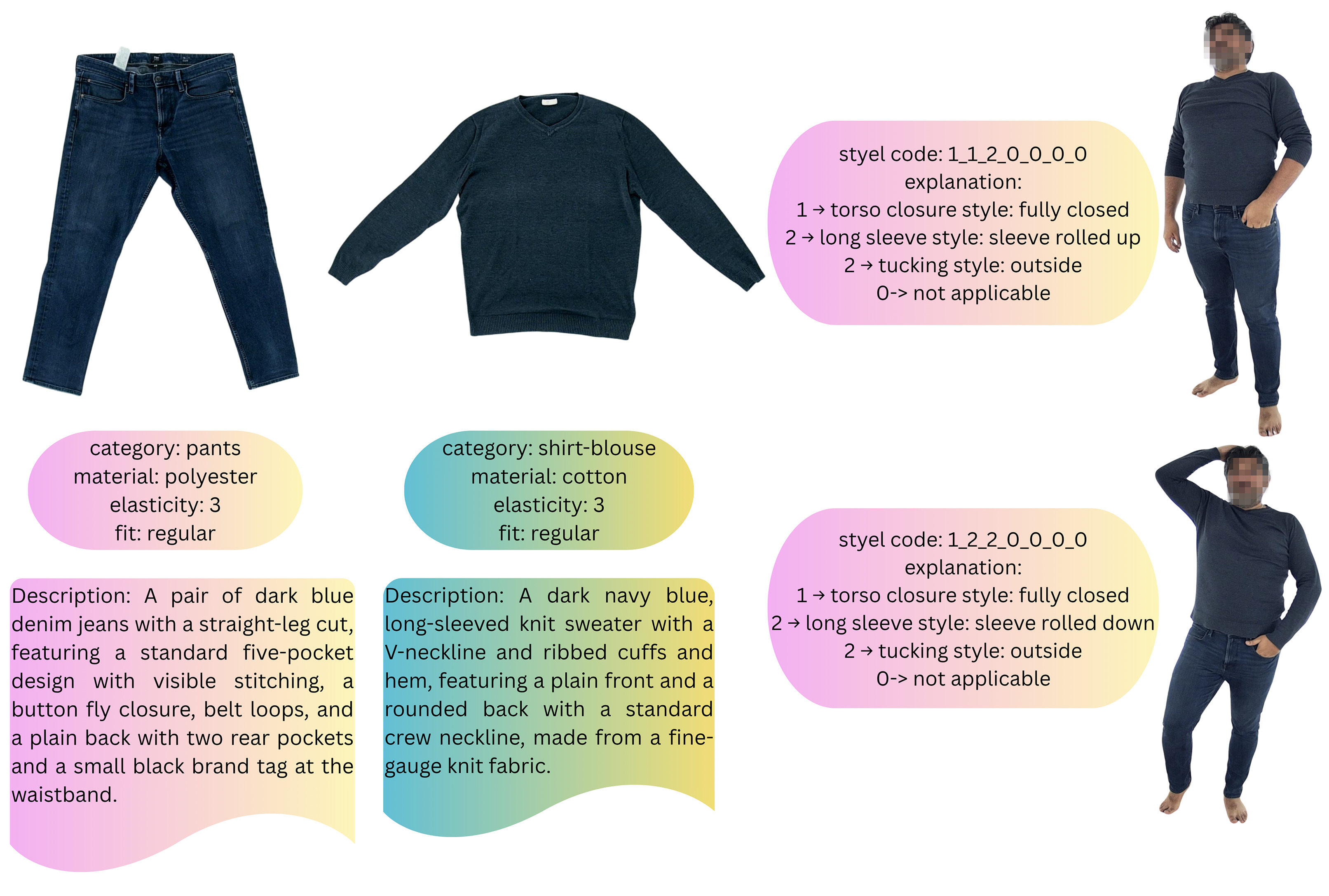}
    \caption{Additional example of text annotations MV-Fashion provides for each outfit. This includes a cloth category, material and elasticity details, our custom styling encoding and the detailed description.}
    \label{fig:supp_text_annotations}
\end{figure}

For more representative examples, we provide clear demonstrations of the garment's draping style and fitting appearance in Fig.~\ref{fig:supp_styles}. Additionally, full annotations set examples, including cloth category, material, elasticity, and textual description are given in Fig.~\ref{fig:supp_text_annotations}.

\subsection{Dataset Statistics}
\begin{figure}[h]
    \centering
    \includegraphics[width=\linewidth]{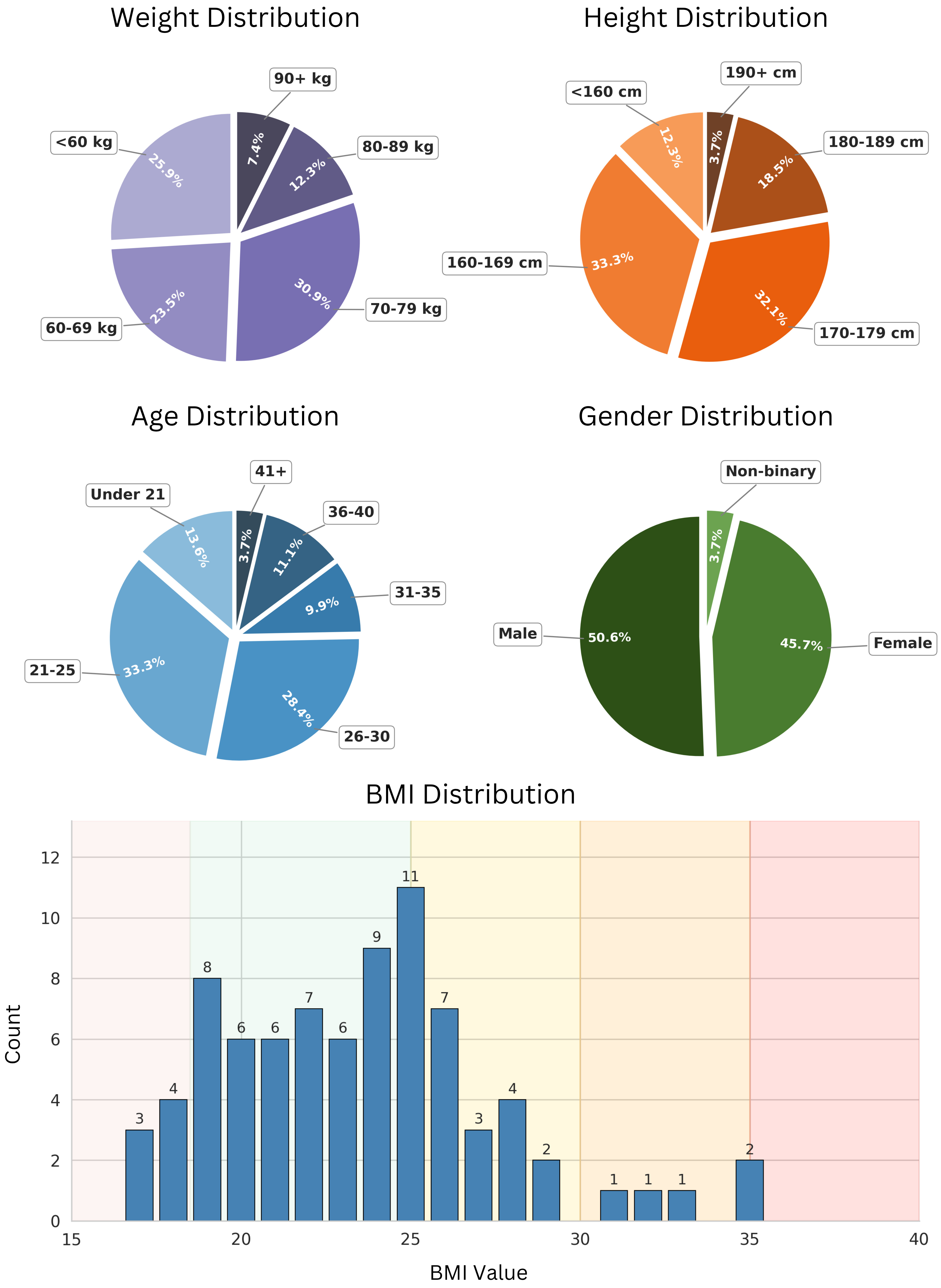}
    \caption{We collected additional information from each subject, like height, weight, age and gender, since these could be important for downstream tasks. They also help us evaluate and ensure a balanced distribution of our dataset.}
    \label{fig:supp_statistics}
\end{figure}

\begin{figure}
    \centering
    \includegraphics[width=\linewidth]{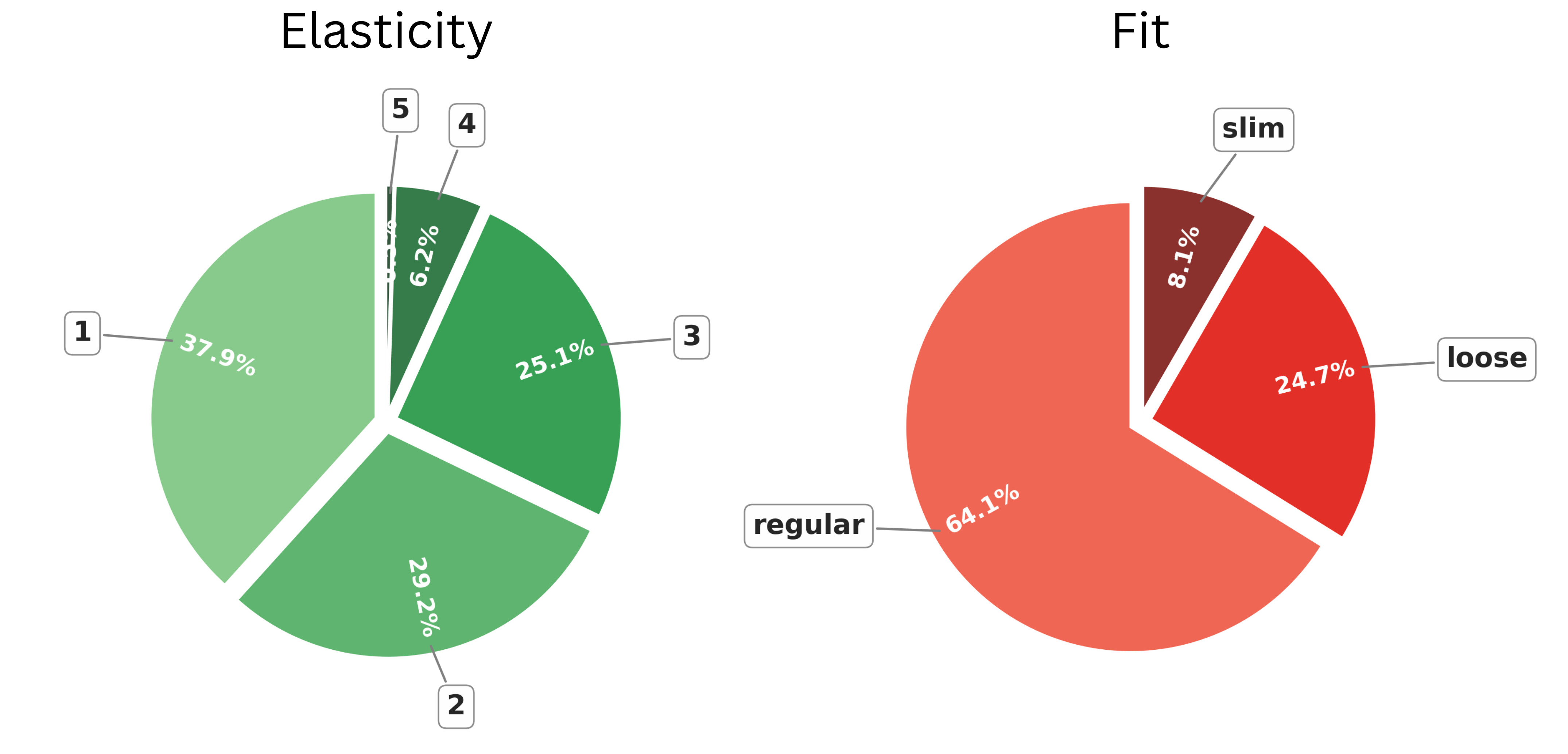}
    \caption{Distribution of different elasticity values and fit styles across the dataset. Highly elastic clothing items are rare, but the more common ones are well represented. The fit styles are also well distributed and MV-Fashion contains a high percentage of challenging, loose clothing. }
    \label{fig:supp_elasticity_fit}
\end{figure}

We provide additional statistics about the dataset in Fig.~\ref{fig:supp_statistics} and Fig.~\ref{fig:supp_elasticity_fit}. The subjects present a balanced distribution across height, gender and weight, while it leans towards the younger age range below 40 for the majority of the dataset. The BMI distribution also confirms that while the extreme ranges have only a few representative samples, the overall distribution is close to a Gaussian. 

We also expand on the distribution of the 14 garment categories in Tab.~\ref{tab:supp_dataset_stats}. It repeats the total, multi-layered, and styled distributions presented in the main paper. In the \textit{total} distribution, shirt-blouse represents the majority, followed by pants and shorts, which is expected for these common outfit types. Dresses and skirts also represent more than 8\% of the data. The rest are mostly outerwear, like jacket, sweater, sweatshirt or coat, which can appear in multi-layered clothing. The next \textit{layer} row depicts the percentage of each category that was part of a multi-layered outfit. The outerwear categories are close to or exactly 100\% since these are usually worn on top of another garment. Similarly, shirt-blouse has a high percentage as it is usually worn under an outer layer. Any lower body garment, like pants or skirts, is automatically 0\% since no other garment is worn on top of those in this dataset. Lastly, the \textit{style} row shows the percentage of each category that had multiple style variations. Again, outerwear-type garments have a high percentage, as these can often be worn open or closed. Shirts and sweaters are also present, since another common style variation is the rolling up of the sleeves or the tucking in of the shirt. Similarly to the layers, we do not define any styles for the lower body garments, as such these have 0\%. 

We also present some new statistics, namely, the garment fit (slim, regular and loose) and the recorded elasticity values' distribution. As expected, the regular fit is the most common in the majority of the categories, followed by the loose clothing in the range of 15-30\% in most cases. The slim fit is limited to some specific categories, including shirt-blouse, dress, sweatshirt, and cardigan, showing below 15\% representation. The exception is the tights-stockings with a 100\% slim fit. Similarly, the recorded elasticity values show that the dataset has a balanced coverage of different materials, as the values from 1 to 3 are well represented in most categories. Values 4 and 5 appear more often in specific categories, including dress, sweater, cardigan, and tights-stockings, which is expected in real life.

\begin{table*}[h]
\centering
\scriptsize

\begin{tabular}{l *{15}{c}}
\toprule
 & 
\rotatebox[origin=l]{90}{\parbox{1cm}{\raggedright shirt- \\ blouse}} & 
\rotatebox[origin=l]{90}{\parbox{1cm}{\raggedright pants}} & 
\rotatebox[origin=l]{90}{\parbox{1cm}{\raggedright shorts}} & 
\rotatebox[origin=l]{90}{\parbox{1cm}{\raggedright dress}} & 
\rotatebox[origin=l]{90}{\parbox{1cm}{\raggedright jacket}} & 
\rotatebox[origin=l]{90}{\parbox{1cm}{\raggedright sweater}} & 
\rotatebox[origin=l]{90}{\parbox{1cm}{\raggedright sweatshirt}} & 
\rotatebox[origin=l]{90}{\parbox{1cm}{\raggedright skirt}} & 
\rotatebox[origin=l]{90}{\parbox{1cm}{\raggedright coat}} & 
\rotatebox[origin=l]{90}{\parbox{1cm}{\raggedright jumpsuit}} & 
\rotatebox[origin=l]{90}{\parbox{1cm}{\raggedright vest}} & 
\rotatebox[origin=l]{90}{\parbox{1cm}{\raggedright cardigan}} & 
\rotatebox[origin=l]{90}{\parbox{1cm}{\raggedright blazer}} & 
\rotatebox[origin=l]{90}{\parbox{1cm}{\raggedright tights- \\ stockings}} &
\rotatebox[origin=l]{90}{\parbox{1cm}{\raggedright \textbf{Total}}} \\
\midrule
\textbf{Total} & $38.3$ & $21.2$ & $11.7$ & $5.2$ & $5.1$ & $3.8$ & $3.6$ & $3.4$ & $3.0$ & $1.3$ & $1.1$ & $0.9$ & $0.7$ & $0.1$ & $100.0$ \\
\textbf{Layer} & $51.6$ & $0.0$ & $0.0$ & $33.3$ & $97.4$ & $82.1$ & $81.5$ & $0.0$ & $100.0$ & $70.0$ & $100.0$ & $100.0$ & $100.0$ & $0.0$ & $39.02$ \\
\textbf{Style} & $7.7$ & $0.0$ & $0.0$ & $0.0$ & $86.8$ & $39.2$ & $55.5$ & $0.0$ & $59.1$ & $10.0$ & $62.5$ & $28.6$ & $60.0$ & $0.0$ & $14.13$ \\
\midrule
\textbf{Slim} & $12.2$ & $5.0$ & $0.0$ & $15.4$ & $7.9$ & $3.6$ & $14.3$ & $7.7$ & $0.0$ & $10.0$ & $0.0$ & $14.3$ & $0.0$ & $100.0$ & $8.1$ \\
\textbf{Regular} & $68.3$ & $55.3$ & $79.1$ & $59.0$ & $71.1$ & $64.3$ & $71.4$ & $65.4$ & $68.2$ & $70.0$ & $100.0$ & $42.9$ & $40.0$ & $0.0$ & $64.1$ \\
\textbf{Loose} & $19.5$ & $39.6$ & $20.9$ & $25.6$ & $21.1$ & $32.1$ & $14.3$ & $26.9$ & $31.8$ & $20.0$ & $0.0$ & $42.9$ & $60.0 $& $0.0$ & $24.7$ \\
\midrule
\textbf{Elasticity 1} & $18.9$ & $60.7$ & $59.3$ & $25.0$ & $69.2$ & $0.0$  & $13.8$ & $48.1$ & $76.2$ & 11.1 & $62.5$ & $28.6$ & $83.3$ & $0.0$ & $37.9$ \\
\textbf{Elasticity 2} & $30.2$ & $27.6$ & $27.5$ & $30.0$ & $23.1$ & $32.1$ & $51.7$ & $44.4$ & $23.8$ & 44.4 & $0.0$ & $0.0$ & $0.0$ & $0.0$ & $29.2$ \\
\textbf{Elasticity 3} & $40.9$ & $11.7$ & $9.9$  & $25.0$ & $7.7$  & $46.4$ & $27.6$ & $7.4$ & $0.0$ & 33.3 & $37.5$ & $42.9$ & $16.7$ & $0.0$ & $25.1$ \\
\textbf{Elasticity 4} & $9.6$& $0.0$ & $2.2$ & $15.0$ & $0.0$ & $21.4$ & $6.9$ & $0.0$ & $0.0$ & $11.1$ & $0.0$ & $28.6$ & $0.0$ & $100.0$ & $6.2$ \\
\textbf{Elasticity 5} & $0.3$ & $0.0$ & $1.1$ & $5.0$ & $0.0$ & $0.0$ & $0.0$ & $0.0$ & $0.0$ & $0.0$ & $0.0$ & $0.0$ & $0.0$ & $0.0$ & $0.5$ \\
\bottomrule
\end{tabular}
\caption{This table shows the distribution of different clothing categories across the dataset (Total). It also shows how often each clothing type shows up in multi-layered outfits (Layer) or in outfits with style variations (Style). Additionally, we show the distribution of our fitting style labels (slim, regular or loose) and the distribution of our elasticity annotation values for each garment category. The last \textit{Total} column shows the percentage of each annotation in the full dataset.  All values are percentages. The percentages for layers are calculated outfit-wise, while the rest of the values are calculated garment-wise.}
\label{tab:supp_dataset_stats}
\end{table*}

\section{Baselines Implementation Details} \label{supp:baseline_tech_details}

\subsection{Virtual Try-On}
\label{supp:vton_baseline}

\begin{figure*}[h]
    \centering
    \includegraphics[width=\linewidth]{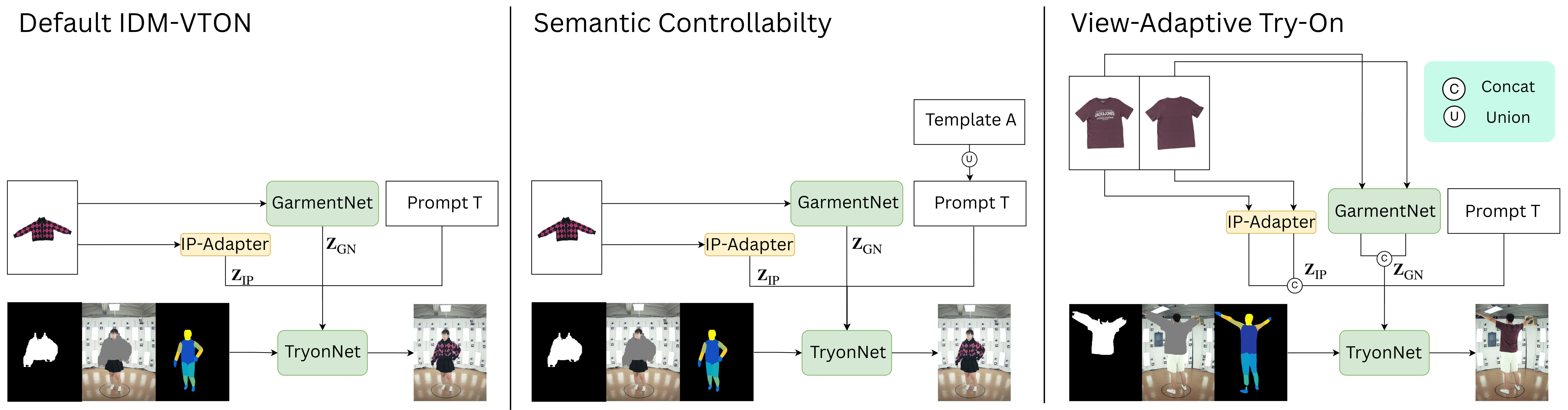}
    \caption{IDM-VTON architecture adaptation to perform Semantic Controllability and View-Adaptive Try-On tasks. For Semantic Controllability, we augment the prompt $T$ by styling annotation ($\mathbf{A}$). For View-Adaptive Try-On, we introduce feature fusion before conditioning, which concatenates the IP-Adapter's features ($\mathbf{Z}_{IP}$) and GarmentNet's features ($\mathbf{Z}_{GN}$) of both the frontal and rear garment views.
    }
    \label{fig:supp_vton_archmod}
\end{figure*}

As detailed in Sec.~\ref{sec:vton_baseline} of the main paper, we leverage the state-of-the-art architectures IDM-VTON~\cite{choi2024idm} and InsertAnything~\cite{song2025insert} for Single-View baseline with their default training configurations and adapt IDM-VTON for the Semantic Controllability and Multi-View Geometric Analysis baselines. We provide the corresponding modifications in Fig.~\ref{fig:supp_vton_archmod} and the following sections. 

\textbf{Semantic Controllability}. This experiment extends IDM-VTON to utilize the novel styling (draping and fitting) annotation for fine-grained control over the synthesized garment appearance.
We define the augmented text prompt $T'$ as the union of the original text prompt $T$ and a structured styling template derived from the categorical styling annotation $\mathbf{A}$ (defined in Sec.~\ref{sec:annotations} of the main paper and Sec.~\ref{supp:garment_attr}):
\begin{equation}
    T' = T \cup \text{Template}(\mathbf{A})
\end{equation}
where $T$ describes the base garment attributes and $\text{Template}(\mathbf{A})$ converts the categorical styling annotation $\mathbf{A}$ into a precise textual suffix (e.g., "outerwear is fully closed"). This union operation represents the combination of the input strings before encoding, which then conditions the IDM-VTON synthesis.

\textbf{Multi-View Geometric Analysis}. 
For the \textit{View-Adaptive Try-On}, we adapt the IDM-VTON architecture to leverage the paired frontal ($I_{C,F}$) and rear ($I_{C,R}$) garment views. We introduce a minimal modification involving feature fusion before conditioning the core model.
We utilize one shared IP-Adapter \cite{ye2023ipadapter} and one shared GarmentNet for processing both views. The resulting high-level feature tokens from the IP-Adapter ($\mathbf{Z}_{IP,F}, \mathbf{Z}_{IP,R}$) are concatenated to form a single combined feature vector $\mathbf{Z}_{IP} = \text{Concat}(\mathbf{Z}_{IP,F}, \mathbf{Z}_{IP,R})$. 
Similarly, low-level features $\mathbf{Z}_{GN}$ are fused by concatenating GarmentNet attention layers' features.
Additionally, we extend places for IP-Adapter's tokens in the cross-attention operation of TryonNet to handle doubled token length.
Finally, IDM-VTON is conditioned by the combined IP-Adapter feature vector $\mathbf{Z}_{IP}$, the combined GarmentNet features $\mathbf{Z}_{GN}$ and the prompt $T$.
This stands in contrast to the \textit{Cross-View Geometric Test}, where we quantify the challenge of cross-perspective mismatch by conditioning the model only with the frontal garment view $I_{C,F}$.

\subsection{Size Estimation} 
\label{supp:size_baseline}

As explained in Sec.~\ref{sec:size_baseline}, available approaches for garment sizing~\cite{kowaleczko2022neural, automatic2022garment} predict the measurements in controlled scenarios where the garments appear flat and unposed in the images. They rely on the 2D keypoints detection and perform the measurements between keypoints directly in the pixel space, which are not generalizable on our posed data. A natural extension of these works is to reconstruct the garment 3D mesh~\cite{zhuDeepFashion3DDataset2020}, lift the detected 2D keypoints into 3D and compute the sizes on the 3D surface. However, this approach requires complex intermediate processing of 3D data, and we consider 3D surface reconstruction out of the scope of this paper. Instead, SPnet~\cite{lim2024spnet} reconstructs the 3D surface by predicting the sewing pattern parameters from the normal image of the clothing, which is estimated beforehand. This normal image is given in a canonical T-pose to handle challenges like occlusions and wrinkles. Finally, the 3D mesh is generated from the sewing pattern by computer graphics techniques. We find the two-stage pipeline of SPnet is easily adaptable to our data without needing complex 3D processing, i.e., by replacing the sewing patterns data with our sizing charts. Specifically, our baseline consists of a garment normal predictor ($\Psi$) and a garment size regressor ($\Phi$). Since an outfit in a given image can have upper and lower body garments with multiple layers, we only consider the outermost visible layer, assuming the garment category (and hence its group) is known beforehand. Therefore, we exclusively focus on the garment size estimation. Also, we deliberately avoid geometric data augmentation; by relying solely on our data, we assess if it provides sufficient information to estimate garment size on its own, demonstrating that the natural variation provided by our multi-subject, multi-view capture system is robust enough to generalize without synthetic modifications. Furthermore, augmentations like random scaling or cropping would corrupt the relation between $G^t$ and our measurement annotations unless accompanied by highly specific, non-trivial adjustments.

\begin{figure}[htbp]
    \centering
    \includegraphics[width=\linewidth]{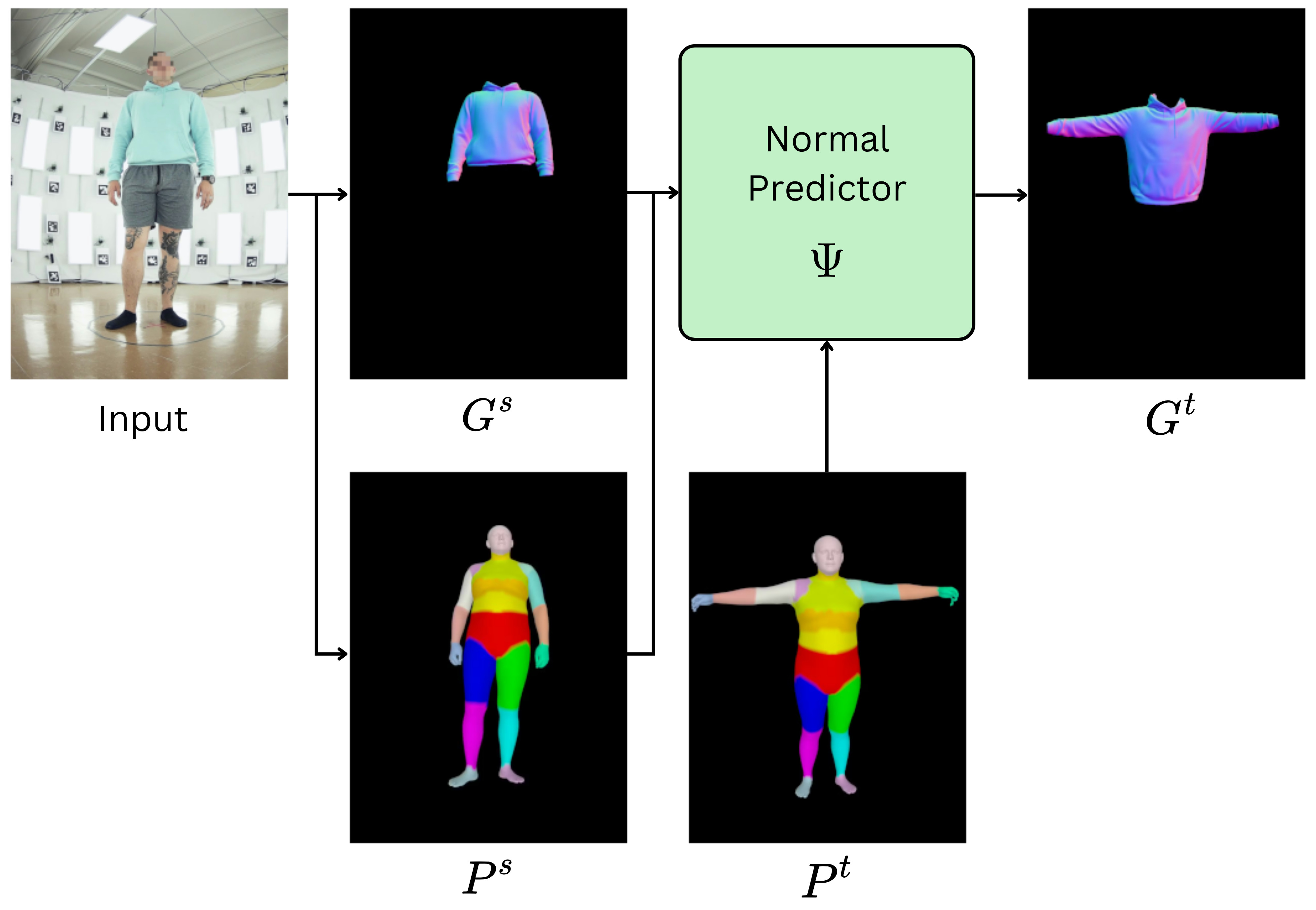}
    \caption{\textbf{The normal predictor ($\Psi$).} To disentangle intrinsic garment geometry from pose deformations, the network transforms the source garment normals $G^s$ into a canonical, unposed state $G^t$. Both $G^s$ and $P^s$ are extracted from the input frame, and the transformation is explicitly conditioned on a target pose $P^t$. Crucially, both pose maps ($P^s, P^t$) utilize semantic body segmentation to provide spatial correspondence cues.}
    \label{fig:normal_predictor}
\end{figure}

\textbf{Normal Predictor.} For $\Psi(G^{s}, P^{s}, P^{t})$, the goal is to transform the garment ($G^{s}$) from a posed state ($P^{s}$) into a canonical, unposed state ($P^{t}$) to facilitate the garment measurement. This process is illustrated in Fig.~\ref{fig:normal_predictor}. All pose maps involved in the baseline ($P^s$ and $P^t$) are semantic body segmentation maps rather than simple skeleton renderings. Specifically, we apply an \texttt{argmax} operation to the SMPL-X blend weights to assign each pixel to a specific body part (e.g., the waist, the hip). By doing so, we provide the model with explicit spatial attention cues, directly linking visual body regions to the different parts of the garment. The data preparation process for $G^s$, $P^s$ and $P^t$ is detailed in Sec.~\ref{sec:size_baseline} of the main paper.

\textbf{Size Regressor.} The second network estimates garment measurements from the canonical normals ($G^t$). We evaluate three architectures for $\Phi$. (1) \textit{Per-Group}: This is the direct implementation of the SPnet sewing pattern regressor repurposed for the garment measurement. Following their protocol, we train a separate network for each garment group (G1–G6, see Sec.~\ref{supp:garment_attr}), limiting the training data for each model to only the samples available for that specific group. (2) \textit{Multi-Task}: To leverage the overlapping data among all the groups, we use our annotation protocol to perform multi-task learning. We train a single unified network by grouping common measurements (e.g., \textit{Chest}) that appear across groups (see Fig.~\ref{fig:supp_sizing_charts} for overlaps). This significantly increases the effective number of training samples per measurement, improving the model's ability to learn robust features for shared body regions. (3) \textit{Multi-Task + SwinV2}: We replace the original SegNet~\cite{segnet} encoder with a pretrained SwinV2~\cite{liu2021swinv2}, and additionally condition $\Phi$ on the target pose $P^{t}$ to guide the network's spatial information towards the corresponding entries in the sizing chart.

\subsection{Novel View Synthesis}
As detailed in Sec.~\ref{sec:nvs_baseline} of the main paper, to validate the applicability of MV-Fashion for NVS we run several benchmarks and ablations. We choose Nerfstudio as the framework to run our novel view synthesis benchmarks, as it aggregates multiple popular and state-of-the-art methods, making them easy to test. It also supports a common data preprocessing pipeline, which has added features that the original implementations might lack. With the three selected methods (\texttt{instant-ngp}, \texttt{nerfacto}, \texttt{splatfacto}), we can cover both major categories, those being Nerf and 3D Gaussian Splatting. Most recent downstream tasks rely on one of these methods; thus, validating them on our dataset is important. 

Additionally, we briefly explore CEM-4DGS~\cite{kangClusteredErrorCorrection2025} a 4D Gaussian Splatting model to validate that our dataset can support dynamic novel view synthesis as well.

\section{Experiments and Results}
\label{supp:experiments}

\subsection{Virtual Try-On}

\textbf{Training Setup.}
We fine-tuned IDM-VTON and InsertAnything using their original training protocols. For all experiments, we uniformly resize input images to ${1024 \times 768}$. For the IDM-VTON protocol (applied to all IDM-VTON baselines), we utilize a batch size of ${12}$ for ${130}$ epochs, employing the AdamW optimizer \cite{adamW} with a learning rate of ${2 \times 10^{-5}}$. For the InsertAnything protocol, we utilize the Prodigy optimizer \cite{prodigy} as the default with safeguard warmup and bias correction enabled. The initial step size $(\gamma_k)$ is set to $1$, and we use a weight decay of $0.01$. We train the model for ${15,000}$ steps with a batch size of ${6}$, leveraging Low-Rank Adaptation (LoRA) with $r={256}$ and $\alpha={256}$. We perform all training on a single NVIDIA H100 GPU (80GB), requiring approximately ${90}$ hours for IDM-VTON and ${100}$ hours for InsertAnything.

\begin{figure*}[!h]
    \centering
    \includegraphics[width=\linewidth]{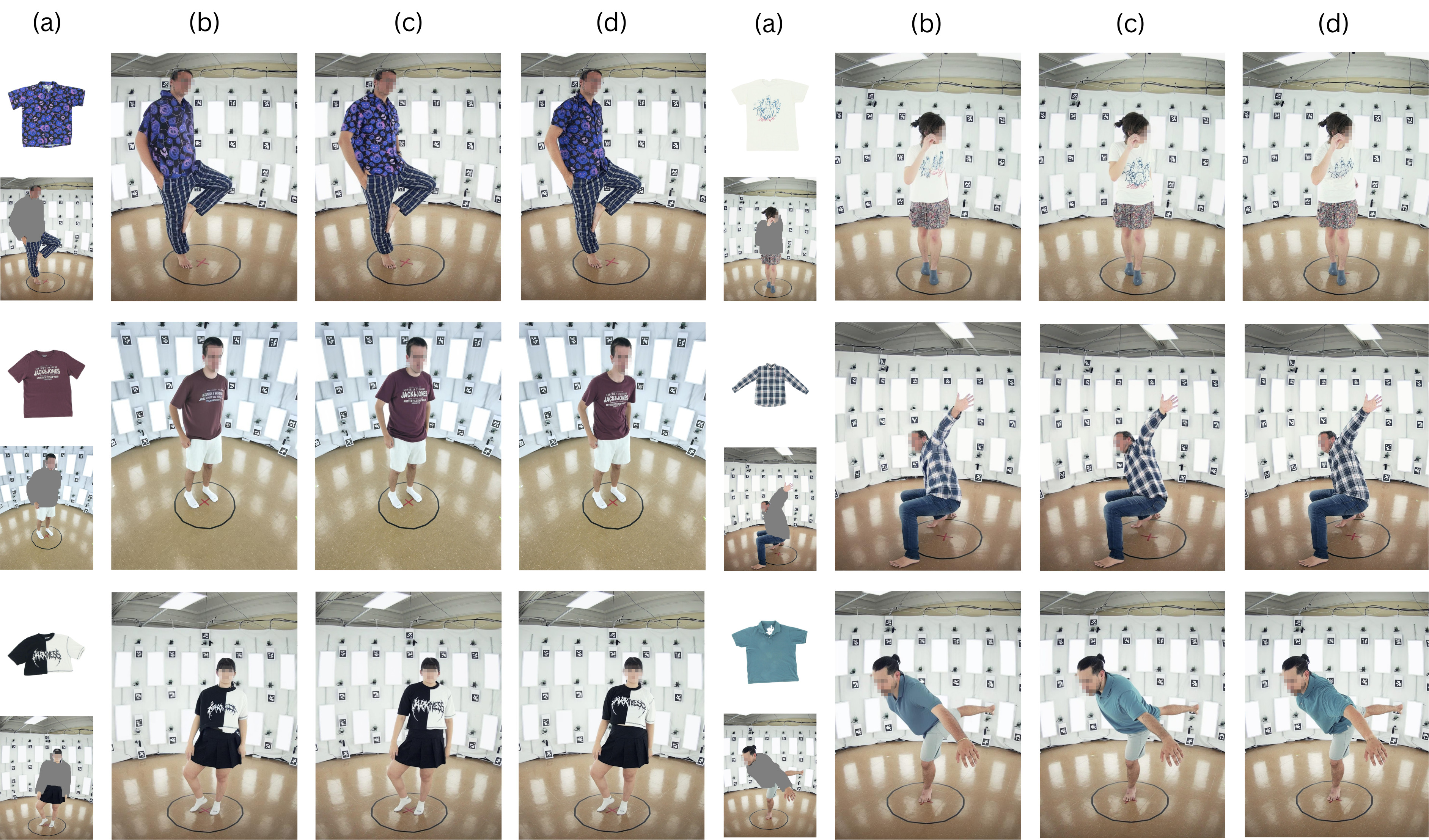}
    \caption{Qualitative results of IDM-VTON and InsertAnything trained on the Single-View MV-Fashion benchmark. The high fidelity achieved by both models confirms the direct compatibility of MV-Fashion with existing state-of-the-art VTON datasets. The columns represent: (a) Input (Garment and Person Mask), (b) IDM-VTON Try-On Result, (c) InsertAnything Try-On Result, (d) Ground Truth.}
    \label{fig:supp_sv_compare}
\end{figure*}

\textbf{Qualitative Results for Single-View Baselines.}
Fig.~\ref{fig:supp_sv_compare} provides visual examples of the Single-View VTON Baselines, complementing the quantitative metrics reported in the main paper (Tab.~\ref{tab:vton_benchmark}). Both models achieve visually high-fidelity and geometrically plausible try-on results, successfully transferring the garment texture and structure onto the target person's pose. Notably, InsertAnything, which yielded the best quantitative scores, demonstrates exceptional realism, confirming that the frontal-paired subset of MV-Fashion is directly compatible with existing state-of-the-art VTON pipelines.

\begin{figure*}[!h]
    \centering
    \includegraphics[width=\linewidth]{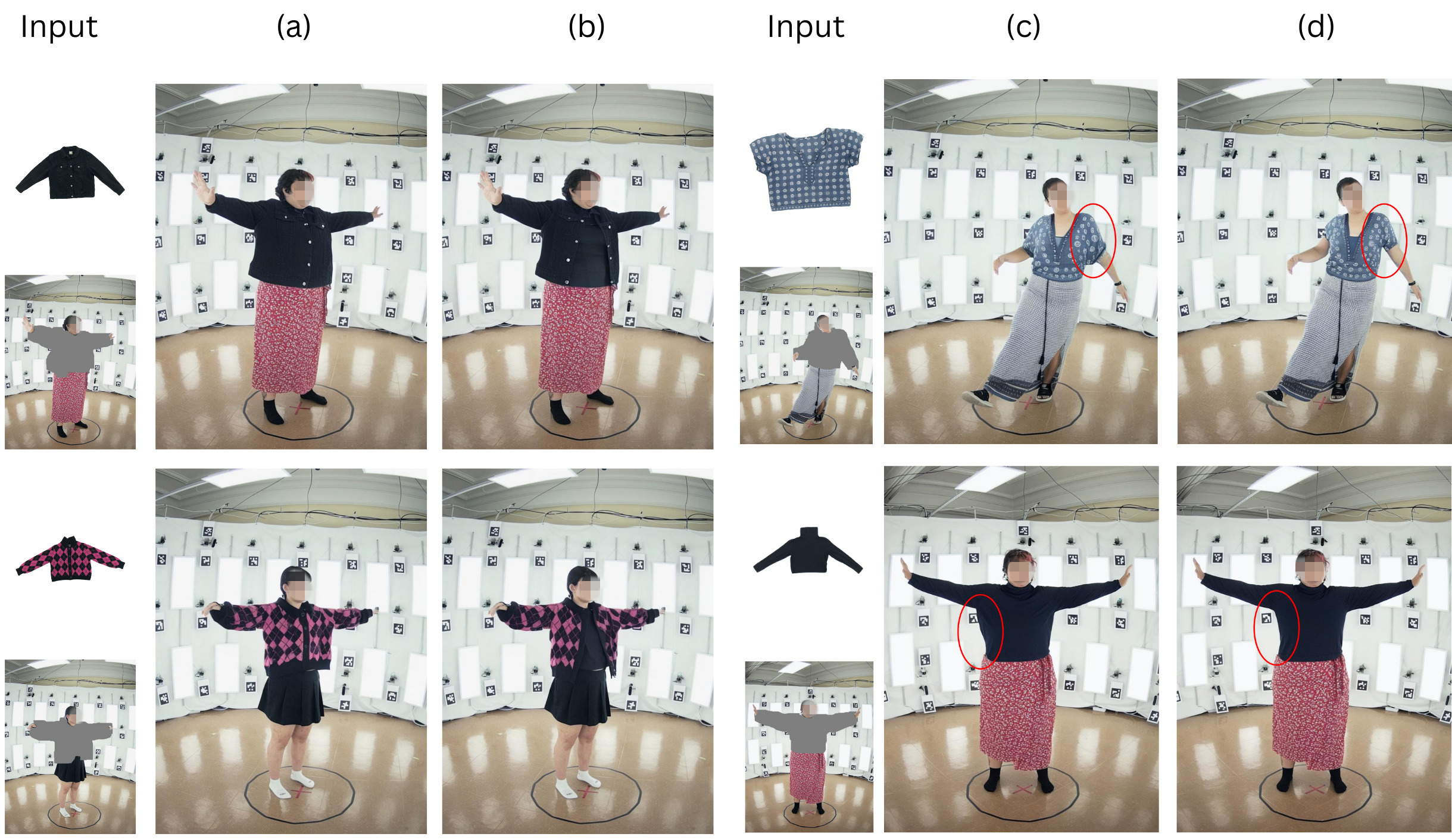}
    \caption{Extended qualitative results for Semantic Controllability. The original experiment augments draping style in prompt when training and compare (a) no-style to (b) draping styling prompt (\eg \textit{outerwear is fully open} for both rows) at test time. Additionally, we test the second experiment to train with fitting style and compare (c) no-style to (b) fitting styling prompt (\eg \textit{regular fit} for row 1 and \textit{slim fit} for row 2) at test time.}
    \label{fig:supp_styling_pt}
\end{figure*}

\textbf{Semantic Controllability Qualitative Results.}
Fig.~\ref{fig:supp_styling_pt} provides additional examples of the Semantic Controllability experiment with draping style. 
As discussed in the main paper, when finetuning with MV-Fashion draping style-augmented data, we observe that the model responds to the styling prompt (column (b)) compared to the no-style prompt (column (a)). In both rows shown, the garments visually react to the prompt by opening the jacket buttons/zip.
Additionally, we also expand the experiment that tests the model's capability when finetuning with fitting style-augmented data. Interestingly, the model shows an emerging ability to follow these styling instructions when comparing the styling prompt (column (d)) to the no-style prompt (column (c)). Specifically, the upper garment in the first row fits closer in the sleeve region, aligning with the "regular fit" prompt, while the garment in the second row becomes tighter in the chest area, aligning with the "slim fit" prompt.

\begin{figure}[!ht]
    \centering
    \includegraphics[width=\linewidth]{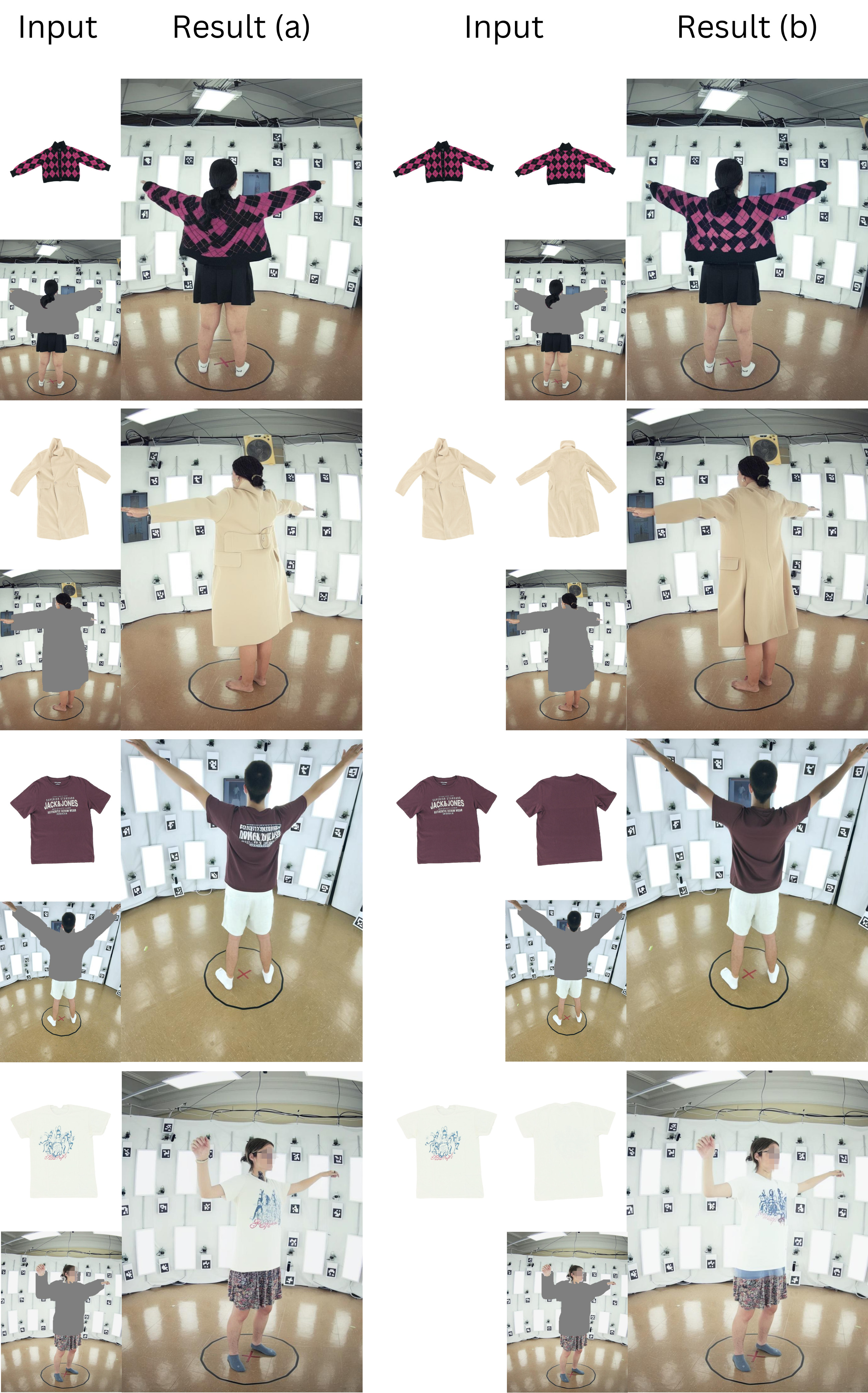}
    \caption{More qualitative results of IDM-VTON on (a) \textit{Cross-View Geometric Test} vs (b) \textit{View-Adaptive Try-On}. The updated IDM-VTON architecture can map between the catalogue view and the person's pose when both frontal and rear images of the garment are provided to the model.}
    \label{fig:supp_mv_qualitative}
\end{figure}

\textbf{Multi-View Qualitative Comparison}.
Fig.~\ref{fig:supp_mv_qualitative} provides the qualitative comparison between the two Multi-View Geometric Analysis experiments. Column (a), representing the Cross-View Geometric Test, shows the model's limitations when synthesizing rear target poses using only the frontal catalogue image. We observe issues stemming from perspective misinferring and poor cross-perspective consistency, leading to structural distortion and inaccurate pattern mapping.
In contrast, column (b), the View-Adaptive Try-On, demonstrates a slight visual improvement. By utilizing both frontal and rear garment images, the model successfully mitigates several geometric and texture errors, achieving better pattern alignment and structural fidelity in the rear poses across examples.

\begin{figure}[h]
    \centering
    \includegraphics[width=1\linewidth]{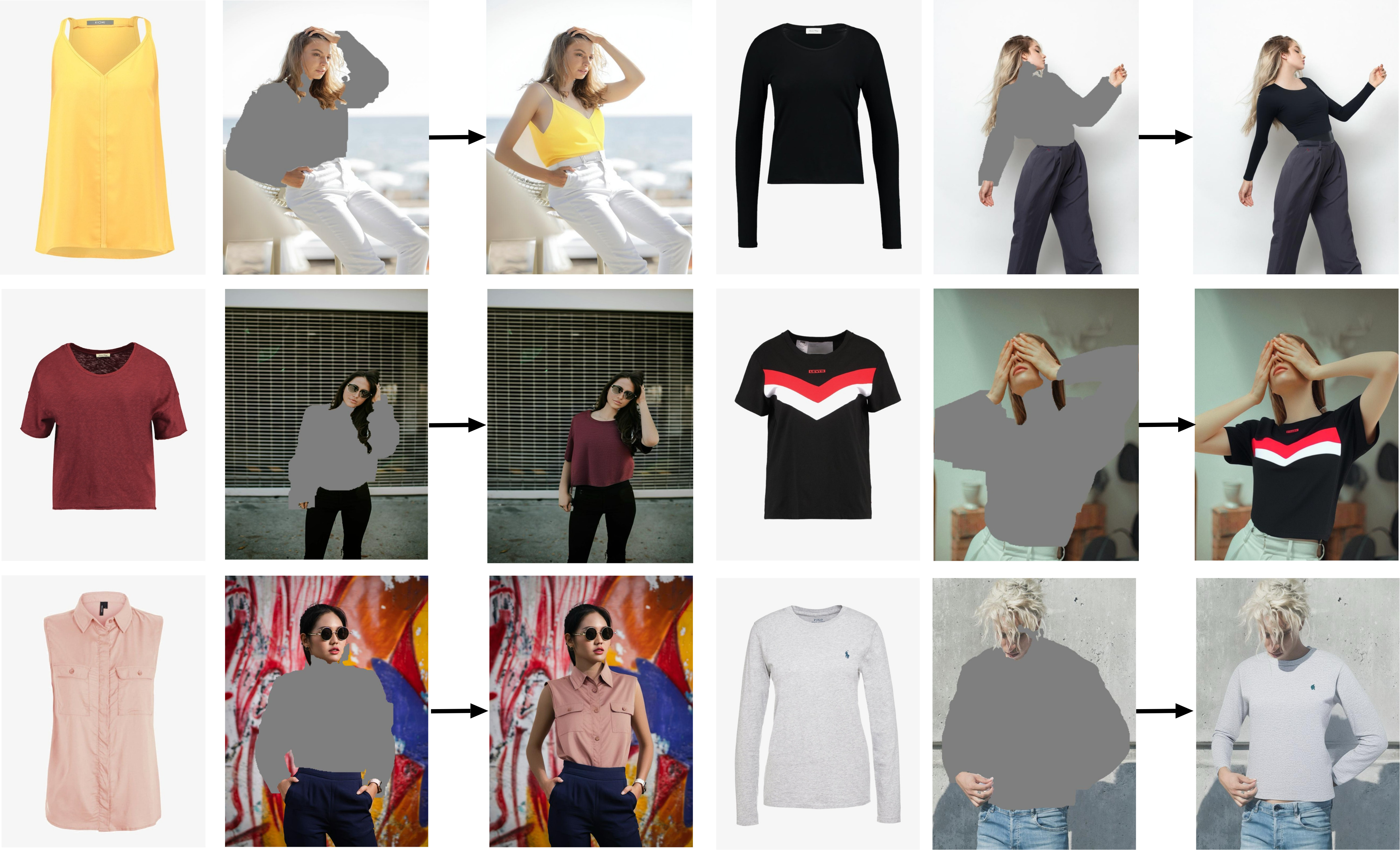}
    \caption{Qualitative examples of our VTON pipeline when using in-the-wild images with complex backgrounds. Our model demonstrates effective generalization to these out-of-distribution samples. (images obtained from unsplash.com)}
    \label{fig:supp_vton_wild}
\end{figure}

\textbf{In The Wild Images}
We further test our VTON baseline trained on MV-Fashion on in-the-wild images from Unsplash to evaluate its robustness. Our model shows promising generalization to these out-of-distribution samples, suggesting that our controlled setup translates effectively to less constrained environments. For a qualitative evaluation, please refer to Fig.~\ref{fig:supp_vton_wild}.

\textbf{Conclusion}.
Overall, the results show that MV-Fashion dataset is directly compatible with standard Single-View VTON research. 
For Semantic Controllability, despite the observed successes, the overall task of fine-grained semantic control remains highly difficult and will require significant architectural advancements to achieve reliable and consistent control, for which our dataset serves as an essential testbed. 
For View-Adaptive Try-On, a partial reduction in failure cases validates the potential of MV-Fashion's multi-view pairs to enable effective viewpoint-aware VTON research, though achieving perfectly seamless cross-perspective fusion remains a valuable future investigation.

\subsection{Size Estimation}

\textbf{Training Setup.} We adhere to a consistent protocol for all baseline variants. Input images are resized to $256 \times 256$, while we utilize the standard Adam optimizer~\cite{adam} with a learning rate of $1 \times 10^{-4}$ and a batch size of 24. All models are trained on a single NVIDIA GeForce RTX 4090 GPU (24GB). Training duration varies by component: the canonical normal predictor ($\Psi$) requires approximately 34 hours to train for 5 epochs. In contrast, the size regressor ($\Phi$) is lightweight; it takes less than one hour to train for 100 epochs (per model variant). We also train an \textit{End-to-End} baseline in which $\Psi$ and $\Phi$ are initialized from their independent pretrained models, $\Psi$ is frozen and $\Phi$ is fine-tuned for 5 epochs in approximately 18 hours.

\begin{figure*}[!h]
    \centering
    \includegraphics[width=\linewidth]{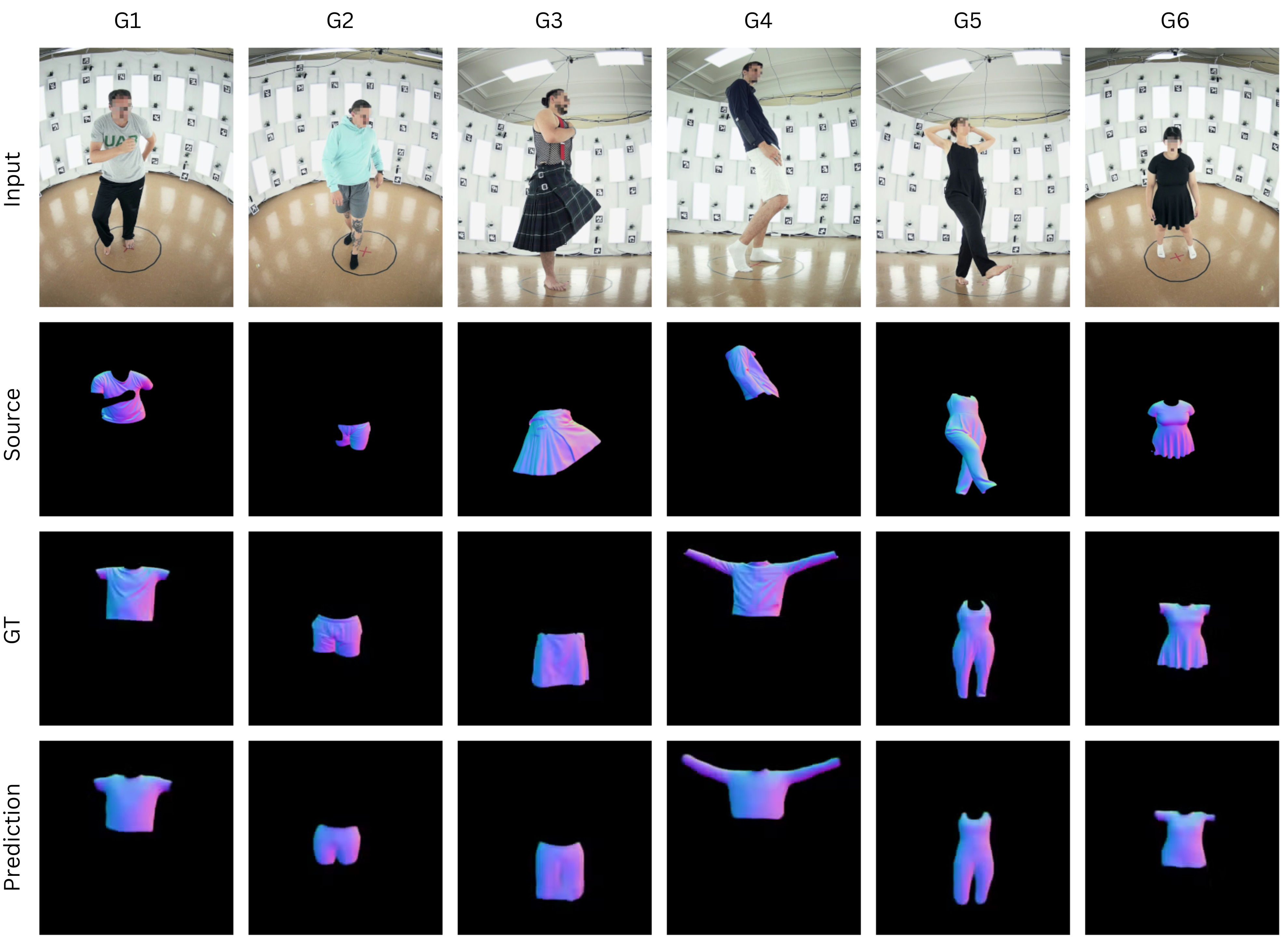}
    \caption{Qualitative results of the garment normal predictor ($\Psi$) across all garment groups (G1-G6). From top to bottom: the input RGB frame, the source garment normal map ($G^s$), the target ground truth canonical normal map ($G^t$), and the predicted canonical normals. The results demonstrate the model's ability to disentangle pose from garment geometry, effectively recovering the intrinsic shape even in cases of occlusion (e.g., G1) or complex deformations (e.g., G4).}
    \label{fig:supp_sizing}
\end{figure*}

\textbf{Qualitative Results.}
In Fig.~\ref{fig:supp_sizing}, we provide qualitative results of the normal predictor across all six garment groups (G1-G6). These examples illustrate the inherent difficulty of the task: the source normals ($G^s$) are often heavily occluded (see G1) or distorted by perspective and body leaning (see G4 and G5). Despite these geometric challenges, the model successfully recovers the flattened, canonical shape of the garment. Furthermore, in the case of G3, the input normal map exhibits severe self-occlusion due to the leg movement; however, the predictor successfully recovers the missing geometry to reconstruct a complete canonical skirt.

\textbf{Quantitative Results.} Tab.~\ref{tab:sizing_metrics} reveals a clear correlation between data frequency (see Tab.~\ref{tab:dataset_stats}) and model performance for the \textit{Per-Group} variant. Groups G1 and G2, which constitute the majority of the dataset ($38.3\%$ and $33.0\%$, respectively), achieve significantly lower errors than the statistically rarer groups. However, the \textit{Multi-Task} strategy effectively mitigates this data imbalance; notably, it reduces the error for G5 from 12.109\,cm to 4.295\,cm.

To further analyse this improvement, we provide detailed breakdowns of Tab.~\ref{tab:sizing_metrics} in Tabs.~\ref{tab:g1_metrics}-\ref{tab:g6_metrics}, reporting the error per measurement part (as defined in Sec.~\ref{supp:garment_attr}) for each garment group. These breakdowns demonstrate how the \textit{Multi-Task} variant leverages shared measurement parts across different garment groups. For instance, the \textit{Leg Length} error in G5 is reduced by over 26\,cm compared to the \textit{Per-Group} baseline.

Finally, to assess the feasibility of estimating size without relying on ground truth canonical garment normals ($G^t$), we introduce the \textit{End-to-End} baseline results in Tab.~\ref{tab:suppl_sizing}. This model fine-tunes the best-performing variant of $\Phi$ (\textit{Multi-Task + SwinV2}) to regress sizes directly from the input image. While this approach naturally yields a higher MAE than the others, it achieves a respectable average error of 6.134\,cm. This demonstrates that the signals within MV-Fashion enable direct, image-based garment size learning.

\textbf{Conclusion.} The results indicate our annotations are sufficient to train viable garment size estimation models. By enabling the learning of mappings from in-the-wild deformations to garment measurements, MV-Fashion establishes a strong foundation for future research in cloth sizing.

\begin{table}[h]
\centering
\scriptsize
\setlength{\tabcolsep}{2.7pt}
\begin{tabular}{l c c c}
\toprule
\textbf{Measurement} & \textbf{Per-Group} & \textbf{Multi-Task} & \textbf{Multi-Task + SwinV2} \\
\midrule
\textbf{Neck} & $2.400$ $(1.976)$ & $2.581$ $(2.163)$ & $2.246$ $(1.852)$ \\
\textbf{Chest} & $5.170$ $(4.053)$ & $6.370$ $(4.423)$ & $5.339$ $(3.946)$ \\
\textbf{Waist} & $4.265$ $(4.114)$ & $4.516$ $(4.344)$ & $4.581$ $(4.490)$ \\
\textbf{Bottom} & $3.553$ $(2.921)$ & $4.347$ $(2.652)$ & $3.659$ $(3.118)$ \\
\textbf{Sleeve} & $7.903$ $(14.074)$ & $8.022$ $(14.391)$ & $7.058$ $(14.679)$ \\
\textbf{Bicep} & $1.672$ $(1.177)$ & $1.838$ $(1.418)$ & $1.620$ $(1.096)$ \\
\textbf{Armhole} & $2.461$ $(2.068)$ & $2.538$ $(2.044)$ & $2.459$ $(2.057)$ \\
\textbf{Shoulder} & $4.605$ $(3.748)$ & $4.592$ $(4.024)$ & $4.986$ $(3.768)$ \\
\textbf{Body Height} & $6.612$ $(6.712)$ & $6.965$ $(6.925)$ & $5.568$ $(6.954)$ \\
\textbf{Sleeve Cuff} & $2.026$ $(1.646)$ & $1.782$ $(1.492)$ & $1.758$ $(1.513)$ \\
\midrule
\textbf{Average} & $4.069$ $(5.861)$ & $4.361$ $(6.061)$ & $3.933$ $(5.994)$ \\
\bottomrule
\end{tabular}
\caption{Detailed breakdown of Tab.~\ref{tab:sizing_metrics} for group G1.}
\label{tab:g1_metrics}
\end{table}

\begin{table}[h]
\centering
\scriptsize
\setlength{\tabcolsep}{2.7pt}
\begin{tabular}{l c c c}
\toprule
\textbf{Measurement} & \textbf{Per-Group} & \textbf{Multi-Task} & \textbf{Multi-Task + SwinV2} \\
\midrule
\textbf{Bottom Waist} & $4.177$ $(2.573)$ & $4.266$ $(2.607)$ & $3.572$ $(2.696)$ \\
\textbf{Bottom Hip} & $3.223$ $(2.249)$ & $3.609$ $(2.263)$ & $2.993$ $(2.291)$ \\
\textbf{Thigh} & $1.697$ $(1.302)$ & $2.027$ $(1.515)$ & $1.767$ $(1.316)$ \\
\textbf{Leg Cuff} & $2.459$ $(2.415)$ & $2.566$ $(2.669)$ & $1.991$ $(2.171)$ \\
\textbf{Front Crotch} & $2.561$ $(1.854)$ & $2.803$ $(1.977)$ & $1.946$ $(1.709)$ \\
\textbf{Back Crotch} & $2.552$ $(2.404)$ & $2.603$ $(2.402)$ & $2.784$ $(2.243)$ \\
\textbf{Leg Length} & $4.569$ $(3.315)$ & $4.493$ $(3.451)$ & $5.069$ $(3.264)$ \\
\textbf{Full Length} & $2.812$ $(2.480)$ & $3.117$ $(2.625)$ & $4.412$ $(3.664)$ \\
\midrule
\textbf{Average} & $3.006$ $(2.541)$ & $3.185$ $(2.619)$ & $3.067$ $(2.762)$ \\
\bottomrule
\end{tabular}
\caption{Detailed breakdown of Tab.~\ref{tab:sizing_metrics} for group G2.}
\label{tab:g2_metrics}
\end{table}

\begin{table}[h]
\centering
\scriptsize
\setlength{\tabcolsep}{2.7pt}
\begin{tabular}{l c c c}
\toprule
\textbf{Measurement} & \textbf{Per-Group} & \textbf{Multi-Task} & \textbf{Multi-Task + SwinV2} \\
\midrule
\textbf{Bottom Waist} & $6.325$ $(4.495)$ & $3.938$ $(3.331)$ & $4.367$ $(2.685)$ \\
\textbf{Bottom Hip} & $8.126$ $(4.236)$ & $3.894$ $(2.101)$ & $3.760$ $(2.766)$ \\
\textbf{Bottom Bottom} & $17.300$ $(12.239)$ & $15.506$ $(14.732)$ & $14.483$ $(17.524)$ \\
\textbf{Full Length} & $8.053$ $(5.055)$ & $4.049$ $(2.694)$ & $3.962$ $(3.377)$ \\
\midrule
\textbf{Average} & $9.951$ $(8.437)$ & $6.847$ $(9.174)$ & $6.643$ $(10.135)$ \\
\bottomrule
\end{tabular}
\caption{Detailed breakdown of Tab.~\ref{tab:sizing_metrics} for group G3.}
\label{tab:g3_metrics}
\end{table}

\begin{table}[h]
\centering
\scriptsize
\setlength{\tabcolsep}{2.7pt}
\begin{tabular}{l c c c}
\toprule
\textbf{Measurement} & \textbf{Per-Group} & \textbf{Multi-Task} & \textbf{Multi-Task + SwinV2} \\
\midrule
\textbf{Neck} & $1.768$ $(1.277)$ & $2.444$ $(2.284)$ & $1.815$ $(1.955)$ \\
\textbf{Chest} & $6.279$ $(4.443)$ & $7.493$ $(5.830)$ & $5.996$ $(4.814)$ \\
\textbf{Waist} & $4.656$ $(3.858)$ & $4.946$ $(5.611)$ & $3.721$ $(3.169)$ \\
\textbf{Bottom} & $6.131$ $(5.767)$ & $6.922$ $(6.804)$ & $5.171$ $(5.246)$ \\
\textbf{Sleeve} & $7.784$ $(6.336)$ & $5.717$ $(4.442)$ & $5.842$ $(4.986)$ \\
\textbf{Bicep} & $2.319$ $(1.907)$ & $2.116$ $(1.439)$ & $1.619$ $(1.300)$ \\
\textbf{Armhole} & $3.378$ $(2.535)$ & $2.655$ $(1.942)$ & $2.789$ $(2.070)$ \\
\textbf{Shoulder} & $4.795$ $(3.342)$ & $6.513$ $(5.562)$ & $4.960$ $(3.969)$ \\
\textbf{Body Height} & $11.390$ $(18.222)$ & $15.403$ $(20.404)$ & $11.106$ $(15.482)$ \\
\textbf{Sleeve Cuff} & $2.419$ $(2.500)$ & $2.230$ $(1.713)$ & $1.344$ $(1.314)$ \\
\midrule
\textbf{Average} & $5.092$ $(7.377)$ & $5.644$ $(8.563)$ & $4.436$ $(6.537)$ \\
\bottomrule
\end{tabular}
\caption{Detailed breakdown of Tab.~\ref{tab:sizing_metrics} for group G4.}
\label{tab:g4_metrics}
\end{table}

\begin{table}[h]
\centering
\scriptsize
\setlength{\tabcolsep}{2.7pt}
\begin{tabular}{l c c c}
\toprule
\textbf{Measurement} & \textbf{Per-Group} & \textbf{Multi-Task} & \textbf{Multi-Task + SwinV2} \\
\midrule
\textbf{Neck} & $8.058$ $(2.290)$ & $3.641$ $(2.401)$ & $1.132$ $(0.590)$ \\
\textbf{Chest} & $10.098$ $(1.426)$ & $4.768$ $(2.114)$ & $5.174$ $(1.231)$ \\
\textbf{Waist} & $10.615$ $(1.759)$ & $2.701$ $(1.207)$ & $2.721$ $(1.042)$ \\
\textbf{Bottom} & $11.753$ $(2.834)$ & $3.985$ $(2.504)$ & $3.945$ $(2.169)$ \\
\textbf{Armhole} & $2.083$ $(1.326)$ & $6.951$ $(1.245)$ & $10.211$ $(2.226)$ \\
\textbf{Shoulder} & $9.791$ $(1.785)$ & $6.528$ $(3.385)$ & $6.931$ $(3.096)$ \\
\textbf{Body Height} & $2.657$ $(1.690)$ & $6.099$ $(3.673)$ & $18.046$ $(3.976)$ \\
\textbf{Bottom Waist} & $13.330$ $(2.995)$ & $2.774$ $(1.139)$ & $3.435$ $(2.390)$ \\
\textbf{Bottom Hip} & $17.475$ $(2.204)$ & $2.865$ $(1.650)$ & $4.415$ $(2.980)$ \\
\textbf{Thigh} & $6.026$ $(2.525)$ & $1.842$ $(1.769)$ & $1.733$ $(0.960)$ \\
\textbf{Leg Cuff} & $7.653$ $(1.858)$ & $7.202$ $(1.799)$ & $2.856$ $(2.309)$ \\
\textbf{Front Crotch} & $3.712$ $(2.599)$ & $6.807$ $(1.926)$ & $4.859$ $(1.187)$ \\
\textbf{Back Crotch} & $10.100$ $(2.872)$ & $2.740$ $(1.297)$ & $2.812$ $(1.832)$ \\
\textbf{Leg Length} & $28.879$ $(6.945)$ & $2.204$ $(2.118)$ & $3.070$ $(1.579)$ \\
\textbf{Full Length} & $39.399$ $(9.268)$ & $3.314$ $(1.711)$ & $1.991$ $(1.654)$ \\
\midrule
\textbf{Average} & $12.109$ $(10.313)$ & $4.295$ $(2.746)$ & $4.889$ $(4.633)$ \\
\bottomrule
\end{tabular}
\caption{Detailed breakdown of Tab.~\ref{tab:sizing_metrics} for group G5. \textbf{Note:} \textit{Sleeve}, \textit{Bicep}, and \textit{Sleeve Cuff} measurement parts are missing because all instances in the test set are sleeveless.}
\label{tab:g5_metrics}
\end{table}

\begin{table}[h]
\centering
\scriptsize
\setlength{\tabcolsep}{2.7pt}
\begin{tabular}{l c c c}
\toprule
\textbf{Measurement} & \textbf{Per-Group} & \textbf{Multi-Task} & \textbf{Multi-Task + SwinV2} \\
\midrule
\textbf{Neck} & $4.285$ $(2.699)$ & $2.944$ $(2.135)$ & $2.995$ $(2.366)$ \\
\textbf{Chest} & $8.361$ $(10.686)$ & $6.902$ $(7.048)$ & $6.050$ $(5.117)$ \\
\textbf{Waist} & $7.555$ $(7.777)$ & $4.553$ $(3.660)$ & $3.700$ $(3.120)$ \\
\textbf{Bottom} & $8.080$ $(7.073)$ & $4.268$ $(3.630)$ & $3.787$ $(2.954)$ \\
\textbf{Sleeve} & $6.575$ $(4.483)$ & $4.183$ $(2.600)$ & $6.146$ $(4.889)$ \\
\textbf{Bicep} & $3.446$ $(2.441)$ & $2.957$ $(2.528)$ & $3.781$ $(2.119)$ \\
\textbf{Armhole} & $3.419$ $(2.241)$ & $3.965$ $(2.777)$ & $3.307$ $(2.515)$ \\
\textbf{Shoulder} & $10.579$ $(12.815)$ & $8.317$ $(8.678)$ & $7.850$ $(6.972)$ \\
\textbf{Body Height} & $12.486$ $(7.420)$ & $15.193$ $(11.383)$ & $13.174$ $(7.387)$ \\
\textbf{Sleeve Cuff} & $3.619$ $(2.373)$ & $2.041$ $(1.122)$ & $2.410$ $(1.326)$ \\
\textbf{Bottom Waist} & $8.159$ $(8.206)$ & $5.857$ $(4.638)$ & $5.724$ $(5.929)$ \\
\textbf{Bottom Hip} & $9.029$ $(6.733)$ & $7.764$ $(7.555)$ & $8.634$ $(8.486)$ \\
\textbf{Bottom Bottom} & $14.577$ $(9.603)$ & $13.631$ $(10.468)$ & $10.106$ $(9.316)$ \\
\textbf{Full Length} & $5.804$ $(4.066)$ & $6.116$ $(3.966)$ & $9.826$ $(8.118)$ \\
\midrule
\textbf{Average} & $7.767$ $(7.954)$ & $6.549$ $(7.234)$ & $6.389$ $(6.562)$ \\
\bottomrule
\end{tabular}
\caption{Detailed breakdown of Tab.~\ref{tab:sizing_metrics} for group G6.}
\label{tab:g6_metrics}
\end{table}

\begin{table*}[h]
\centering
\setlength{\tabcolsep}{4pt}
\begin{tabular}{l c c c c}
\toprule
\textbf{Group} & \textbf{Per-Group} & \textbf{Multi-Task} & \textbf{Multi-Task + SwinV2} & \textbf{End-to-End} \\
\midrule
\textbf{G1} & $4.069$ $(5.861)$ & $4.361$ $(6.061)$ & $3.933$ $(5.994)$ & $5.316$ $(6.672)$ \\
\textbf{G2} & $3.006$ $(2.541)$ & $3.185$ $(2.619)$ & $3.067$ $(2.762)$ & $3.965$ $(3.789)$ \\
\textbf{G3} & $9.951$ $(8.437)$ & $6.847$ $(9.174)$ & $6.643$ $(10.135)$ & $10.565$ $(13.070)$ \\
\textbf{G4} & $5.092$ $(7.377)$ & $5.644$ $(8.563)$ & $4.436$ $(6.537)$ & $5.924$ $(6.705)$ \\
\textbf{G5} & $12.109$ $(10.313)$ & $4.295$ $(2.746)$ & $4.889$ $(4.633)$ & $6.347$ $(5.192)$ \\
\textbf{G6} & $7.767$ $(7.954)$ & $6.549$ $(7.234)$ & $6.389$ $(6.562)$ & $11.076$ $(11.976)$ \\
\midrule
\textbf{Average} & $4.904$ $(6.533)$ & $4.710$ $(6.392)$ & $4.279$ $(5.870)$ & $6.134$ $(7.832)$ \\
\bottomrule
\end{tabular}
\caption{Quantitative results for the size regressor ($\Phi$). We report MAE in cm as mean (std). This table expands Tab.~\ref{tab:sizing_metrics} with the \textit{End-to-End} baseline, which regresses size directly from the input image. The comparison assesses the performance with and without relying on ground truth canonical garment normals ($G^t$). Groups and model variants are defined in Sec.~\ref{sec:annotations} and Sec.~\ref{sec:size_baseline} of the main paper, respectively.}
\label{tab:suppl_sizing}
\end{table*}

\subsection{Novel View Synthesis}

\begin{figure}[!h]
    \centering
    \includegraphics[width=\linewidth]{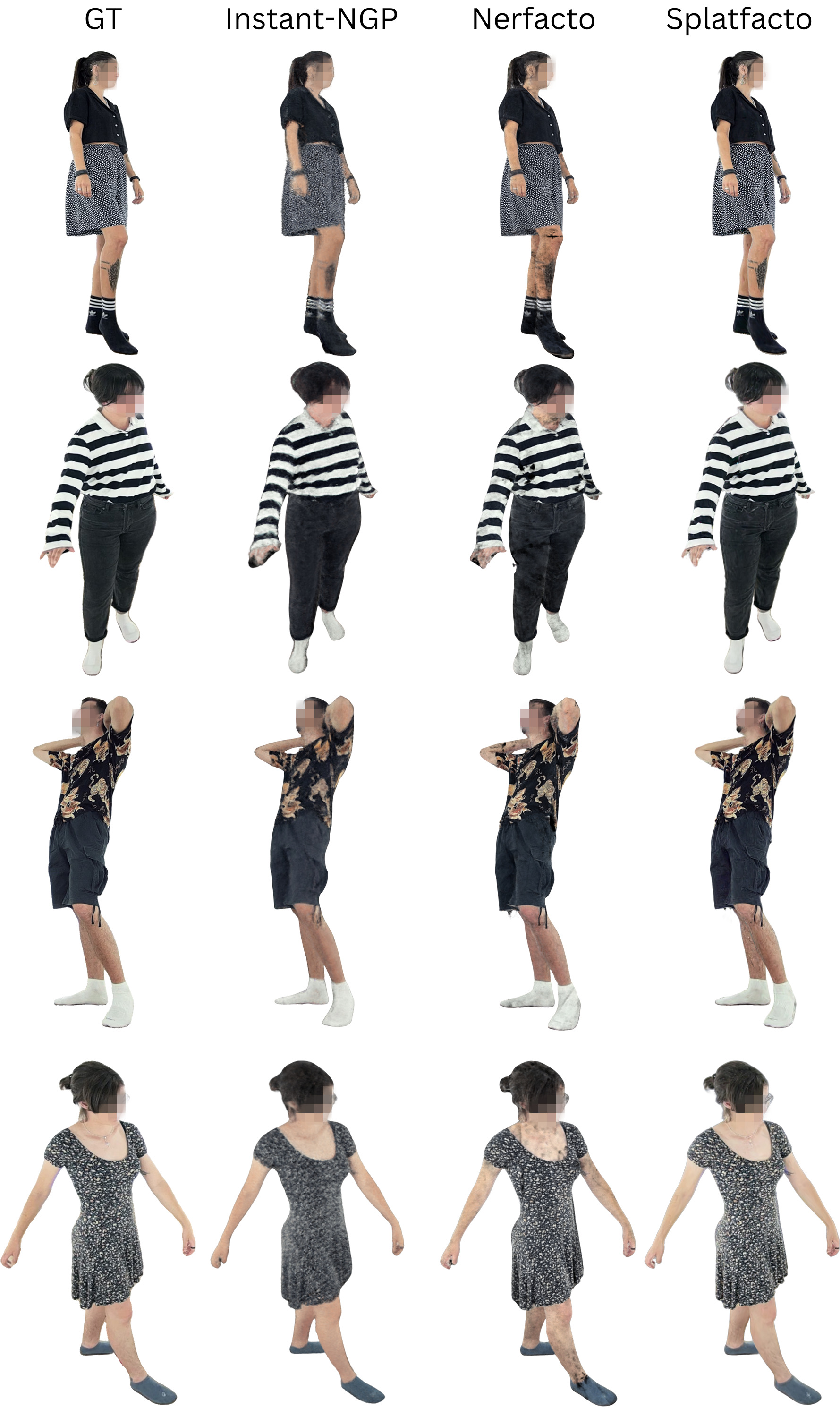}
    \caption{Additional examples of the three NVS methods tested. Instant-NGP produces blurry results, while \texttt{nerfacto} is sharper, but it has some distinct artifacts. \texttt{splatfacto} on the other hand provides superior quality.}
    \label{fig:supp_nvs_samples}
\end{figure}

\begin{figure}[!h]
    \centering
    \includegraphics[width=\linewidth]{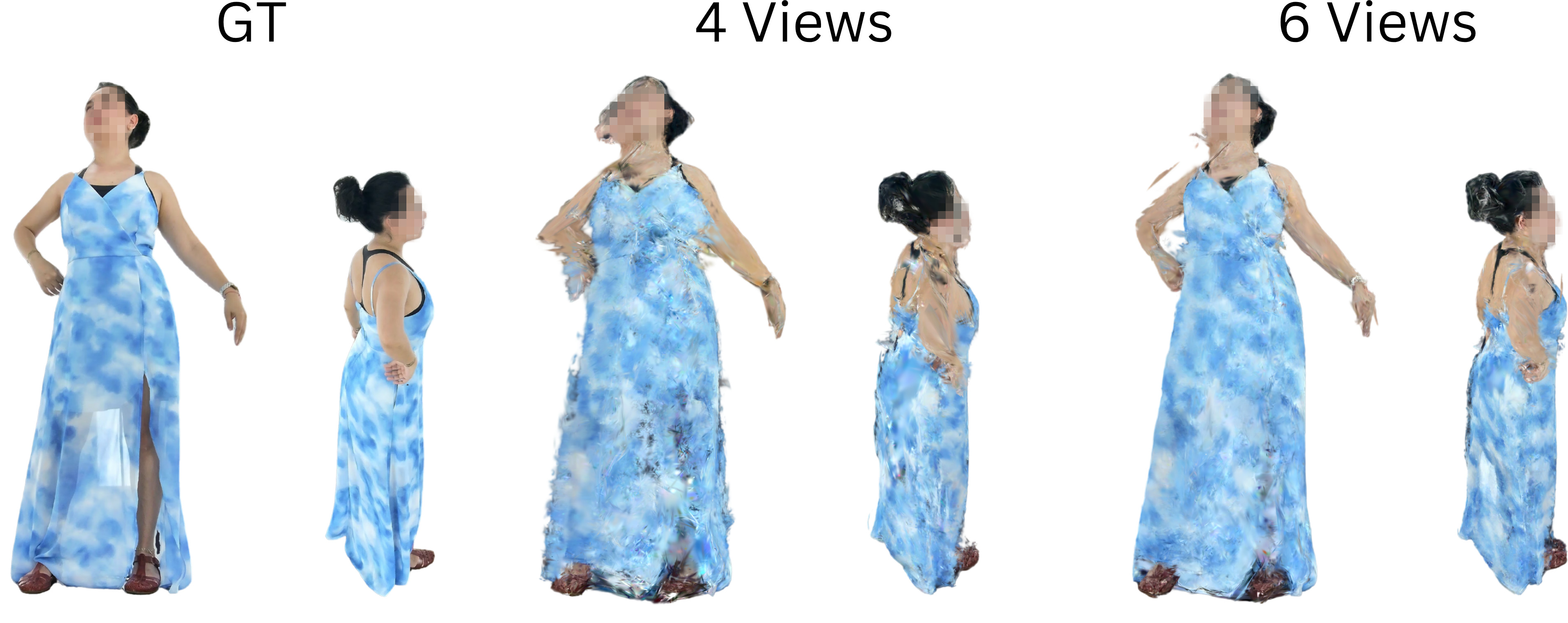}
    \caption{Qualitative results of the ablation over the number of views used for \texttt{splatfacto} when using only Bolt cameras. Because of the limited number of available cameras there are many artifacts. }
    \label{fig:supp_nvs_ablation_bolt}
\end{figure}

\begin{figure*}[h]
    \centering
    \includegraphics[width=\linewidth]{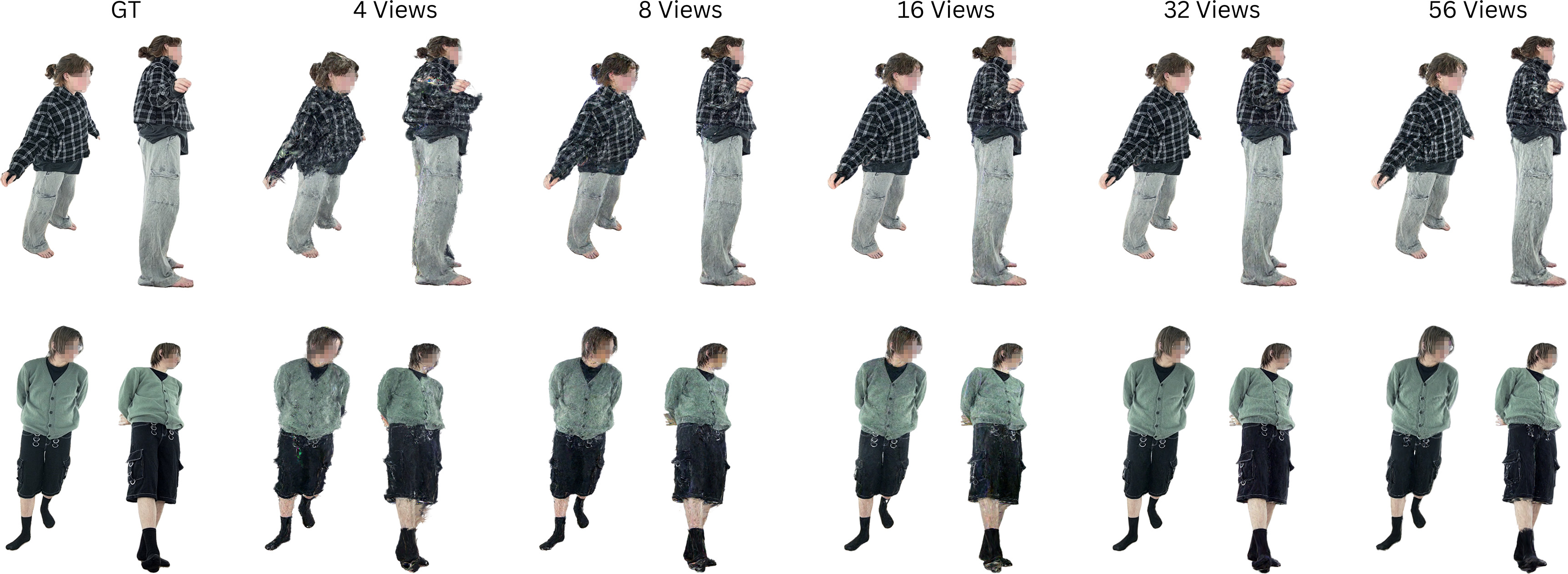}
    \caption{Qualitative results of the ablation over the number of views used for \texttt{splatfacto}. It shows that using very few views produces artifacts, but from 16 views it starts to improve and when using all views the results are almost indistinguishable from the ground truth.}
    \label{fig:supp_nvs_ablation}
\end{figure*}

\textbf{Training Setup.} To train each of the baseline methods, we run them for 50,000 iteration. Unless specified, we use the default parameters. For \textbf{instant-ngp}, we had to reduce the \textit{--pipeline.target-num-samples}: "The target number of samples to use for an entire batch of rays" value to 16,384 due to the implementation, which uses \texttt{nerfacc} library. This initializes the target volume in the beginning of the optimization, and with a large number of rays it requires significant amount of video memory. For \textbf{nerfacto and splatfacto} we use their "big" variants, namely \texttt{nerfacto-big} and \texttt{splatfacto-big} which increases the number of parameters used and achieves slightly better results. 

\textbf{Ablation Study.} We perform an ablation study over the effect of the number of views used for training. We evaluate how the quality of the final results compares when using a very low number of input views (4), increase the number of views until we add all training views. All tests are performed on the RPi cameras only for consistency, but we also evaluate the higher resolution Bolt cameras. Since we only have 8 of Bolts available, the ablation is limited, but we test training on 4 random or 6 total Bolt cameras and evaluate on the remaining two. The metrics we use are affected by the resolution change. Therefore, they are not comparable between RPi and Bolt experiments. We restrict evaluation to same-resolution settings to ensure metric consistency.

\textbf{Comparison of Baseline Methods.}
As discussed in the main paper, instant-ngp struggles with the fine details and even some geometrical features are missing or present artifacts. \texttt{nerfacto} reproduces the details better, though it is still prone to showing some geometrically incorrect artifacts. On the other hand, \texttt{splatfacto} can reproduce fine detail and is more geometrically correct. It can still present some artifacts specific to 3d gaussian splatting when the novel viewpoints are highly different compared to the training views. We present additional qualitative results from more subjects and from multiple views in Fig.~\ref{fig:supp_nvs_samples}.

\textbf{Ablation of Different Number of Training Views.}
As shown in Tab.~\ref{tab:nvs_metrics}, adding more training views improves the results as expected, though the results start saturating with more views. Performance saturation at higher view counts indicates diminishing returns. This suggests that the 60-camera setup provides an optimal trade-off between reconstruction fidelity and system complexity. We show some qualitative examples in Fig.~\ref{fig:supp_nvs_ablation}.

Additionally, we evaluate the results with only the Bolt cameras, as can be seen qualitatively in Fig.~\ref{fig:supp_nvs_ablation_bolt} and quantitatively in Tab.~\ref{tab:supp_nvs_ablation_bolt}. One can observe that \texttt{splatfacto} produces similar quality images, using the higher resolution Bolt cameras, to RPis. This is despite the fact that all Bolt cameras are at a low or high viewing angle, and none have a straight, perpendicular view to the subject. Since there are only 8 Bolt cameras designed to be spatially distributed, artifacts are visible in the validation views, as these are highly different from the training ones.

\begin{table}[h]
\centering
\scriptsize
\begin{tabular}{cccc}
\toprule
\textbf{Views} & \textbf{PSNR $\uparrow$} & \textbf{SSIM $\uparrow$}& \textbf{LPIPS $\downarrow$} \\
\midrule
\textbf{4}     & $24.089 (1.806)$ & $0.952 (0.012)$ & $0.073 (0.016)$  \\
\textbf{6}     & $25.831 (1.777)$ & $0.956 (0.012)$ & $0.061 (0.013)$ \\
\bottomrule
\end{tabular}
\caption{Image quality metrics when using only Bolt cameras with \texttt{splatfacto}. Note that these values are not directly comparable with Tab.~\ref{tab:nvs_metrics} due to the resolution difference. Also, because of the limited number of cameras, these values are relatively low.}
\label{tab:supp_nvs_ablation_bolt}
\end{table}

\begin{figure*}[h]
    \centering
    \includegraphics[width=0.99\linewidth]{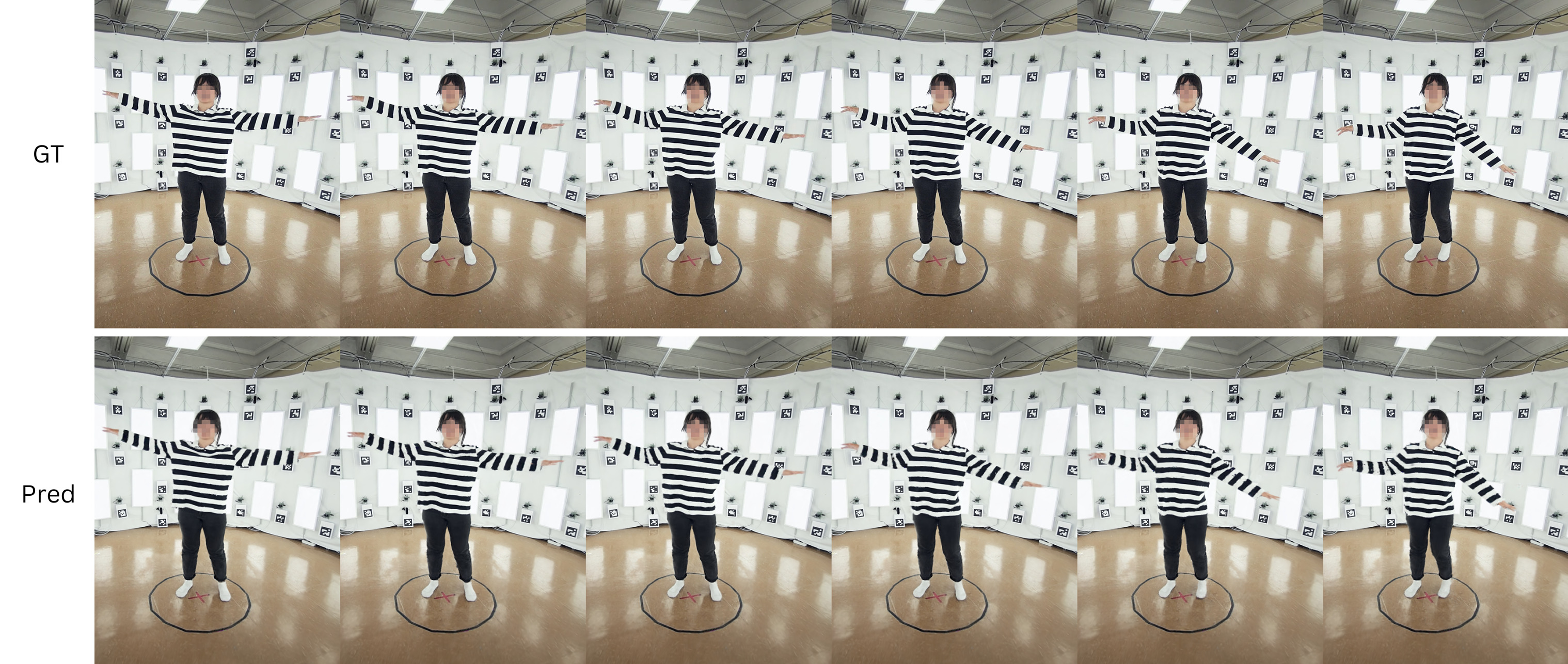}
    \caption{Qualitative results of the dynamic novel view synthesis experiments. The results show slight blurring around the hands, but otherwise prvode good quality outputs. The frames shown are 6 consecutive frames from a 20 frame sequence.}
    \label{fig:supp_nvs_4dgs}
\end{figure*}

\textbf{4D Gaussian Splatting}
We run CEM-4DGS~\cite{kangClusteredErrorCorrection2025} with the default settings from the authors to validate our dataset for the application of dynamic novel view synthesis. We ran the method on a sequence of 20 frames from which the first seven can be seen in Fig.~\ref{fig:supp_nvs_4dgs}. This shows that while parts of the person with more movement like the hands are slightly blurry, the overall reconstruction is of acceptable quality. CEM-4DGS and similar methods treat static and dynamic elements of the scene separately, as such, they produce a sharp background similar to our 3DGS tests. Quantitative results can be seen in Tab.~\ref{tab:supp_nvs_4dgs}.

\begin{table}[h]
\centering
\scriptsize
\begin{tabular}{cccc}
\toprule
\textbf{Method} & \textbf{PSNR $\uparrow$} & \textbf{SSIM $\uparrow$}& \textbf{LPIPS $\downarrow$} \\
\midrule
\textbf{CEM-4DGS}     & $24.395$ & $0.875$ & $0.151$  \\
\bottomrule
\end{tabular}
\caption{Image quality metrics of the 4DGS reconstruction. As expected, slightly lower than static reconstruction, but still good quality. Values shown are means across all frames of the sequence.}
\label{tab:supp_nvs_4dgs}
\end{table}

\textbf{Conclusion.}
Even with our relatively low-resolution cameras, we are able to achieve highly realistic novel view synthesis, which leads us to believe that MV-Fashion is suited for downstream tasks involving the discussed methods.